\documentclass{article}

\usepackage{microtype}
\usepackage{graphicx}
\usepackage{booktabs} 
\usepackage{fontawesome5} 

\usepackage{hyperref}

\usepackage[accepted]{icml2026}

\usepackage{amsmath}
\usepackage{amssymb}
\usepackage{mathtools}
\usepackage{amsthm}
\usepackage[utf8]{inputenc}
\usepackage[T1]{fontenc}
\usepackage{url}
\usepackage{empheq}
\usepackage{booktabs}
\usepackage{amsmath}
\usepackage{amsfonts}
\usepackage{nicefrac}
\usepackage{microtype}
\usepackage{graphicx}
\usepackage{makecell}
\usepackage{dblfloatfix}

\usepackage{fontawesome5}

\usepackage[capitalize,noabbrev,nameinlink]{cleveref}

\theoremstyle{plain}

\theoremstyle{definition}

\theoremstyle{remark}

\usepackage[textsize=tiny]{todonotes}

\usepackage[separate-uncertainty=true,reset-math-version=false,detect-weight=true,detect-family=true]{siunitx}
\usepackage{silence}
\usepackage[utf8]{inputenc}
\usepackage[T1]{fontenc}
\usepackage{tabularx}
\usepackage{lipsum}
\usepackage{url}
\usepackage{placeins}
\usepackage{booktabs}
\usepackage{soul}
\usepackage{amsfonts}
\usepackage{nicefrac}
\usepackage{microtype}
\usepackage{wrapfig}
\usepackage{caption}
\usepackage{amssymb}
\usepackage{pifont}

\usepackage{subcaption}
\usepackage{float}
\usepackage{rotating}
\usepackage{etoolbox}
\usepackage{xparse}
\usepackage{grffile}
\usepackage{amsmath}
\usepackage{xspace}
\usepackage{euscript}
\usepackage{enumitem}
\usepackage{mathtools}
\usepackage{tikz}
\usepackage{tablefootnote}
\usepackage[flushleft]{threeparttable}
\usepackage{multirow}
\usepackage[title]{appendix}
\usepackage{arydshln}
\usepackage{array, booktabs}
\usepackage{comment}
\usepackage{graphicx}
\usepackage{adjustbox}
\usepackage{amsmath,amsthm}
\usepackage{dsfont}
\usepackage{balance}
\usepackage{multicol}
\usepackage{xcolor}
\usepackage{nameref}
\usepackage{pdfpages}
\usepackage{braket}
\usepackage[normalem]{ulem}
\usepackage{bbding}
\usepackage{xfrac}
\usepackage[usestackEOL]{stackengine}
\usepackage{indentfirst}
\usepackage{csquotes}
\usepackage{pgf}
\usepackage{varwidth}
\usepackage{verbatimbox}
\usepackage{verbatim}
\usepackage{bm}
\usepackage{anyfontsize}
\usepackage{todonotes}
\usepackage[svgnames]{xcolor}

\usepackage{empheq}
\definecolor{myblue}{rgb}{.8, .8, 1}

\definecolor{pastelblue}{RGB}{76,113,175}
\definecolor{pastelgreen}{RGB}{144,238,144}
\definecolor{pastelred}{RGB}{196,78,82}
\definecolor{pastelgrey}{RGB}{230,230,230}
\definecolor{pastelbeige}{RGB}{243,236,221}
\definecolor{pastelpurple}{RGB}{154,139,192}
\definecolor{salmon}{RGB}{250, 128, 114}
\definecolor{darkgreen}{rgb}{0,0.6,0}
\definecolor{darkred}{rgb}{0.5,0,0}
\definecolor{verylightgreen}{HTML}{F6FFF9}
\definecolor{verylightred}{HTML}{FFF4F3}
\definecolor{verylightgray}{HTML}{F4F6F6}
\definecolor{babyblueeyes}{rgb}{0.63, 0.79, 0.95}
\definecolor{lightpink}{rgb}{1.00, 0.714, 0.757}

\makeatletter
\def\thmt@refnamewithcomma #1#2#3,#4,#5\@nil{%
	\@xa\def\csname\thmt@envname #1utorefname\endcsname{#3}%
	\ifcsname #2refname\endcsname
	\csname #2refname\expandafter\endcsname\expandafter{\thmt@envname}{#3}{#4}%
	\fi}
\makeatother
\crefname{equation}{Equation}{Equations}
\crefname{section}{Section}{Sections}
\crefname{corollary}{Corollary}{Corollaries}
\crefname{proposition}{Proposition}{Propositions}
\crefname{appendix}{Appendix}{Appendices}
\crefname{theorem}{Theorem}{Theorems}
\crefname{figure}{Figure}{Figures}
\crefname{tabular}{Table}{Tables}
\crefname{algorithm}{Algorithm}{Algorithm}
\Crefname{conjecture}{Conjecture}{Conjectures}
\Crefname{definition}{Definition}{Definitions}
\Crefname{observation}{Observation}{Observations}
\Crefname{assumption}{Assumption}{Assumptions}
\Crefname{axiom}{Axiom}{Axioms}
\Crefname{case}{Case}{Cases}
\Crefname{claim}{Claim}{Claims}
\Crefname{conclusion}{Conclusion}{Conclusions}
\Crefname{condition}{Condition}{Conditions}
\Crefname{criterion}{Criterion}{Criteria}
\Crefname{exercise}{Exercise}{Exercises}
\Crefname{example}{Example}{Examples}
\Crefname{notation}{Notation}{Notations}
\Crefname{problem}{Problem}{Problems}
\Crefname{property}{Property}{Properties}
\Crefname{remark}{Remark}{Remarks}
\Crefname{solution}{Solution}{Solutions}
\Crefname{summary}{Summary}{Summaries}
\Crefname{motivation}{Motivation}{Motivations}
\Crefname{query}{Query}{Queries}

\newcommand*\dbar[1]{\overline{\overline{\lower0.2ex\hbox{$#1$}}}}

\def\ba{{\bm{a}}}

\def\bh{{\bm{h}}}

\def\bp{{\bm{p}}}

\def\bv{{\bm{v}}}

\def\bx{{\bm{x}}}

\DeclareFontFamily{U}{BOONDOX-calo}{\skewchar\font=45 }
\DeclareFontShape{U}{BOONDOX-calo}{m}{n}{
  <-> s*[1.05] BOONDOX-r-calo}{}
\DeclareFontShape{U}{BOONDOX-calo}{b}{n}{
  <-> s*[1.05] BOONDOX-b-calo}{}
\DeclareMathAlphabet{\mathcalb}{U}{BOONDOX-calo}{m}{n}
\SetMathAlphabet{\mathcalb}{bold}{U}{BOONDOX-calo}{b}{n}
\DeclareMathAlphabet{\mathbcalb}{U}{BOONDOX-calo}{b}{n}

\usepackage{xfp}
\ExplSyntaxOn
\newcommand{\printdimall}[2][4]{%
  \begingroup\leavevmode
  \fpeval{ round( \dim_to_fp:n {#2} , #1 ) }\,pt%
  \space(%
    \fpeval{ round( \dim_to_fp:n {#2} / \dim_to_fp:n {1cm} , #1 ) }\,cm; %
    \fpeval{ round( \dim_to_fp:n {#2} / \dim_to_fp:n {1in} , #1 ) }\,in%
  )%
  \endgroup
}

\ExplSyntaxOff

\newcommand\thefontsize{The current font size is: \f@size pt}

\newcommand{\pos}{{\bx}}
\newcommand{\vel}{{\bv}}
\newcommand{\mvel}{{\bar{\bv}}}
\newcommand{\acc}{{\ba}}
\newcommand{\force}{{\boldsymbol{f}}}
\newcommand{\mforce}{{\bar{\boldsymbol{f}}}}
\newcommand{\mass}{{\boldsymbol{m}}}
\newcommand{\mom}{{\bp}}

\renewcommand{\t}{{\Delta t}}
\newcommand{\hamiltonian}{{\mathcal{H}}}
\newcommand{\potential}{{V}}
\newcommand{\kinetic}{T}

\newcommand{\hmap}{u}
\newcommand{\hdisp}{\bar{\hmap}}
\newcommand{\params}{{\boldsymbol{\theta}}}

\newcommand{\apos}{{t}}
\newcommand{\bpos}{{t^*}}
\newcommand{\dense}{{\boldsymbol{W}}}
\newcommand{\MLP}{{\boldsymbol{\mathrm{MLP}}}}
\newcommand{\softmax}{{\boldsymbol{\mathrm{softmax}}}}
\newcommand{\emb}{{\boldsymbol{\mathrm{Embed}}}}
\newcommand{\rpe}{{\boldsymbol{e}_{ij}^{\text{rPE}}}}
\newcommand{\pe}{{\boldsymbol{e}_{ij}^{\text{PE(3)}}}}
\newcommand{\AdaLN}{{\mathrm{AdaLN}}}
\newcommand{\AdaScale}{{\mathrm{AdaScale}}}

\newlength{\maxwidth}
\newcommand{\algalign}[2]
{\makebox[\maxwidth][c]{$#1{}$}${}#2$}

\usepackage[acronym,nowarn,section,nogroupskip,nonumberlist,nolist]{glossaries}
\glsdisablehyper
\newglossaryentry{SDE}
{
  name={SDE},
  description={stochastic differential equation},
  first={stochastic differential equation (SDE)},
  plural={SDEs},
  descriptionplural={stochastic differential equations},
  firstplural={stochastic differential equations (SDEs)}
} 

\newglossaryentry{FF}
{
  name={FF},
  description={force field},
  first={force field (FF)},
  plural={FFs},
  descriptionplural={force fields},
  firstplural={force fields (FFs)}
} 
\newglossaryentry{MLFF}
{
  name={MLFF},
  description={machine-learned force field},
  first={machine-learned force field (MLFF)},
  plural={MLFFs},
  descriptionplural={machine-learned force fields},
  firstplural={machine-learned force fields (MLFFs)}
}

\newacronym{QM}{QM}{quantum mechanical}
\newacronym{BG}{BG}{Boltzmann generator}
\newacronym{MD}{MD}{molecular dynamics}
\newacronym{CG}{CG}{coarse-graining}
\newacronym{JS}{JS}{Jensen-Shannon}
\newacronym{PMF}{PMF}{potential of mean force}
\newacronym{MCMC}{MCMC}{Markov chain Monte Carlo}
\newacronym{TICA}{TICA}{time-lagged independent component analysis}
\newacronym{PWD}{PWD}{pairwise distance}

\definecolor{cite_color}{HTML}{114083}
\definecolor{url_color}{RGB}{153, 102,  0}
\hypersetup{
 colorlinks = true,
 citecolor = cite_color,
 linkcolor = url_color,
 urlcolor = black
}

\icmltitlerunning{Learning Hamiltonian Flow Maps: Mean Flow Consistency for Large-Timestep Molecular Dynamics}

\def\ourCode{\url{https://ml4molsim.github.io/hamiltonian-flow-maps}}

\begin{document}

\twocolumn[
\icmltitle{Learning Hamiltonian Flow Maps:\\Mean Flow Consistency for Large-Timestep Molecular Dynamics}



\icmlsetsymbol{equal}{*}

\begin{icmlauthorlist}
\icmlauthor{Winfried Ripken}{equal,tu,bifold}
\icmlauthor{Michael Plainer}{equal,fu,eliza,tu,bifold}
\icmlauthor{Gregor Lied}{equal,tu}
\icmlauthor{J. Thorben Frank}{equal,tu,bifold}
\end{icmlauthorlist}
\begin{icmlauthorlist}
\icmlauthor{Oliver T. Unke}{google}
\icmlauthor{Stefan Chmiela}{tu,bifold}
\icmlauthor{Frank Noé}{fu,rice,microsoft}
\icmlauthor{Klaus-Robert Müller}{tu,bifold,google,mpi,korea}
\end{icmlauthorlist}

\icmlaffiliation{tu}{Technical University Berlin}
\icmlaffiliation{fu}{Free University of Berlin}
\icmlaffiliation{eliza}{Zuse School ELIZA}
\icmlaffiliation{bifold}{BIFOLD Berlin}
\icmlaffiliation{google}{Google DeepMind}
\icmlaffiliation{microsoft}{Microsoft Research AI4Science}
\icmlaffiliation{mpi}{MPI for Informatics, Saarbrücken}
\icmlaffiliation{korea}{Department of Artificial Intelligence, Korea University}
\icmlaffiliation{rice}{Rice University}

\icmlcorrespondingauthor{Thorben Frank}{thorben.frank@tu-berlin.de}
\icmlcorrespondingauthor{Frank Noé}{franknoe@microsoft.com}
\icmlcorrespondingauthor{Klaus-Robert Müller}{klaus-robert.mueller@tu-berlin.de}

\icmlkeywords{Machine Learning, Hamiltonian Mechanics, Molecular Dynamics, Integration, Flow Maps, Mean Flow}

\vskip 0.3in
]



\printAffiliationsAndNotice{\icmlEqualContribution} 

\begin{abstract}
Simulating the long-time evolution of Hamiltonian systems is limited by the small timesteps required for stable numerical integration. To overcome this constraint, we introduce a framework to learn \emph{Hamiltonian Flow Maps} by predicting the \emph{mean} phase-space evolution over a chosen time span~$\t$, enabling stable large-timestep updates far beyond the stability limits of classical integrators. To this end, we impose a \emph{Mean Flow} consistency condition for time-averaged Hamiltonian dynamics. Unlike prior approaches, this allows training on independent phase-space samples without access to future states, avoiding expensive trajectory generation. Validated across diverse Hamiltonian systems, our method in particular improves upon molecular dynamics simulations using machine-learned force fields (MLFF). Our models maintain comparable training and inference cost, but support significantly larger integration timesteps while trained directly on widely-available \emph{trajectory-free} MLFF datasets. Our code, model weights, and self-contained JAX and PyTorch notebooks are available at \ourCode{}.
\end{abstract}

\section{Introduction}

Solving Hamilton's equations of motion is central to modeling physical systems, with \gls{MD} simulations as a prominent application, where atomic motion is governed by interatomic forces~\cite{hollingsworth2018molecular, goldstein2001classical}. In practice, numerical integration of these equations requires small timesteps~$\t$ to ensure stability, rendering long-time simulations computationally expensive~\cite{hairer2006geometric}. To reduce cost, forces are commonly computed using fast but approximate force fields~\cite{frenkel2002understanding} as a replacement for highly accurate but computationally prohibitive \gls{QM} methods~\cite{tuckerman2000understanding}. While \glspl{MLFF} have started to bridge the gap between \gls{QM} accuracy and computational cost~\cite{noe2020ml4ms, keith2021combining, unke2021machine, bonneau2026breaking}, the integration bottleneck remains, limiting the accessible time scales~\cite{wang2025design}.

As a result, accelerating \glspl{MLFF} is an active area of research, focusing on efficient architectures~\cite{xie2023ultra, frank2024euclidean} and improved implementations~\cite{pelaez2024torchmd, park2024scalable}. An orthogonal line of work bypasses integration by predicting future states directly~\citep{bigi2025flashmd,thiemann2025forcefree,thompson2025atom,diez2025transferable}. However, these approaches rely on reference trajectories for training, which are prohibitively expensive to generate with \emph{ab-initio} \gls{QM} methods. To mitigate this, \citet{bigi2025flashmd} propose training an \gls{MLFF} on \emph{ab-initio} data and then using it as a teacher to generate trajectories for distillation. While effective, this strategy remains costly for chemically diverse datasets, introduces teacher biases, and requires retraining whenever the teacher changes.

\begin{figure*}[t]
    \centering
    \includegraphics{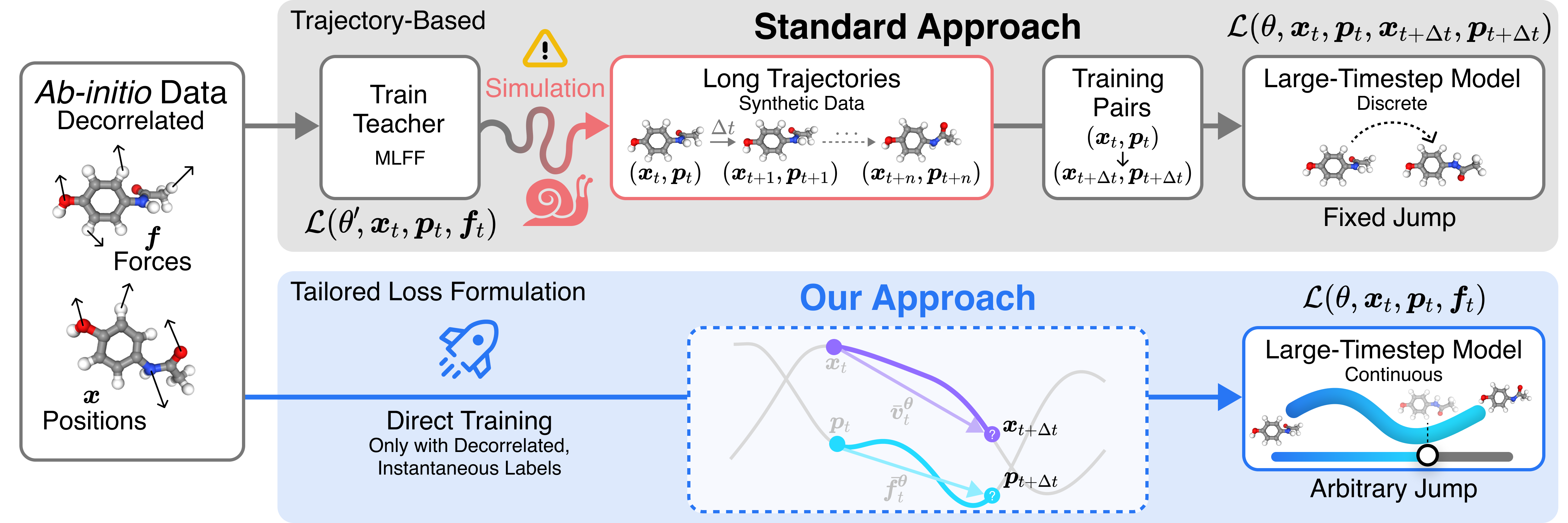}
    \caption{Hamiltonian Flow Maps (HFMs) for large timesteps in phase space. \textbf{Top row}: 
    Existing approaches rely on trajectory data, typically generated using a teacher \gls{MLFF} through sequential simulation with small timesteps. While this enables training large-timestep models via direct regression, the resulting models are limited to a fixed set of predefined timesteps that cannot be adjusted after training.
    \textbf{Bottom row}:
    Our approach learns continuous-time, large-timestep dynamics directly from decorrelated \emph{ab-initio} samples, without requiring trajectories. The model supports arbitrary timesteps at inference. Our tailored loss combines force matching with a consistency constraint that enforces agreement of the predicted flow across different time horizons (see \cref{sec:computational-approach}).}
    \vspace{-0.4cm}
    \label{fig:intro-figure}
\end{figure*}

This work proposes an alternative approach: Rather than regressing future states from reference trajectories, we learn the cumulative dynamics directly from a single phase-space configuration and its instantaneous time derivative. Since Hamilton's equations are deterministic, this information is sufficient to learn the correct \emph{Hamiltonian flow map}. We reframe learning accurate large-timestep dynamics as optimizing a consistency condition: Inspired by the seminal work of \emph{Flow Maps} for generative modeling~\citep{boffi2025flow}, we adapt the mathematical formalism of \emph{Mean Flows}~\citep{geng2025mean}, originally designed for stochastic generative modeling, to the deterministic integration of Hamilton's equations. This objective requires \emph{neither numerical integration nor time-ordered data}. We can therefore train directly on standard \gls{MLFF} datasets, which typically consist of decorrelated molecular geometries with force labels~\cite{eastman2023spice, ganscha2025qcml, levine2025open}, without relying on a teacher \gls{MLFF} or trajectory generation. The same objective recovers instantaneous forces as well as their time-averaged counterparts, yielding a single model that functions both as a conventional force field and as a large-timestep integrator. Consequently, our models rely purely on \emph{ab-initio} data (see \cref{fig:intro-figure}) and can be trained with a compute budget comparable to standard \glspl{MLFF}~\citep{batzner2022e3}.

Our contributions can be summarized as follows:
\begin{itemize}[topsep=0pt, partopsep=0pt, itemsep=0pt, parsep=0pt,leftmargin=10pt]
    \item We derive a trajectory-free training objective to learn large-timestep Hamiltonian flow maps from single-time phase-space samples via a consistency condition.
    \item We show that our method can be trained on widely available \gls{MLFF} datasets, which typically contain independent snapshots rather than equilibrium samples.
    \item We apply our method to molecular dynamics, enabling stable rollouts at timesteps far beyond the stability limits of standard \glspl{MLFF} while maintaining high accuracy.
\end{itemize}
\section{Background}
\subsection{Hamiltonian Mechanics}
Hamiltonian mechanics \citep{goldstein2001classical} provide a unifying description for a wide range of physical systems, including harmonic oscillators, rigid-body dynamics, gravitational $N$-body systems, and molecular systems governed by interatomic potentials. In all cases, the system state evolves in phase space according to an energy function, the \emph{Hamiltonian} $\hamiltonian$, where the time evolution is governed by
\begin{equation}
\vel \coloneqq \dot{\pos} = \frac{\partial \hamiltonian (\pos,\mom)}{\partial \mom}, \qquad
\force \coloneqq \dot{\mom} = -\frac{\partial \hamiltonian (\pos,\mom)}{\partial \pos},
\label{eq:hamilton_eqs}
\end{equation}
with $\pos$ and $\mom$ denoting positions and momenta, and $\vel$ and $\force$ the corresponding instantaneous velocities and forces. These dynamics are fully coupled, meaning that changes in positions affect the evolution of momenta and vice versa.

The state at a later time $\bpos$ can be obtained by integrating these instantaneous quantities over a time interval $[\apos,\bpos]$,
\begin{equation}
\begin{pmatrix}
\pos_\bpos\\
\mom_\bpos
\end{pmatrix}
=
\begin{pmatrix}
\pos_\apos\\
\mom_\apos
\end{pmatrix}
+
\int_\apos^\bpos
\begin{pmatrix}
\vel_\tau \\
\force_\tau
\end{pmatrix}
\mathrm{d}\tau ,
\label{eq:forward_hamiltonian_map}
\end{equation}
with $\vel_\tau$ and $\force_\tau$ denote the instantaneous velocity and force along the trajectory at time $\tau$. \cref{eq:forward_hamiltonian_map} is an exact identity, where classical numerical integrators approximate this integral by evaluating $\vel_\tau$ and $\force_\tau$ at discrete timesteps.

\subsection{Molecular Dynamics} \label{sec:background-md}
\glsreset{MD} 
\Gls{MD} are a concrete instantiation of Hamiltonian mechanics in which the Hamiltonian $\hamiltonian$ is assumed to be separable and takes the form
\begin{equation}
\hamiltonian(\pos,\mom) = \kinetic(\mom) + \potential(\pos),
\end{equation}
with kinetic energy $\kinetic(\mom) = \sum_{i=1}^{N} \tfrac{\|\mom^{(i)}\|^2_2}{2 \mass^{(i)}}$, where $\mom^{(i)}$ and $\mass^{(i)}$ denote the momentum and mass of particle $i$, and the potential energy $\potential(\pos)$ depends only on the positions $\pos$. Under this assumption, the equations of motion reduce to
\begin{equation}
\vel \coloneqq \frac{\partial \kinetic(\mom)}{\partial \mom} = \frac{\mom}{\mass}, \qquad
\force \coloneqq -\frac{\partial \potential(\pos)}{\partial \pos}.
\end{equation}
Conventional \gls{MD} simulations model particle interactions using an empirical \emph{force field} that parameterizes the potential energy $\potential(\pos)$, and integrate the resulting equations of motion with explicit numerical schemes, such as Velocity Verlet, advancing the system through many small timesteps of size $\Delta t$~\citep{frenkel2002understanding}.
\section{Related Work}
\textbf{Designing faster \glspl{MLFF}.}
\Glspl{MLFF} have emerged as a powerful tool to obtain cheap but \gls{QM}-accurate MD simulations~\citep{behler2007generalized, chmiela2017machine,schutt2018schnet,batzner2022e3}. Although much faster than \gls{QM} methods, simulations with \glspl{MLFF} remain limited by computational cost, restricting sampling and applications to large systems~\citep{wang2025design}. Recent work has focused on improving \gls{MLFF} efficiency via architectural advances~\citep{xie2023ultra, frank2024euclidean}, kernel optimization~\cite{pelaez2024torchmd}, or GPU parallelization~\citep{musaelian2023learning,park2024scalable}. In contrast, our approach accelerates simulations by increasing the integration timestep instead of speeding up models.

\textbf{Learning larger timesteps from trajectories.}
A complementary line of work accelerates simulations by learning from trajectory data to enable larger integration steps~\citep{xu2024equivariant,bowen2024generative}, typically in two regimes.

The first operates in the \emph{deterministic} setting, where the current state uniquely determines the future state. Similar to our approach, these methods aim to predict long integration steps for Hamiltonian dynamics~\citep{bigi2025flashmd,thiemann2025forcefree}, but they require reference trajectories, whose generation at \emph{ab-initio} accuracy is prohibitively expensive~\citep{thompson2025atom}. Using pre-trained \Glspl{MLFF} surrogates~\citep{bigi2025flashmd} alleviates this cost but remains computationally expensive and can introduce teacher-induced biases. In contrast, our method trains directly on \emph{ab-initio} force data without trajectories or surrogate teachers. Moreover, while prior approaches are usually trained for a fixed timestep~$\Delta t$, our model can be evaluated for any $\Delta t \in [0,\Delta t_{\mathrm{max}}]$, unifying instantaneous predictions and large-timestep propagation within a single model.

The second regime goes beyond the deterministic limit and reframes long-time propagation as a \emph{stochastic sampling problem} \citep{noe2019boltzmann,bioemu2025}, typically addressed with generative models~\citep{klein2023timewarp,schreiner2023ito,diez2025transferable,olsson2026generative}. As the timestep increases, systems increasingly behave like an equilibrium sampler, shifting emphasis from temporal evolution to matching stationary statistics. These methods target this equilibrium-sampling limit, but sacrifice fine-grained kinetic information and typically restrict dynamics to a fixed thermodynamic ensemble.

\textbf{Interpolation for Hamiltonian dynamics.}
Other methods exploit time-invertibility to interpolate between states in phase space and recover ground truth dynamics with high accuracy~\cite{winkler2022high,wang2023interpolating}.

\textbf{Similarities with few-step generative models.}
Our problem shares similarities with accelerated sampling in diffusion models~\citep[see also \cref{app:sec:few-step-diffusion-similarity}]{sohldickstein2015nonequilibrium, ho2020ddpm, song2021score}, where recent works employ few-step generation to bypass expensive ODE integration~\citep{song2023consistency, song2024improved, lu2025simplifying}. In contrast to those probabilistic methods, our training objective is entirely deterministic and purely motivated by physics without any noise injection.

Another key distinction lies in the training signal: Flow-based generative models interpolate between known source and target distributions over a fixed time interval $t \in [0,1]$. In physical simulations without trajectory data, neither the target samples nor the path are known, precluding trajectory-matching objectives~\citep{salimans2022progressive,zheng2023fast}. Instead, we train solely on instantaneous forces over an unbounded time horizon, adapting Mean Flow-style formulations~\citep{geng2025mean, geng2025improved, boffi2025build, sabour2025align} to Hamiltonian dynamics.

\textbf{Similarities with PINNs.}
Our objective can be viewed as minimizing the residual of the Liouville equation. From this perspective it resembles a PDE loss found in Physics-Informed Neural Networks (PINNs)~\cite{raissi2019pinns}.

In contrast to PINNs, our method uses self-distillation to learn an operator capable of advancing any phase-space state. Consequently, unlike standard PINNs that fit specific solutions to fixed boundary conditions, our framework dynamically bootstraps its own targets and trains exclusively on instantaneous labels.
\section{Method}
\label{sec:methods}
In this work, we present a framework that accelerates Hamiltonian dynamics by shifting the focus from \emph{faster force evaluation} to \emph{faster integration}. In the following, we introduce a trajectory-free training objective for a model capable of recovering instantaneous forces while supporting accurate predictions over large timesteps, reducing integration steps.

\subsection{Hamiltonian Flow Maps}
The core idea is to model the Hamiltonian evolution directly in phase space over a finite time interval. Concretely, we learn a \emph{Hamiltonian flow map} $\hmap_{\apos\to\bpos}$ that advances positions and momenta from time $\apos$ to $\bpos$ according to \cref{eq:forward_hamiltonian_map},
\begin{equation}
\begin{pmatrix}
\pos_\apos\\
\mom_\apos
\end{pmatrix}
\mapsto 
\begin{pmatrix}
\pos_\bpos\\
\mom_\bpos
\end{pmatrix}
= \hmap_{\apos\to\bpos}(\pos_\apos, \mom_\apos).
\end{equation}

A direct way to learn this mapping is to regress paired initial $(\pos_\apos,\mom_\apos)$ and final states $(\pos_\bpos,\mom_\bpos)$ from trajectories \citep{klein2023timewarp, thiemann2025forcefree, bigi2025flashmd}, or to unroll a numerical integrator during training \citep{zugec2025dynamic}. However, both strategies are inherently sequential, leading to high computational cost and limited scalability. Instead, we seek a formulation that avoids simulating intermediate states during training and learns large-step predictions directly from \emph{single-time phase-space samples}.

To achieve this, we build on recent advances in few-step generative modeling, in particular the \emph{Mean Flow} framework of \citet{geng2025mean} and the related self-distillation approach of \citet{boffi2025build}, which reduce the number of integration steps required during sampling in flow matching and diffusion models~\citep{lipman2023flow}. While conceptually related, we do not train a flow matching model. Instead, we adapt the underlying consistency identity to deterministic Hamiltonian phase-space dynamics and to a training regime in which only instantaneous force labels are available. In~\cref{app:sec:few-step-diffusion-similarity}, we discuss this in more detail.

\textbf{Mean displacement field.}
Motivated by this view, we define the startpoint-conditioned mean displacement field
\begin{equation}
\hdisp(\pos_\apos, \mom_\apos, \bpos - \apos) \coloneqq \frac{1}{\bpos - \apos}
\int_\apos^\bpos
\begin{pmatrix}
\vel_\tau \\
\force_\tau
\end{pmatrix}
\mathrm{d}\tau ,
\label{eq:displacement-def}
\end{equation}
which represents the time-averaged velocity and force accumulated along the trajectory over the interval $[\apos,\bpos]$. This quantity captures the non-trivial, integration-dependent component of the flow map in \cref{eq:forward_hamiltonian_map} that we seek to approximate with a neural network. Formulating the dynamics in terms of this mean displacement normalizes small and large time intervals to the same scale and ensures a well-defined limit as $\t=\bpos-\apos$ approaches 0, in which $\hdisp$ recovers the instantaneous velocity and force. In the following, we show how $\hdisp$ can be learned efficiently and subsequently used to reconstruct the Hamiltonian flow map $\hmap_{\apos\to\bpos}$.

\textbf{Trajectory-free consistency equation.}
The parameterization of the average displacement field $\hdisp$ suffers from the same computational drawbacks as the full flow map since it still involves an explicit time integral. Differentiating with respect to time yields an equivalent formulation that no longer involves an explicit integral. To this end, we multiply \cref{eq:displacement-def} by $(\bpos-\apos)$ and take the time derivative ${\mathrm{d}}/{\mathrm{d}\apos}$,
\begin{equation}
\frac{\mathrm{d}}{\mathrm{d}\apos}
\left[
(\bpos - \apos)\hdisp(\pos_\apos, \mom_\apos, \bpos-\apos)
\right]
=
\frac{\mathrm{d}}{\mathrm{d}\apos}
\int_\apos^\bpos
\begin{pmatrix}
\vel_\tau \\
\force_\tau
\end{pmatrix}
\mathrm{d}\tau.
\label{eq:derivative-both-sides}
\end{equation}
We use the product rule on the left, the fundamental theorem of calculus on the right, and multiply with $-1$ to get
\begin{equation}
\hdisp(\pos_\apos, \mom_\apos, \bpos-\apos)
-
(\bpos - \apos) \frac{\mathrm{d}}{\mathrm{d}\apos}
\hdisp(\pos_\apos, \mom_\apos, \bpos - \apos)
=
\begin{pmatrix}
\vel_\apos \\
\force_\apos
\end{pmatrix}.
\label{eq:mean-identity}
\end{equation}
The total derivative $\tfrac{\mathrm{d}}{\mathrm{d}\apos}\hdisp$ follows from the chain rule,
\begin{equation}
\begin{aligned}
\frac{\mathrm{d}}{\mathrm{d}\apos}\hdisp&(\pos_\apos, \mom_\apos, \bpos - \apos) =
\\
&= \frac{\mathrm{d}\pos_\apos}{\mathrm{d}\apos}\partial_{\pos_\apos}\hdisp
+
\frac{\mathrm{d}\mom_\apos}{\mathrm{d}\apos}\partial_{\mom_\apos}\hdisp
+
\frac{\mathrm{d} \bpos - \apos}{\mathrm{d}\apos}\partial_{\bpos - \apos}\hdisp \\
&=
(\vel_{\apos} \cdot \partial_{\pos_\apos})\hdisp
+
(\force_\apos \cdot \partial_{\mom_\apos})\hdisp
-
\partial_{\bpos - \apos} \hdisp ,
\end{aligned}
\label{eq:u-total-derivative}
\end{equation}
where we used Hamilton's equations to replace the time derivatives of position $\pos$ and momentum $\mom$ with velocity $\vel$ and force $\force$, respectively.

Together, \cref{eq:mean-identity,eq:u-total-derivative} define a necessary integral-free consistency condition that any mean displacement field must satisfy under Hamiltonian dynamics.

\subsection{Computational Approach}
\label{sec:computational-approach}

This consistency condition directly induces a regression objective for learning the displacement field. Concretely, this allows us to train a neural network $\hdisp^\params$ by minimizing
\begin{equation}
\begin{gathered}
\mathcal{L}(\params, \pos, \mom, \vel, \force) = \mathbb{E}_{\t} \left[ \left\| \hdisp^\params(\pos, \mom, \t) - \hdisp_{\text{tgt}}\right\|^2_2 \right] \\
\hdisp_{\text{tgt}} = \begin{pmatrix}\vel \\ \force \end{pmatrix}
+ \t\left[ (\vel \cdot \partial_{\pos})\hdisp^\params +(\force \cdot \partial_{\mom})\hdisp^\params - \partial_\t \hdisp^\params \right] ,
\end{gathered}
\label{eq:loss}
\end{equation}
where the time interval is denoted by $\t = \bpos - \apos$. 
For a system with $N$ particles and $d$ dimensions, the neural network is a map $\hdisp^\params: \mathbb{R}^{2dN+1}\to\mathbb{R}^{2dN}$, predicting mean forces and velocities $\mforce, \mvel \in \mathbb{R}^{dN}$ from positions $\pos \in \mathbb{R}^{dN}$, momenta $\mom \in \mathbb{R}^{dN}$ and time interval $\t \in [0,\t_\mathrm{max}]$.

In \autoref{app:sec:sufficiency}, we show that minimizing this loss is sufficient to learn Hamiltonian flow maps. Importantly, the objective uses only instantaneous labels and \emph{does not} require time integrals or any explicit simulation during training.

\begin{algorithm}[b!]
\caption{Efficiently Learning Hamiltonian Flow Maps}
\begin{algorithmic}[1]
\REQUIRE Neural network \( \hdisp^\params \), sample \( (\pos, \mom, \mom / \mass, \force) \),  ~~ distribution \( q(\tau) \), maximal timestep $\t_{\mathrm{max}}$
\STATE \( \t \gets \tau\cdot \t_{\mathrm{max}},\ \tau \sim q(\tau) \)
\STATE \( (\hdisp, \frac{\mathrm{d}}{\mathrm{d}t}\hdisp) \gets \texttt{jvp}(\hdisp^\params, (\pos, \mom, \t), (\mom/\mass, \force, -1)) \)
\STATE \( \hdisp_{\text{tgt}} \gets \texttt{stack}([\mom/\mass, \force]) + \t \cdot \frac{\mathrm{d}}{\mathrm{d}t}\hdisp \)
\STATE \( \mathcal{L}_\params \gets ||\hdisp - \texttt{stopgrad}(\hdisp_{\text{tgt}})||^2_2 \)
\end{algorithmic}
\label{alg:training}
\end{algorithm}

\textbf{Training data and sampling.} Our training objective requires phase-space samples $(\pos_\apos,\mom_\apos,\vel_\apos,\force_\apos)$ together with a randomly sampled time interval $\t$. Because we restrict to time-independent Hamiltonians, the phase-space state is fully determined by $(\pos,\mom)$ and is independent of absolute time $\apos$. Training can therefore use arbitrary instantaneous tuples $(\pos,\mom,\vel,\force)$ that do not need to come from trajectories. For the systems we study, we consider the common case of \emph{separable} Hamiltonians, so that $\vel=\mom / \mass$. In molecular systems, however, most \gls{MLFF} datasets provide only $(\pos,\force)$ and no $\mom$ labels. We therefore sample $\mom$ independently from Maxwell--Boltzmann distributions at varying temperatures, resulting in the training sample $(\pos,\mom,\mom/\mass,\force)$. Details on the sampling of $\t$ and $\mom$ are given in \cref{app:implementation}.

\textbf{Efficient implementation.} The loss in \cref{eq:loss} can be evaluated efficiently using Jacobian--vector products (\texttt{jvp}) available in standard deep learning frameworks. It requires only a single forward pass of the network and one backward pass. Optimizing the full objective, including gradients through the target $\hdisp_{\text{tgt}}$, can yield the best results \citep{you2025modular} but incurs substantial computational cost due to higher-order derivatives. In practice, we therefore adopt a \emph{stop-gradient} formulation, which treats the right-hand side as a fixed regression target and avoids these higher-order terms. This strategy is well established \citep{song2023consistency, song2024improved, geng2025mean, he2025rne} and reduces the training overhead to approximately $30\%$ compared to standard \gls{MLFF} training on a single GPU. The full training procedure is given in \cref{alg:training}.

\textbf{Force matching and consistency across time.}
In the case where $\t = 0$, the force component of the loss reduces to
\begin{equation}
    \mathcal{L}(\params, \pos,\mom,\vel,\force)_{dN+1:} = \left\| \hdisp^\params(\pos, \mom, \t=0)_{dN+1:} - \force \right\|^2_2,
\end{equation}
which recovers the standard loss of \glspl{MLFF}~\cite{unke2021machine} and therefore the network learns to predict the interatomic forces. 
For $\t > 0$, our loss encourages consistency between the predicted displacement field and the time-integrated instantaneous quantities.

Intuitively, this view amounts to a form of self-distillation across time: instantaneous force predictions act as a base signal, while predictions at larger $\t$ are constrained to match the accumulation of these local dynamics (see bottom row of~\autoref{fig:intro-figure}). \citet{zhang2025alphaflow} analyze these two components of the \textit{Mean Flow} loss in the generative modeling setting, which is useful for understanding our loss as well. Accurate learning at $\t = 0$ is essential, as it anchors the consistency condition for larger timesteps. Therefore, we sample $\t = 0$ in 75\% of training cases, balancing stable learning of instantaneous quantities with gradual enforcement of long-timescale consistency (see~\cref{app:subsec:force-error}).

\textbf{Recovering the Hamiltonian flow map.} \label{paragraph:time-reversal-of-inverse}
The proposed loss learns the \emph{average displacement} Hamiltonian flow $\hdisp$, which does not directly predict the next state. However, by construction, this field recovers the Hamiltonian flow map $\hmap_{\apos\to\bpos}$ (compare \cref{eq:forward_hamiltonian_map}) by multiplying with $\t$
\begin{equation}
\hmap_{\apos\to\bpos}(\pos_\apos, \mom_\apos) 
 =
 \begin{pmatrix}
    \pos_\apos \\\mom_\apos
\end{pmatrix}
 + \t \cdot \hdisp(\pos_\apos, \mom_\apos, \t).
\label{eq:flow-parameterization}
\end{equation}
Simulations can be obtained by repeating this update, effectively reducing the number of materialized states.

\textbf{Example: The 1D harmonic oscillator.}
\begin{figure}[t]
    \centering
    \includegraphics{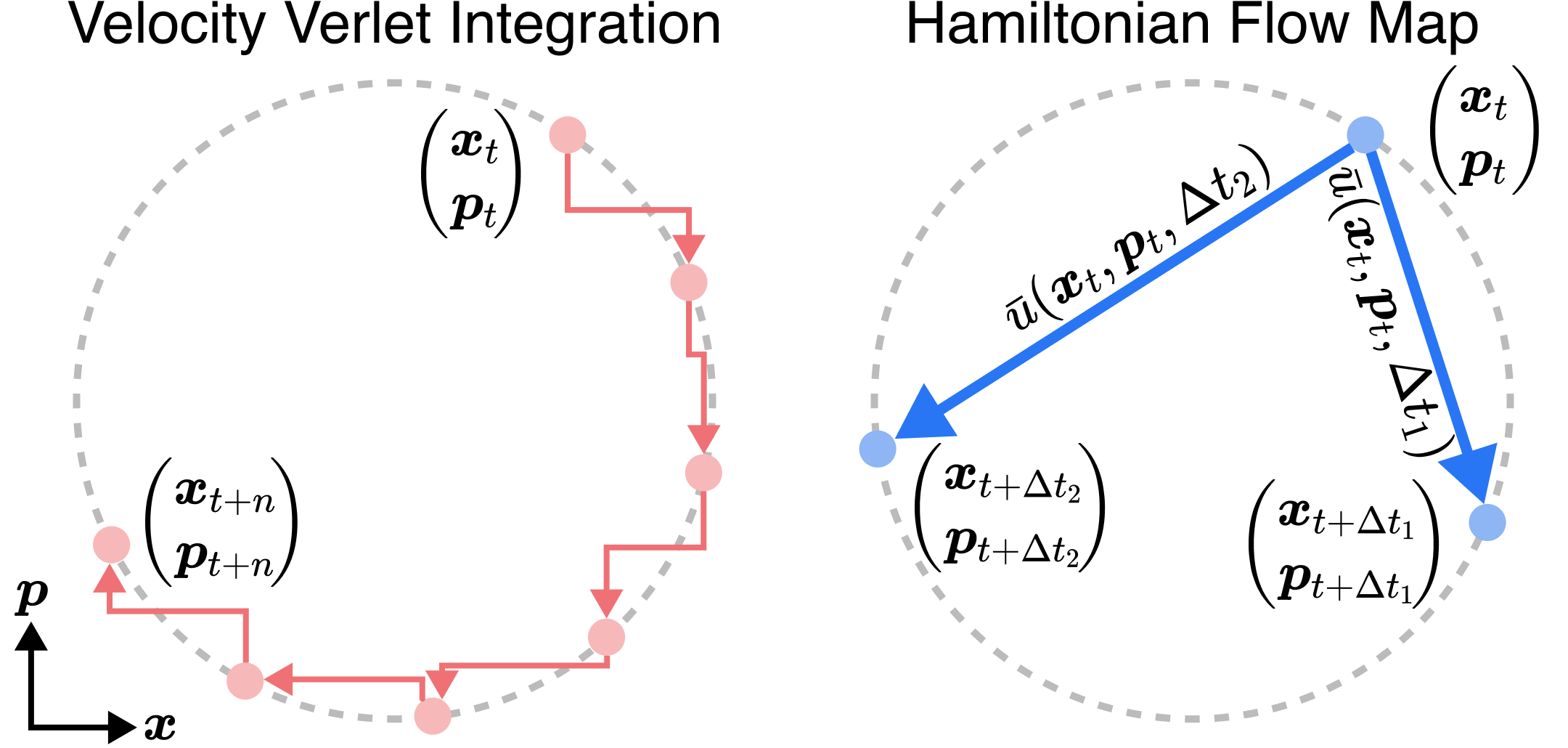}
    \caption{HFMs vs. classical integration: Symplectic integrators such as Velocity Verlet (VV) advance the system through many local half-steps (left). HFMs instead predict the phase-space displacement over the interval directly (right). By modeling the mean velocity and force $(\bar{\vel},\bar{\force})$ over the interval, HFMs apply a single large update (blue arrows) from the current state.}
    \vspace{-0.4cm}
    \label{fig:harmonic-oscillator}
\end{figure}
We illustrate the proposed method using an analytically tractable toy system: the one-dimensional harmonic oscillator, which models, e.g., a frictionless spring. The trajectories for this system are circular orbits in phase space. \cref{fig:harmonic-oscillator} schematically compares Velocity Verlet integration with our approach. 

In contrast to classical integrators, such as Velocity Verlet, which rely on local half-step corrections, Hamiltonian flow maps directly apply finite displacements in phase space. We define these displacements through mean velocities and mean forces, obtained by learning the averaged instantaneous vector field over a continuous time interval.

\subsection{Model Architecture and Inference Filters} \label{sec:architecture-filters}
\begin{figure}[t]
    \centering
    \includegraphics{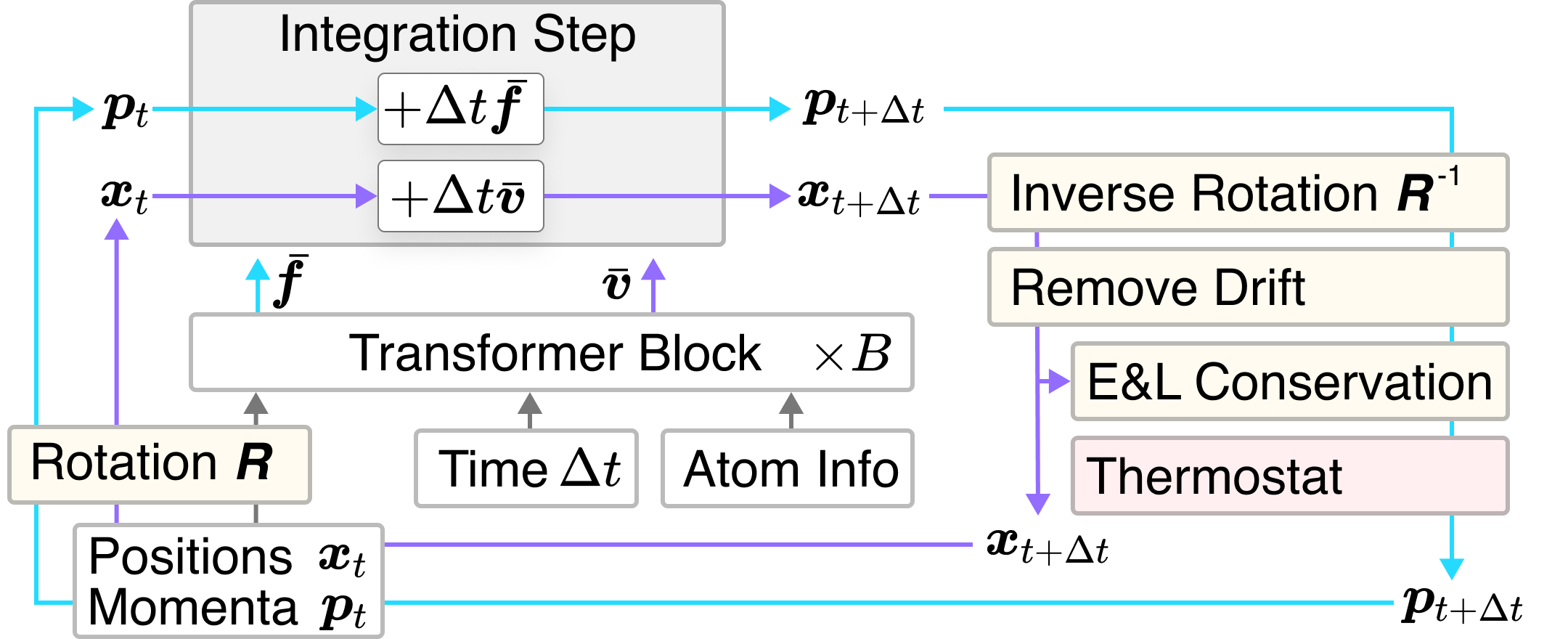}
    \caption{Simulation with a trained HFM. The model predicts mean forces $\mforce$ and mean velocities $\mvel$, conditioned on the current phase-space state $(\pos_t, \mom_t)$, the time interval $\t$, and atom information (types and masses). The predictions advance the system by one integration step. To ensure stable rollouts, we refine the updated state using three filters (right): removal of global translation drift, coupled conservation of energy and angular momentum (E\&L), and random rotation for approximate rotation equivariance.
    }
    \vspace{-0.42cm}
    \label{fig:architecture}
\end{figure}
We adopt a recent transformer for 3D geometries proposed by~\citet{frank2025sampling} to predict mean velocities and mean forces (see~\cref{app:sec:model-architecture}). Directly predicting these quantities yields non-conservative models with energy drift potentially causing instability~\citep{bigi2025dark}, which is critical in NVE simulations~\citep{thiemann2025forcefree,bigi2025learning} and can persist in NVT ensembles despite thermostats~\citep{bigi2025flashmd}. In our experiments, NVT rollouts remain stable even without inference filters, but still show some thermodynamic drift (see~\cref{app:subsec:filters-fix-stuff}).

To mitigate these problems, previous work has introduced inference filters to preserve physical constraints~\citep{bigi2025flashmd, thiemann2025forcefree}. However, conservation of total angular momentum and total energy have so far been looked at only separately. We solve a constrained optimization problem in closed form, to minimally adjust the momenta after each update step, such that both total energy and angular momentum are preserved (see~\cref{app:inference,app:subsec:coupled-conservation-filter}). Compared to sequential or decoupled correction schemes, this coupled formulation crucially avoids interference between the two constraints. To apply this filter, we need to evaluate the change in total energy that is caused by our model to correct it: for the potential energy part, we employ an additional prediction head in our model, or use a separately trained \gls{MLFF}. Note, that these filters do not correct fundamental integration errors and do not enable larger timesteps on their own (see~\cref{app:subsec:filters-fix-stuff}).

 In summary, we employ three lightweight filters during simulation (see~\cref{fig:architecture}):
\begin{itemize}[topsep=0pt, partopsep=0pt, itemsep=0pt, parsep=0pt,leftmargin=10pt]
\item \textbf{Random rotation.} Reduce effects of non-equivariance by applying a random global rotation before each step.
\item \textbf{Remove drift.} Remove total momentum of the system to prevent flying-ice-cube instabilities \citep{harvey1998flying}. 
\item \textbf{Energy and angular momentum conservation.} Enforce coupled conservation of total energy and angular momentum by modifying and rescaling the momenta.
\end{itemize}

\section{Experiments}
\label{sec:experiments}
In this section, we train Hamiltonian Flow Maps (HFMs) with our proposed approach and evaluate them. Our code and model weights are available at:\\ \ourCode{}.

\subsection{Classical Mechanics}
We first consider three systems described by classical mechanics, including two simple single-particle problems and a more complex gravitational 100-body system.

\textbf{Single particle systems.}
We begin with two classic Hamiltonian systems, the Barbanis potential and the spring pendulum, and stress-test integration stability at large timesteps. For simplicity, we use a dimensionless coordinate system. For each system, we train HFMs on temporally uncorrelated phase-space samples by uniformly sampling positions and momenta at fixed total energy and computing forces from the ground-truth potential (see \cref{app:sec:experimental-setup-single-particle} for details).

\begin{figure}[tb]
    \centering
    \includegraphics{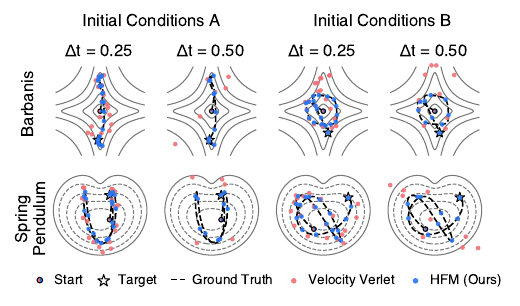}
    \caption{Generated trajectories for two potentials (Barbanis, spring pendulum) from two initial conditions (left/right). We compare the ground truth (dashed), VV (red), and HFM (blue) at two large timesteps. VV deviates even with a small timestep increase, while HFM stays aligned at larger steps.
    }
    \vspace{-0.45cm}
    \label{fig:classic}
\end{figure}

Once trained, we simulate dynamics either with Velocity Verlet (VV) using the ground-truth potential, or with the learned HFM at increasing timestep $\t$ (\cref{fig:classic}). Even at $\t = 0.25$, VV no longer matches the reference trajectory and exhibits phase and amplitude errors as $\t$ increases. In contrast, the learned HFM remains stable for larger $\t$ and closely follows the reference dynamics. This demonstrates that HFMs trained with our objective can take timesteps well beyond the regime in which VV is accurate.

\textbf{Gravitational $N$-body system.}
To consider a more complex setting, we use the 100-particle gravitational dataset from~\citet{brandstettergeometric}. Although the dataset contains trajectories, we ignore temporal ordering and train only on individual force labels, sampling momenta independently for each configuration. We again omit units for simplicity.

\begin{figure}[t]
    \centering
    \includegraphics{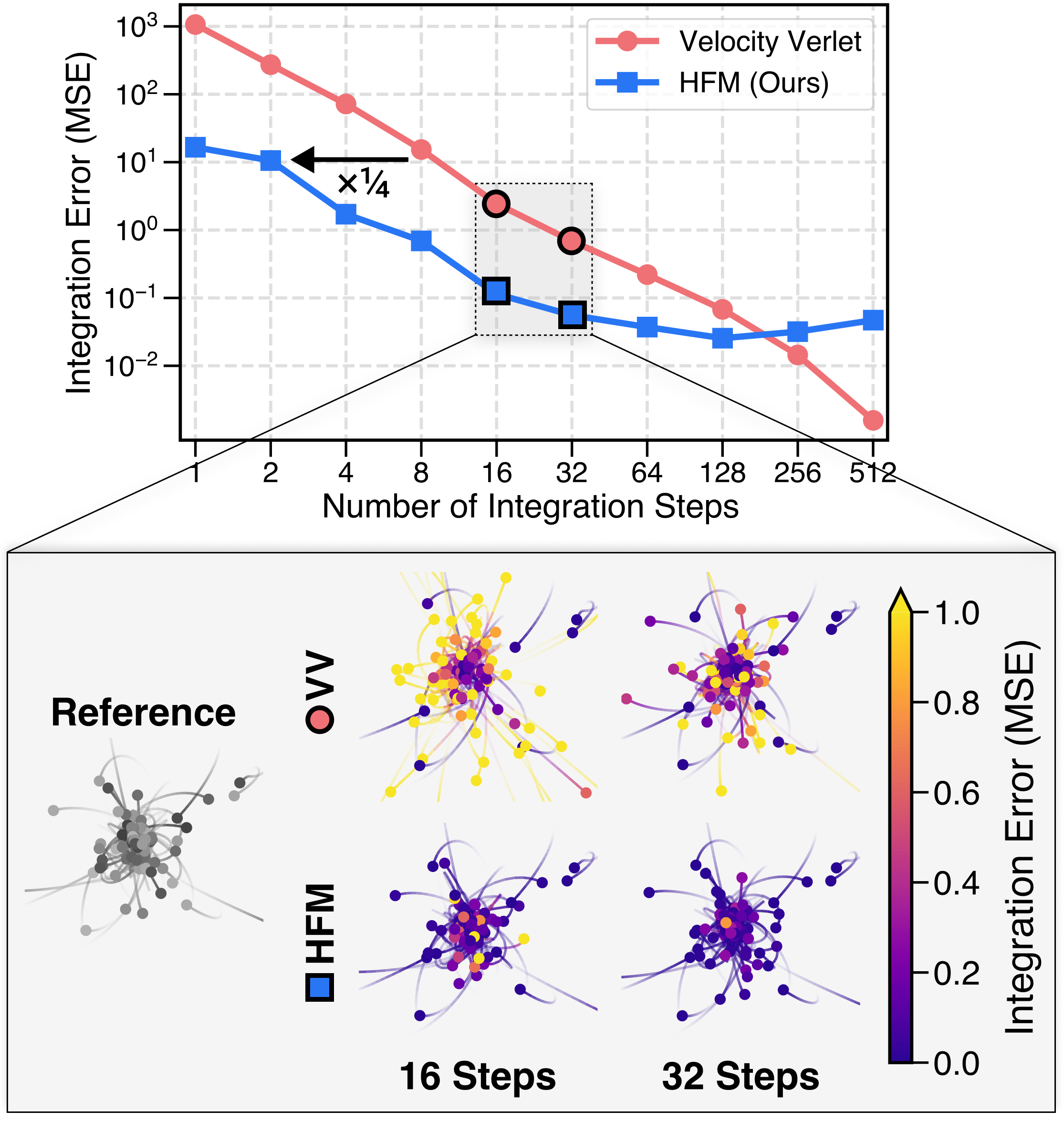}
    \caption{
    Gravitational $N$-body rollout ($\apos=3$, $\bpos=4$). 
    \textbf{Top:} MSE as a function of the number of integration steps $n$, averaged over the test set. The HFM remains accurate at coarse discretization, while VV diverges rapidly. Overall, the HFM requires $4\times$ fewer steps than VV to reach the same accuracy.
    \textbf{Bottom:} Particle trajectories for 16 and 32 integration steps, colored by per-particle deviation from the ground truth. The HFM yields physically consistent rollouts, whereas VV leads to unphysical scattering.
    }
    \vspace{-0.4cm}
    \label{fig:gravity}
\end{figure}

Following the protocol in~\citet{brandstettergeometric}, we assess extrapolation over the interval from $\apos=3$ to $\bpos=4$ (details in \cref{app:sec:experimental-setup-gravity}). We vary the number of integration steps $n$ used to bridge this interval, which sets the step size to $\t = (\bpos-\apos)/n$. We compare HFMs to VV integration of the ground-truth analytical potential in \cref{fig:gravity}. HFMs significantly outperform VV at large step sizes (small $n$), where the classical integrator becomes unstable. As $\t$ decreases, the task approaches instantaneous force prediction, and performance saturates due to inherent model error. 

\subsection{Molecular Dynamics}
In the following, we show that our framework enables long-time \gls{MD} simulations with fewer integration steps while faithfully reproducing the target equilibrium statistics. 

\begin{table*}[b]
\centering
\vspace{-0.2cm}
\caption{Interatomic distances $h(r)$ MAE [unitless] for 300\,ps of MD simulation in the NVT ensemble using a Langevin thermostat w.r.t. the reference data from \emph{ab-initio} calculations. For each system, we compare a single HFM model with $\t_{\max} = 10$\,fs at increasing timesteps $\t \in \{0.5,1,3,5,7,9\}$\,fs against an \gls{MLFF} baseline with $\t = 0.5$\,fs. We run the simulation for 5 different initial conditions and report the average performance with standard deviations shown in parentheses. Lower values indicate better performance.}
\vspace{-0.1cm}
\label{tab:fane_langevin}
\begin{tabular}{l|c|cccccc}
\hline
Dataset & MLFF & \multicolumn{6}{c}{Hamiltonian Flow Map} \\
 & 0.5\,fs & 0.5\,fs & 1\,fs & 3\,fs & 5\,fs & 7\,fs & 9\,fs \\
\hline
Aspirin & 0.030 \scriptsize{(0.002)} & 0.026 \scriptsize{(0.001)} & 0.027 \scriptsize{(0.001)} & 0.035 \scriptsize{(0.001)} & 0.038 \scriptsize{(0.001)} & 0.042 \scriptsize{(0.001)} & 0.046 \scriptsize{(0.001)} \\
Ethanol & 0.073 \scriptsize{(0.001)} & 0.075 \scriptsize{(0.001)} & 0.076 \scriptsize{(0.002)} & 0.083 \scriptsize{(0.001)} & 0.093 \scriptsize{(0.001)} & 0.089 \scriptsize{(0.001)} & 0.115 \scriptsize{(0.001)} \\
Naphthalene & 0.043 \scriptsize{(0.000)} & 0.044 \scriptsize{(0.000)} & 0.047 \scriptsize{(0.000)} & 0.053 \scriptsize{(0.000)} & 0.051 \scriptsize{(0.000)} & 0.052 \scriptsize{(0.000)} & 0.063 \scriptsize{(0.000)} \\
Salicylic Acid & 0.035 \scriptsize{(0.000)} & 0.035 \scriptsize{(0.000)} & 0.039 \scriptsize{(0.000)} & 0.042 \scriptsize{(0.000)} & 0.051 \scriptsize{(0.001)} & 0.052 \scriptsize{(0.000)} & 0.058 \scriptsize{(0.003)} \\
\hline
\end{tabular}
\vspace{-0.3cm}
\end{table*}

We train and analyze HFM models on a selection of the MD17 and MD22 datasets~\cite{chmiela2017machine,chmiela2023accurate}, alanine dipeptide in implicit solvent~\citep{kohler2021smooth} and coarse-grained proteins~\cite{deshaw20211fastfolding}. We use experiment-dependent training splits with details in \cref{app:sec:experimental-setup-small-mols,app:sec:experimental-setup-paracetamol,app:sec:experimental-setup-alanine}. As baselines, we train SO(3)-equivariant \glspl{MLFF} and use the force outputs for numerical integration within VV with $\t = 0.5$\,fs. All MLFF models use the same data splits as the HFM models.

\begin{figure}[tb]
    \centering
    \includegraphics{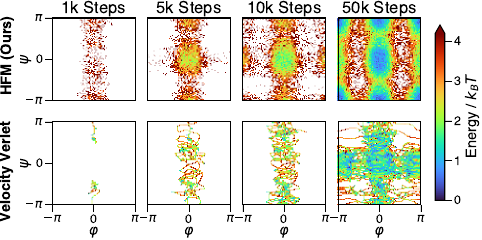}
    \vspace{-0.12cm}
    \caption{State space exploration in an NVE simulation of the paracetamol molecule for an increasing number of integration steps. \textbf{Top}: Our HFM with $\t = 9$\,fs. \textbf{Bottom}: MLFF + VV integrator with $\t=0.5$\,fs. Since our HFM allows for a larger integration timestep, it covers the state space much faster.}
    \vspace{-0.53cm}
    \label{fig:paracetamol_nve_nstep}
\end{figure}

\textbf{NVE simulations.} Before performing realistic constant temperature NVT simulations, we analyze our learned HFM in the NVE setting. This poses an absolute stress test for physical validity, as thermostats in an NVT can absorb numerical integration errors. Choosing the right timestep $\t$ is a trade-off between faster state space exploration and simulation accuracy, which can be addressed with our learned HFM. To test this, we compare the Ramachandran plots of the dihedral angles in paracetamol for HFM ($\t=9$\,fs) and the VV baseline given a fixed budget of integration steps. We find that the learned HFM explores the conformational space significantly faster, discovering relevant modes and uncorrelated samples more efficiently than the baseline (\cref{fig:paracetamol_nve_nstep}). Additional ablations testing energy and angular momentum conservation can be found in \cref{app:subsec:nve-E-L-conservation}.

\textbf{Structural observables.} We then transition to the realistic setting of NVT simulations and perform 300\,ps \gls{MD} simulations for four organic molecules using a Langevin thermostat. To asses the structural accuracy and stability of our simulations, we follow \citet{fu2023forces} and calculate the distribution $h(r)$ over interatomic distances $r$ and report the MAE w.r.t.\ \emph{ab-initio} reference data. Our learned HFM produces accurate simulation statistics for all test molecules (see ~\cref{tab:fane_langevin}). At $\t=0.5$\,fs, it achieves distance distribution accuracies comparable to the \gls{MLFF} baseline. As expected, the error increases in $\t$, but the HFM model maintains decent accuracy and stability up to $\t=9$\,fs for most molecules. Our results demonstrate robustness close to the training horizon $\t_{\mathrm{max}}=10$\,fs and well beyond standard integration limits. \cref{app:subsec:wall-clock-time,app:subsec:force-error,app:subsec:simulation-stability} contain further metrics for wall clock time comparison, force MAE and simulation stability. We further ran simulations with other thermostats (\cref{app:subsec:other-thermostats}), where we observe that a global and deterministic Nos\'{e}-Hoover thermostat~\cite{martyna1996explicit} can lead to simulation artifacts. Similar effects have been observed in other long-timestep models~\citep{bigi2025flashmd} and have been attributed to failure of the thermostat to maintain kinetic energy equipartition across atom types.

\begin{figure*}[tb]
    \centering
    \includegraphics{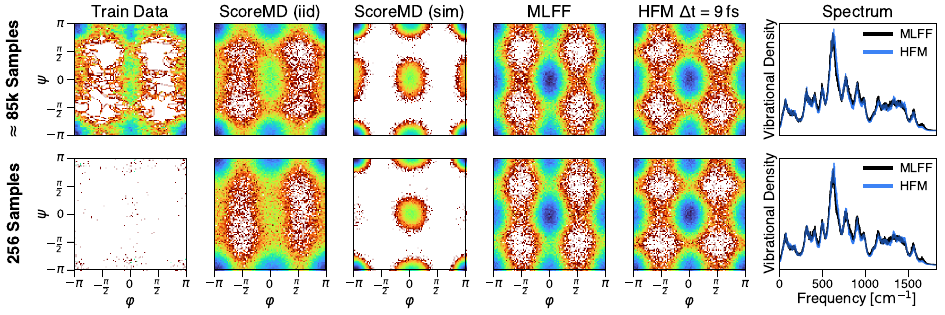}
    \caption{Data efficiency for Paracetamol in NVT simulation (3\,ns). We compare models trained on $\approx 85$k samples (\textbf{top}) and on only 256 samples (\textbf{bottom}). We show the free energy surface projected onto the dihedral angles $\varphi, \psi$. Both the \gls{MLFF} and the HFM recover the reference distribution despite the biased training distribution. In contrast, ScoreMD, like many other generative approaches, assumes converged training samples and therefore reproduces the biased training distribution in both settings. The HFM additionally yields accurate large-timestep dynamics, as indicated by a vibrational spectrum that closely matches the small-timestep \gls{MLFF}.}
    \vspace{-0.4cm}
    \label{fig:paracetamol_nvt_training_data}
\end{figure*}
\textbf{Temporal observables.} Although distributions are well suited to validate structural statistics, they do not probe temporal dynamics. To do so, we perform an NVT simulation of paracetamol and calculate the power spectrum from the atomic velocities (right panel of~\cref{fig:paracetamol_nvt_training_data}). Close alignment between HFM and the MLFF baseline indicates that our model describes the underlying dynamics with high fidelity.

\textbf{Training data distribution.} Next, we train two HFMs on sparse (256 samples) and saturated (85k samples) paracetamol data (bottom and top row in \cref{fig:paracetamol_nvt_training_data}). Within NVT simulations, both HFMs accurately describe the conformational space (Ramachandran plots) and the temporal dynamics (Power spectra) compared to the MLFF baseline, demonstrating that our HFMs capture the essential flow topology even from sparse data and potentially biased distributions. 

In contrast, many generative approaches assume specific training data, e.g., samples from a converged \gls{MD} simulation. To highlight this, we also compare our HFM with ScoreMD~\citep{plainer2025consistent}, a recent diffusion-based approach that can generate independent samples (iid) by denoising but can also be used for simulation (sim). On the respective splits, it reproduces the training distribution rather than yielding converged distributions (\cref{fig:paracetamol_nvt_training_data}), which in practice requires computationally expensive reweighting~\citep{klein2024tbg}. Further, many generative approaches cannot recover dynamics such as spectra.

\begin{figure*}[tb]
    \centering
    \includegraphics{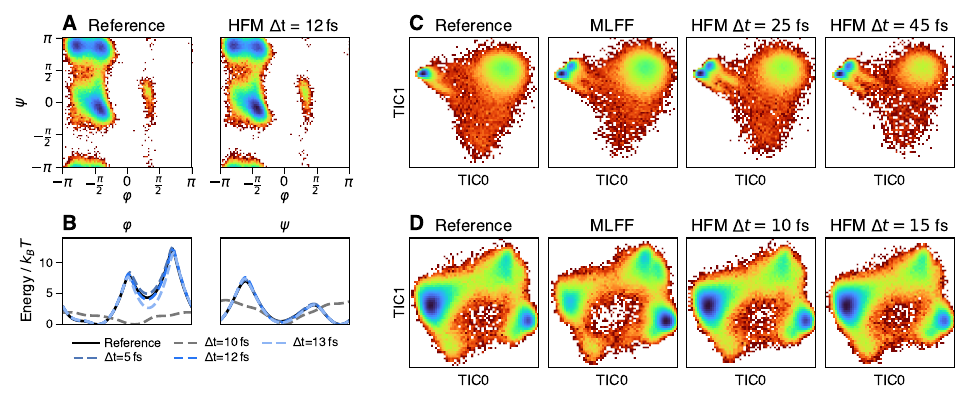}
    \caption{Long-running NVT simulations. 
\textbf{A}~Ramachandran plot of the backbone dihedral angles $\phi$ and $\psi$ for alanine dipeptide, comparing the reference simulation to the HFM evaluated at $\t = 12$\,fs. 
\textbf{B}~Free energy projections for alanine dipeptide across different step sizes $\t$ compared to the reference distribution. Across most step sizes, the HFM correctly samples the metastable basins; as $\t$ increases, the systematic deviation grows. 
\textbf{C}~Time-lagged independent component analysis (TICA) projections for Chignolin simulations ($10$\,ns, 10 replicas) comparing the all-atom reference simulation, rollouts using a CG MLFF and HFM trained on CG data. The HFM accurately captures the conformational landscape even at large time steps up to $\t = 45$\,fs compared to the reference MLFF simulation ($\t = 0.5$\,fs). 
\textbf{D}~TICA projections for BBA simulations ($10$\,ns, 10 replicas). HFM converges significantly faster with $\t = 10/15$\,fs.}
    \vspace{-0.4cm}
    \label{fig:aldp}
\end{figure*}

\textbf{Long, stable peptide dynamics.}
To demonstrate scalability to systems with complex metastable states, we simulate alanine dipeptide.
Parts A and B of~\cref{fig:aldp} show the Ramachandran plot and marginal distributions of the backbone dihedral angles $(\varphi, \psi)$. We generate a total trajectory length of 1\,$\mu$s by running 10 parallel simulations of 100\,ns each with the HFM at varying timesteps. The learned integrator captures transitions between metastable states and reproduces the characteristic free energy landscape.

\textbf{Stepsize.} For alanine dipeptide, we train the HFM with $\t_{\mathrm{max}}=15$\,fs and obtain accurate free energy surfaces with $\t=12$\,fs. This is practically relevant because standard \gls{MD} simulations often require additional stabilization beyond $\t=4$\,fs \citep{hopkins2015long}, with $\t\approx 6$\,fs being achievable in principle \citep{izaguirre1999longer}, which is still only about half of the step size we use here. Such regularization techniques may also further improve stability for our method, but are beyond the scope of this work.

Across experiments and hyperparameter choices, we observe that long rollouts around $\t \approx 10$\,fs can become unstable for alanine dipeptide. This instability does not consistently correlate with one-step prediction error, suggesting that it arises from the interaction between the learned map and the simulation setup rather than from degraded local accuracy. We report free energy surfaces and compute errors for a wider range of timesteps in \cref{app:subsec:ablations-aldp}.

\begin{figure}[tb]
    \centering
    \includegraphics{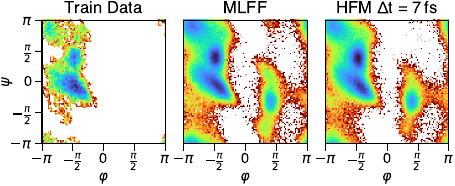}
    \caption{Ac-Ala3-NHMe NVT simulations. When trained on biased data where a mode is completely missing (left), our HFM recovers the hidden mode ($3\text{ns}$ rollouts, $10$ replicas) and stays comparable to an MLFF reference (right).}
    \vspace{-0.5cm}
    \label{fig:ala3_ramachandran_comparison}
\end{figure}

\textbf{Inference-time Extrapolation.} We test extrapolation with HFMs beyond incomplete data by training on MD22 Ac-Ala3-NHMe~\cite{chmiela2023accurate}, where a conformational mode is completely missing. The HFM extrapolates during inference to recover the missing mode, producing the same free energy landscape like an MLFF (\autoref{fig:ala3_ramachandran_comparison}).

\textbf{Coarse-grained systems.} To evaluate robustness on larger macromolecules, we study our model on coarse-grained (CG) representations of Chignolin and BBA, retaining only the $C_\alpha$ atoms. Instead of direct ab initio data, we generate force labels by querying ScoreMD~\citep{plainer2025consistent}, a pre-trained coarse-grained machine learning force field for fast folding proteins \citep{deshaw20211fastfolding}. We distill this model into large-timestep dynamics using our HFM framework. To account for any potential model bias, we additionally train an MLFF for comparison on the same CG data. As shown in C and D of \cref{fig:aldp}, HFM maintains stable simulations and captures the correct ensemble statistics across complex landscapes and large timesteps.

For Chignolin, all of our models overrepresent a second folded state. However, since this mode is also present in the MLFF, we attribute this to the distillation of ScoreMD, where errors can easily accumulate. Our model remains highly stable and accurate at a step size of $\t = 45$\,fs ($\t_{\max} = 50$\,fs), exceeding a speedup of $20\times$ compared to the underlying reference force field while recovering correct transition rates (see \cref{app:subsec:transition-rates-chignolin}).

For BBA, our HFM rollouts stay closer to the reference distribution and better recover the respective weights between states. Overall, we found our HFM model to be more stable during inference (compare \cref{app:subsec:bba-individual-rollouts}). We show stable rollouts up to $\t = 15$\,fs steps ($\t_{\max} = 20$\,fs).

\section{Conclusion and Limitations}
A central challenge in computational physics is the need for small integration steps to ensure numerical stability. This constraint often prevents simulations from reaching the timescales of interest and has motivated data-driven approaches to accelerate time integration. Existing methods, however, rely on expensive pre-computed trajectories. 

We address this limitation by introducing a trajectory-free training objective for Hamiltonian Flow Maps (HFMs) from instantaneous supervision (forces and positions). Across systems from single-particle dynamics to multi-body molecular systems, we show that HFMs enable stable large-timestep simulation beyond the limits of classical integrators.

The proposed maps approximate the underlying dynamics and therefore only approximately preserve physical structure such as symplecticity, energy conservation, or equivariance. In practice, we combine the model with inference filters and local stochastic thermostats to improve long-rollout stability. Furthermore, we observe that optimizing the training objective becomes increasingly challenging as the timestep grows, particularly in strongly chaotic regimes.

Overall, we show that useful dynamical maps can be learned without access to trajectories. Our HFMs are trained on the same data as classical \glspl{MLFF} while maintaining similar training and inference costs. Beyond this, they combine learning instantaneous forces and large-timestep dynamics simultaneously. We hope this perspective broadens the set of practical tools for long-time simulation, encouraging practitioners to augment standard \glspl{MLFF} with HFMs to obtain stable, large-timestep simulations.

\clearpage
\section*{Impact Statement}
This work develops a machine learning framework for accelerating Hamiltonian dynamics simulations, with a focus on molecular dynamics. By enabling larger integration timesteps without requiring trajectory data during training, the proposed method can substantially reduce the computational cost of atomistic simulations. This may lower barriers to studying long-timescale processes in chemistry, materials science, and biophysics, and support applications such as drug design. We do not anticipate direct negative societal impacts; the method is intended to complement, not replace, first-principles simulations, and does not introduce ethical concerns beyond those common in computational modeling in the physical sciences.

\section*{Acknowledgements}
We would like to thank Klara Bonneau, Jonas Köhler, Hartmut Maennel, Tim Ebert, Khaled Kahouli and Martin Michajlow for fruitful discussions and their helpful input. JTF, WR, KRM, and SC acknowledge support by the German Federal Ministry of Research, Technology and Space (BMFTR) under Grants BIFOLD24B, BIFOLD25B, 01IS18037A, 01IS18025A, and 01IS24087C. MP is supported by the Konrad Zuse School of Excellence in Learning and Intelligent Systems (\href{https://eliza.school/}{ELIZA}) through the DAAD programme Konrad Zuse Schools of Excellence in Artificial Intelligence, sponsored by the Federal Ministry of Education and Research. Further, this work was in part supported by the BMBF under Grants 01IS14013A-E, 01GQ1115, 01GQ0850, 01IS18025A, 031L0207D, and 01IS18037A. KRM was partly supported by the Institute of Information \& Communications Technology Planning \& Evaluation (IITP) grants funded by the Korea government (MSIT) (No.2019-0-00079, Artificial Intelligence Graduate School Program, Korea University and No. 2022-0-00984, Development of Artificial Intelligence Technology for Personalized Plug-and-Play Explanation and Verification of Explanation). Moreover, we gratefully acknowledge support by the Deutsche Forschungsgemeinschaft (SFB1114, Projects No. A04 and No. B08) and the Berlin Mathematics center MATH+ (AA1-10, AA2-20, and AA-Health). 




\clearpage
\bibliography{bibliography}
\bibliographystyle{icml2026}

\newpage
{
\appendix
\onecolumn
\crefalias{section}{appendix}
\crefalias{subsection}{appendix}
\section{Notation} \label{app:notation}

For better readability, we will use $\square_i$ instead of $\square^{(i)}$ to index particles throughout the appendix. For the state $(\pos_t, \force_t, \mom_t)$ of an $N$-particle system at time $t$, with $\pos_t, \force_t, \mom_t \in \mathbb{R}^{dN}$, we write $\pos_{t,i}, \force_{t,i}, \mom_{t,i} \in \mathbb{R}^d$ for the position, force, and momentum of particle $i$, respectively. In contrast, for an independent sample $(\pos, \force, \mom)$, with $\pos, \force, \mom \in \mathbb{R}^{dN}$, we write $\pos_i, \force_i, \mom_i \in \mathbb{R}^d$. To avoid ambiguity, we reserve $i, j$ exclusively for particle indices. All other single symbol subscripts denote time labels.
\begin{table}[ht]
   \centering
   \begin{tabular}{lcc}
   \toprule
   \textbf{Description} & \textbf{Symbol} & \textbf{Identities} \\
   \midrule
   Hamiltonian & $\hamiltonian(\pos,\mom)$ & \\
   Position & $\pos$ & \\
   Momentum & $\mom$ & $\mom=\mass\vel$ \\
   Velocity & $\vel$ & $\boldsymbol{v}=\dot{\boldsymbol{x}}$\\
   Force & $\force$ & $\force = \mass\acc = \dot{\mom}$\\
   Mass & $\mass$ & \\
   Timestep & $\t$ & \\
   Start point & $\apos$ & $\bpos - \apos$ \\
   End point & $\bpos$ & \\
   Time derivative & $\dot{\square}$ & $\frac{\mathrm{d}}{\mathrm{d}t} \square$\\
    \bottomrule
    \end{tabular}
    \caption{Notation and symbols used in the paper.}
\end{table}
\section{Implementation Details} \label{app:implementation}

In this section, we will discuss additional implementation details for our approach.
\subsection{Training Hamiltonian Flow Maps}

\paragraph{Training objective.} We implement the loss in \cref{eq:loss} for a sample $(\pos, \mom, \vel, \force)$ and timestep $\t$ as a weighted sum:
\begin{align}
    \mathcal{L}(\params, \pos, \mom, \vel, \force, \t) = \lambda_\vel \mathcal{L}_\vel(\params, \pos, \mom, \vel, \force, \t) + \lambda_\force \mathcal{L}_\force(\params, \pos, \mom, \vel, \force, \t).
\end{align}
The individual loss terms, i.e., the velocity loss term $\mathcal{L}_\vel(\params, \pos, \mom, \force, \vel)$ and force loss term $\mathcal{L}_\force(\params, \pos, \mom, \force, \vel)$, are defined as
\begin{align}
    \mathcal{L}_\vel(\params, \pos, \mom, \force, \vel, \t) &= w_\vel \cdot \dfrac{1}{3N} \sum_{i=1}^{N} \mass_i \cdot \left\| \bar{\vel}^\params(\pos, \mom, \t)_i - \mathrm{sg}(\bar{\vel}_{\text{tgt}})_i\right\|^2_2, \label{app:eq:vel-loss}\\
    \mathcal{L}_\force(\params, \pos, \mom, \force, \vel, \t) &= w_\force \cdot \dfrac{1}{3N} \sum_{i=1}^{N} \left\| \bar{\force}^\params(\pos, \mom, \t)_i - \mathrm{sg}(\bar{\force}_{\text{tgt}})_i\right\|^2_2,\label{app:eq:force-loss}
\end{align}
where $N$ denotes the number of particles in the system, $\bar{\vel}^\params(\pos, \mom, \t)_i, \bar{\force}^\params(\pos, \mom, \t)_i \in \mathbb{R}^d$ are the predicted mean velocities and forces for the $i$-th particle, $\mathrm{sg}(\bar{\vel}_{\text{tgt}})_i, \mathrm{sg}(\bar{\force}_{\text{tgt}})_i \in \mathbb{R}^d$ are the regression targets, and $\mathrm{sg}(\cdot)$ is a stop-gradient operator (see \cref{sec:methods}). In general, the targets $\bar{\vel}_{\text{tgt}}, \bar{\force}_{\text{tgt}} \in \mathbb{R}^{dN}$ are defined as
\begin{align}
    \bar{\vel}_{\text{tgt}} &= \vel + \t\left[ (\vel \cdot \partial_{\pos})\bar{\vel}^\params +(\force \cdot \partial_{\mom})\bar{\vel}^\params - \partial_\t \bar{\vel}^\params \right],\\
    \bar{\force}_{\text{tgt}} &= \force + \t\left[ (\vel \cdot \partial_{\pos})\bar{\force}^\params +(\force \cdot \partial_{\mom})\bar{\force}^\params - \partial_\t \bar{\force}^\params \right],
\end{align}
and can be efficiently computed using Jacobian–vector products (\texttt{jvp}) available in \texttt{jax}~\citep{bradburry2018jax} with minimal computational overhead (see \cref{sec:methods}). The scalars $w_\vel, w_\force \in \mathbb{R}$ in \cref{app:eq:vel-loss,app:eq:force-loss} are adaptive loss weights~\citep{geng2024consistency} with
\begin{align}
    w_\vel &= \left( \mathrm{sg}\left( \dfrac{1}{3N} \sum_{i=1}^{N}  \mass_i \cdot \left\| \bar{\vel}^\params(\pos, \mom, \t)_i - \mathrm{sg}(\bar{\vel}_{\text{tgt}})_i \right\|^2_2 \right) + c \right)^{-p},\\
    w_\force &= \left( \mathrm{sg} \left( \dfrac{1}{3N} \sum_{i=1}^{N} \left\| \bar{\force}^\params(\pos, \mom, \t)_i - \mathrm{sg}(\bar{\force}_{\text{tgt}})_i \right\|^2_2 \right) + c \right)^{-p}.
\end{align}
In our experiments, we set $c=10^{-3}$ and $p=0.5$, yielding a weighting similar to the Pseudo-Huber loss~\citep{song2024improved}, which we find to improve the training stability.
Additionally, in \cref{app:eq:vel-loss}, we weight the contribution of the $i$-th particle to the velocity loss by its mass $\mass_i$. We apply mass-weighted loss scaling, rather than predicting mass-scaled quantities~\citep{zheng2021learning, bigi2025flashmd}, as doing so would violate the underlying physics in our loss formulation. The networks $\bar{\vel}^\params$ and $\bar{\force}^\params$ share the same backbone. We discuss the architectures used throughout this work in \cref{app:sec:model-architecture}.

\paragraph{Rotation augmentation.}
In this work, we learn Hamiltonian flow maps using network architectures that do not transform equivariantly under rotation of the input positions. As a consequence, we apply random rotations during training to enforce approximate rotational equivariance via data augmentation~\citep{abramson2024alphafold3,frank2025sampling,plainer2025consistent}. Specifically, we randomly sample rotation matrices $\boldsymbol{R} \in \mathbb{R}^{3\times 3}$ (orthogonal matrices with determinant +1) and apply them as
\begin{equation} \label{app:eq:rotation-augementation}
    \mathrm{ApplyRotation}(\boldsymbol{R}, \pos)_i = \boldsymbol{R}\pos_i, \quad
    \mathrm{ApplyRotation}(\boldsymbol{R}, \force)_i = \boldsymbol{R}\force_i, \quad
    \mathrm{ApplyRotation}(\boldsymbol{R}, \mom)_i = \boldsymbol{R}\mom_i, 
\end{equation}
where $\pos_i, \force_i, \mom_i \in \mathbb{R}^3$ denote the position, force and momentum of the $i$-th particle, respectively. Based on \cref{app:eq:rotation-augementation}, we can adapt the original training algorithm in \cref{alg:training} with minimal changes to allow for training with rotation augmentation. The modified training algorithm is summarized in \cref{app:alg:training-with-augmentation}.

\begin{algorithm}
\caption{Efficiently Learning Hamiltonian Flow Maps with Rotation Augmentation}
\begin{algorithmic}[1]
\REQUIRE Neural network \( \hdisp^\params \), training sample \( (\pos, \mom, \mom / \mass,  \force) \), distribution \( q(\tau) \), maximal timestep $\t_{\mathrm{max}}$
\STATE \( \tau \sim q(\tau) \)
\STATE \( \boldsymbol{R} \sim \mathrm{SO}(3) \)
\STATE \( \t \gets \tau\cdot \t_{\mathrm{max}} \)
\STATE \( \pos \gets \mathrm{ApplyRotation}(\boldsymbol{R}, \pos) \)
\STATE \( \force \gets \mathrm{ApplyRotation}(\boldsymbol{R}, \force) \)
\STATE \( \mom \gets \mathrm{ApplyRotation}(\boldsymbol{R}, \mom) \)
\STATE \( (\hdisp, \frac{\mathrm{d}}{\mathrm{d}t}\hdisp) \gets \texttt{jvp}(\hdisp^\params, (\pos, \mom, \t), (\mom/\mass, \force, -1)) \)
\STATE \( \hdisp_{\text{tgt}} \gets \texttt{stack}([\mom/\mass, \force]) + \t \cdot \frac{\mathrm{d}}{\mathrm{d}t}\hdisp \)
\STATE \( \mathcal{L}_\params \gets ||\hdisp - \texttt{stopgrad}(\hdisp_{\text{tgt}})||^2_2 \)
\end{algorithmic}
\label{app:alg:training-with-augmentation}
\end{algorithm}

\paragraph{Sampling timesteps.}
The choice of the base distribution $q(\tau)$ for the timestep $\t$ has a significant impact on the structure of the loss landscape and thus the training stability of our method~\citep{zhang2025alphaflow}. We experiment with three different choices: (i) the uniform distribution, (ii) the absolute difference of two logit-normal random variables, and (iii) our proposed mixture distribution. In all cases, $q(\tau)$ is defined over $[0,1]$. During training, we sample $\tau \sim q(\tau)$ and set
\begin{equation}
    \t = \tau \cdot \t_{\max},
\end{equation}
where $\t_{\max}$ is the maximal timestep used during training. First, we draw $\tau \sim \mathcal{U}(0,1)$ as a simple baseline. Second, inspired by~\citet{geng2025mean}, we sample $\tau$ as the absolute difference of two i.i.d. logit-normal random variables~\citep{esser2024scaling}. Specifically, we sample $z_1, z_2 \sim \mathcal{N}(-0.4, 1.0)$, map them to $(0,1)$ via the sigmoid function $\sigma(\cdot)$, and set
\begin{equation}
    \tau = \bigl\lvert \sigma(z_1)-\sigma(z_2) \bigr \rvert.
\end{equation}
In our experiments, we find that emphasizing smaller timesteps can help the training stability for some systems. However, this distribution has low density for $[0.8, 1.0]$, which results in an increased integration error for timesteps $\t \rightarrow \t_{\max}$ (see \cref{app:sec:tau_sampling}). To mitigate this issue, we propose a mixture distribution, which allows learning for larger $\t$, while maintaining the stability benefits of right-skewed distributions. Specifically, we define
\begin{equation}
    q(\tau) = 0.98\mathcal{B}(1,2) + 0.02\mathcal{U}(0,1),
\end{equation}
where $\mathcal{B}(\cdot, \cdot)$ is a beta distribution and $\mathcal{U}(\cdot, \cdot)$ is a uniform distribution. We visualize all three distributions in \cref{app:fig:t-sampling}.

\begin{figure}[h]
    \centering
    \includegraphics[width=0.5\linewidth]{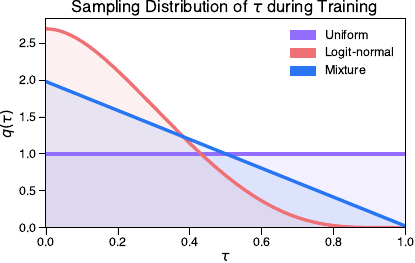}
    \caption{Sampling distributions for $\tau$. We experiment with three different $\tau$-sampling distributions in this work: (i) a uniform distribution, (ii) the logit-normal difference distribution inspired by~\citet{geng2025mean}, and (iii) our proposed mixture distribution.}
    \label{app:fig:t-sampling}
\end{figure}

\paragraph{Force matching ratio.}
As discussed in \cref{sec:methods}, the accurate learning at $\t = 0$ is essential, as it anchors the consistency condition for timesteps $\t> 0$. However, for any continuous distribution $q(\tau)$ over $[0,1]$, it holds $\mathbb{P}(\tau=0)=0$. Therefore, with probability $q_{\t=0}$, we set $\t = 0$ during training. Otherwise, we draw $\tau \sim q(\tau)$ and set $\t = \tau \cdot \t_{\max}$.

\paragraph{Sampling momenta.}
Standard MLFF datasets~\cite{eastman2023spice, ganscha2025qcml, levine2025open} typically consist of single point calculations with molecular geometries and force labels $(\pos, \force)$ but without $\mom$ labels. For each training example, we therefore draw momenta $\mom$ independently from Maxwell--Boltzmann distributions at varying temperatures. Specifically, we sample $T \sim \mathcal{N}(\mu_T, \sigma_T^2)$ and ensure non-negative values via clipping $\tilde{T} = \max\{0, T\}$. We then draw
\begin{equation} \label{app:eq:maxwell-boltzmann}
    \mom_i \sim \mathcal{N}\left(0, \mass_ik_B\tilde{T}\,\mathbb{I}_3\right),
\end{equation}
where $\mass_i$ is the mass of the $i$-th particle, and $k_B$ is the Boltzmann constant. This is justified
for the \textit{separable} Hamiltonians commonly assumed in MD, where the force depends only on $\pos$ and is independent of $\mom$. At inference time, we may similarly sample the initial momenta, but from the Maxwell--Boltzmann distribution at a fixed target temperature, i.e., $\sigma^2_T=0$. 

\paragraph{Training with zero drift, zero angular momentum and zero momentum.} For an $N$-particle system with positions $\{\pos_i\}_{i=1}^N$, momenta $\{\mom_i\}_{i=1}^N$, and masses $\{\mass_i\}_{i=1}^N$, let $M=\sum_{i=1}^N \mass_i$ denote the total mass of the system. To simplify the training objective, we ensure $\sum_{i=1}^N \mom_i = \boldsymbol{0}$ by removing the center-of-mass drift $\vel^{\text{(com)}} = \frac{1}{M} \sum_{i=1}^N \mom_i$ from the momenta:
\begin{equation}
    \label{app:eq:remove-drift}
    \mom_i \leftarrow \mom_i - \mass_i \vel^{\text{(com)}}.
\end{equation}
This transformation corresponds to a transfer to a different inertial system and therefore can be done without changing the physics of the system. During simulation, we ensure that the system is always in the same inertial state as during training via a lightweight inference filter. As the kinetic energy may change under the drift removal, we rescale the momenta,
\begin{equation} \label{app:eq:momenta-rescaling}
    \mom_i \leftarrow \sqrt{\dfrac{T_\text{tgt}}{T_\text{cur}}} \mom_i,
\end{equation}
where $T_\text{cur}$ denotes the temperature after the drift removal and $T_\text{tgt}$ denotes either (i) the temperature before the drift removal during training or (ii) the target temperature of the simulation during inference. We compute temperatures $T_\text{cur}$ (and $T_\text{tgt}$ during training) based on $3N-3$ degrees of freedom, where we have accounted for the zero center-of-mass drift. 

A fundamental property of Hamiltonian mechanics is the conservation of total angular momentum for an isolated system. The total angular momentum is defined as $\mathbf{L} = \sum_{i=1}^N \boldsymbol{r}_i \times \mom_i$ with $\boldsymbol{r}_i = \pos_i - \pos^\text{(com)}$ and $\pos^\text{(com)} = \frac{1}{M} \sum_{i=1}^N \mass_i\pos_i$. However, total angular momentum, unlike total linear momentum, changes the dynamics of the system, and therefore cannot be simply removed. Still, we found it beneficial in our early experiments to occasionally present the edge case of zero total angular momentum to the model. Therefore, with probability $q_{\mathbf{L}=\mathbf{0}}$ we train with zero angular momentum by removing the global rotational component from the momenta,
\begin{equation}
    \mom_i \leftarrow \mom_i - \mass_i \left( \boldsymbol{\omega}\times \boldsymbol{r}_i \right),
\end{equation}
where $\boldsymbol{\omega}$ is obtained by solving $\mathbf{I}\boldsymbol{\omega}=\mathbf{L}$ with the inertia tensor $\mathbf{I} = \sum_{i=1}^N \mass_i \left( \|\boldsymbol{r}_i\|^2 \mathbb{I}_3 - \boldsymbol{r}_i \boldsymbol{r}_i^\top \right)$. 

Another interesting edge case arises if the momenta for the initial state in phase space are entirely zero, where the dynamics are fully determined by the forces of the system. Therefore, with probability $q_{\mom=\mathbf{0}}$ we train with zero momenta: 
\begin{equation}
\mom_i \leftarrow \mathbf{0}.
\end{equation}
Note that for larger $\t$ the momenta still accumulate and have to be taken into account by the model accordingly.

\paragraph{Training hyperparameters.} We use the Adam optimizer~\citep{kingma2015adam} with $\beta_1=0.9$, $\beta_2=0.95$. We linearly warm up the learning rate over the first 1\% of training steps from $\mu_0 = 10^{-6}$ to $\mu_{\text{max}} = 10^{-4}$, followed by a cosine decay to $\mu_{\text{min}} = 10^{-8}$. We do not use weight decay and clip the global gradient norm to $5$. We apply equal loss weighting, i.e., $\lambda_\vel = 1$ and $\lambda_\force = 1$ and use the mixture distribution as the base distribution $q(\tau)$ in our experiments. For all 3D systems, we set $q_{\mathbf{L}=\mathbf{0}}=0.25$, $q_{\mom=\mathbf{0}}=0.25$, and $q_{\t=0}=0.75$. For all other systems, we set $q_{\mathbf{L}=\mathbf{0}}=0$, $q_{\mom=\mathbf{0}}=0$, and $q_{\t=\mathbf{0}}=0$.

\subsection{Inference with Hamiltonian Flow Maps} \label{app:inference}
The direct prediction of forces using a non-equivariant model architecture can introduce artifacts during simulation, including violations of rotational equivariance, momentum drift, and energy instabilities. To mitigate these effects, we can optionally apply a set of filters at inference time, similar to TrajCast~\citep{thiemann2025forcefree} and FlashMD~\citep{bigi2025flashmd}. 

The inference algorithm for a single step is summarized in~\cref{app:alg:inference}, where $\mathrm{ApplyUpdate}$ refers to the HFM update in \cref{eq:flow-parameterization}. Starting from an initial state ($\pos_0$, $\mom_0$) at physical time $t = 0$, we iteratively apply \cref{app:alg:inference} with fixed $\t$, until the desired simulation time is reached. The initial state ($\pos_0$, $\mom_0$) consists of the positions $\pos_0$ of a relaxed structure and momenta $\mom_0$ sampled from the Maxwell--Boltzmann distribution at target temperature $T$. Similar to training, we ensure zero net linear momentum and then rescale $\mom_0$ to match the target temperature $T$ (see \cref{app:eq:remove-drift} and \cref{app:eq:momenta-rescaling}).

\begin{algorithm}
\caption{Inference with Hamiltonian Flow Maps}
\begin{algorithmic}[1]
\REQUIRE Neural network \( \hdisp^\params \), current state \( (\pos_t,  \mom_t) \), timestep $\t$, rotation matrix \( \boldsymbol{R} \sim \mathrm{SO}(3) \)
\STATE \( \pos_t\phantom{\,\, \t}, \mom_t\phantom{\,\, \t} \gets \mathrm{RandomRotationFilter}(\boldsymbol{R}, \pos_t, \mom_t) \)
\STATE \( \pos_{t+\t}, \mom_{t+\t} \gets \mathrm{ApplyUpdate}(\hdisp^\params, \pos_t, \mom_t, \t) \)
\STATE \( \pos_{t+\t}, \mom_{t+\t} \gets \mathrm{RandomRotationFilter}(\boldsymbol{R}^{-1}, \pos_{t+\t}, \mom_{t+\t}) \)
\STATE \( \pos_{t+\t}, \mom_{t+\t} \gets \mathrm{RemoveDriftFilter}(\pos_{t+\t}, \mom_{t+\t}) \)
\STATE \( \phantom{\pos_{t+\t},\ } \mom_{t+\t}\gets \mathrm{CoupledConservationFilter}(\pos_{t+\t}, \mom_{t+\t}) \)
\STATE \( \phantom{\pos_{t+\t},\ } \mom_{t+\t} \gets \mathrm{Thermostat}(\mom_{t+\t}) \)
\end{algorithmic}
\label{app:alg:inference}
\end{algorithm}

In the following, we discuss the remove drift and random rotation filter from FlashMD. Our proposed coupled energy and angular momentum conservation filter and the used thermostats are discussed in~\cref{app:subsec:coupled-conservation-filter,app:subsec:thermostats}, respectively. 

\paragraph{Remove drift filter.} For an $N$-particle system at time $t$ with positions $\{\pos_{t,i}\}_{i=1}^N$, momenta $\{\mom_{t,i}\}_{i=1}^N$, and masses $\{\mass_i\}_{i=1}^N$, let $M=\sum_{i=1}^N \mass_i$ denote the total mass. We aim to achieve the conservation of total linear momentum of the system, i.e.,
\begin{equation}
    \sum_{i=1}^N \mom_{t+\t,i} - \sum_{i=1}^N \mom_{t,i} = 0.
\end{equation}
One can show that this condition is satisfied by applying a \textit{local} translation to the updated momenta
\begin{equation}
    \mom_{t+\t,i} \gets \mom_{t+\t,i} + \mass_i \left( -\vel^{\text{(com)}}_{t+\t} + \vel^{\text{(com)}}_t \right),
\end{equation}
where
\begin{equation}
    \vel^{\text{(com)}}_{t+\t} = \dfrac{1}{M} \sum _{i=1}^N \mom_{t + \t,i},\qquad
    \vel^{\text{(com)}}_t = \dfrac{1}{M} \sum _{i=1}^N \mom_{t, i}, \quad
\end{equation}
denote the center-of-mass for the current velocities $\vel_t$, and updated velocities $\vel_{t+\t}$, respectively.

Since the total momentum is constant, the center-of-mass of the system follows a uniform linear motion, i.e.,
\begin{equation}
    \sum_{i=1}^N \mass_i \pos_{t+\t,i} - \sum_{i=1}^N \mass_i \pos_{t,i} = \t \sum_{i=1}^N \mom_{t,i}.
\end{equation}
One can show that this condition is satisfied by applying a \textit{global} translation to the updated positions
\begin{equation}
    \pos_{t+\t,i} \leftarrow \pos_{t+\t,i} - \pos^{\text{(com)}}_{t+\t} + \pos^{\text{(com)}}_t + \t \vel^{\text{(com)}}_t,
\end{equation}
where
\begin{equation}
    \pos^{\text{(com)}}_{t} = \dfrac{1}{M} \sum_{i=1}^N \mass_i\pos_{t,i}, \quad
    \pos^{\text{(com)}}_{t+\t} = \dfrac{1}{M} \sum_{i=1}^N \mass_i\pos_{t+\t,i}, \quad 
    \vel^{\text{(com)}}_{t} = \dfrac{1}{M} \sum _{i=1}^N \mom_{t,i},
\end{equation}
denote the center-of-mass for the current positions $\pos_t$, updated positions $\pos_{t+\t}$, and current velocities $\vel_t$, respectively.

\paragraph{Random rotation filter.}

During simulation, we apply random rotations to states to average out artifacts introduced due to violations of rotational equivariance. Notice that the rotation has to be applied to the centered positions, as otherwise this operation may shift the frame center. We summarize the $\mathrm{RandomRotationFilter}$ function in \cref{app:alg:random-rotation-filter}, which applies a random rotation $\boldsymbol{R}$ to the current state $(\pos_t, \mom_t)$ and its inverse $\boldsymbol{R}^{-1}$ to the updated state $(\pos_{t+\t}, \mom_{t+\t})$ (see \cref{fig:architecture} and \cref{app:alg:inference}). The rotations $\boldsymbol{R}, \boldsymbol{R}^{-1}$ are applied based on the $\mathrm{ApplyRotation}$ function in \cref{app:eq:rotation-augementation}.

\begin{algorithm}
\caption{$\mathrm{RandomRotationFilter}$}
\begin{algorithmic}[1]
\REQUIRE Neural network \( \hdisp^\params \), state \( (\pos, \mom) \), rotation matrix \( \boldsymbol{R} \sim \mathrm{SO}(3) \)
\STATE \( \pos \gets \mathrm{ApplyRotation}(\boldsymbol{R}, \pos - \pos_{\text{mean}}) + \pos_{\text{mean}} \)
\STATE \( \mom \gets \mathrm{ApplyRotation}(\boldsymbol{R}, \mom) \)
\end{algorithmic}
\label{app:alg:random-rotation-filter}
\end{algorithm}

\subsection{Coupled Conservation Filter for Angular Momentum and Energy} \label{app:subsec:coupled-conservation-filter}

In this subsection, we describe our proposed coupled conservation filter for angular momentum and energy. Compared to sequential update schemes, this filter solves a constrained optimization problem in closed form to minimally adjust the momenta after each update step, such that total energy and angular momentum are conserved simultaneously.

\paragraph{Problem formulation.} For an $N$-particle system with updated positions $\{\pos_{t+\t, i}\}_{i=1}^N$, updated momenta $\{\mom_{t+\t, i}\}_{i=1}^N$, and masses $\{\mass_i\}_{i=1}^N$, let $M=\sum_{i=1}^N \mass_i$ denote the total mass of the system, and $\boldsymbol{r}_{t+\t, i} = \pos_{t+\t, i} - \pos^\text{(com)}_{t+\t}$ denote the updated position of the $i$-particle expressed in the center-of-mass frame with $\pos^\text{(com)}_{t+\t} = \frac{1}{M} \sum_{i=1}^N \mass_i\pos_{t+\t, i}$. 

We seek corrected updated momenta $\{\mom'_{t+\t, i}\}_{i=1}^N$ such that
\begin{align}
\sum_{i=1}^N \frac{\|\mom'_{t+\t, i}\|^2}{2\mass_i} &= K_{\text{tgt}}, \label{eq:energy_constraint}\\
\sum_{i=1}^N \boldsymbol{r}_{t+\t, i} \times \mom'_{t+\t, i} &= \mathbf{L}_{\mathrm{tgt}}, \label{eq:angmom_constraint}
\end{align}
while minimizing the mass-weighted momentum change
\begin{equation}
\min_{\mom'} \;
\sum_{i=1}^N \frac{1}{2\mass_i}\|\mom'_{t+\t, i} - \mom_{t+\t, i}\|^2 .
\end{equation}

\paragraph{Target definitions.} In \cref{eq:angmom_constraint}, the target $\mathbf{L}_{\text{tgt}}$ is the total angular momentum before the simulation step:
\begin{align}
    \mathbf{L}_{\mathrm{tgt}} = \sum_{i=1}^N \boldsymbol{r}_{t,i} \times \mom_{t,i}.
\end{align}

In \cref{eq:energy_constraint}, the target kinetic energy $K_{\text{tgt}}$ refers to 
\begin{align}
    K_{\text{tgt}}=K_{\text{cur}}-\Delta E_{\text{tot}}.
\end{align}

The total energy decomposes into kinetic and potential energy. We need to evaluate a potential $V$
to compute the potential energy before and after the simulation step. For this we can either apply (i) a separately trained \gls{MLFF} or (ii) attach a separate head for predicting the potential energy to the HFM model. In this work, we experiment with both strategies (see \cref{app:sec:experimental-setup}). Notice that the computation of the potential energy can be reused for the next simulation step, such that the computational overhead is a single potential evaluation per simulation step. The kinetic energy
can be obtained directly from the momenta. We use both to compute the total energy before and after the simulation step. The difference in total energy, denoted as $\Delta E_{\text{tot}}$, needs to be accounted for by correction of the momenta after the simulation step.

\paragraph{Lagrangian.}
Introduce a scalar Lagrange multiplier $\alpha \in \mathbb{R}$ for the kinetic-energy
constraint \eqref{eq:energy_constraint} and a vector Lagrange multiplier $\boldsymbol{\beta} \in \mathbb{R}^3$ for the angular-momentum constraint \eqref{eq:angmom_constraint}. The Lagrangian is
\begin{equation}
\mathcal{L}(p',\alpha,\boldsymbol{\beta})
=
\sum_{i=1}^N \frac{1}{2\mass_i}\|\mom'_{t+\t, i} - \mom_{t+\t, i}\|^2
+
\alpha\!\left(
\sum_{i=1}^N \frac{\|\mom'_{t+\t, i}\|^2}{2\mass_i} - K_{\text{tgt}}
\right)
+
\boldsymbol{\beta} \cdot
\left(
\sum_{i=1}^N \boldsymbol{r}_{t+\t, i} \times \mom'_{t+\t, i} - \mathbf{L}_{\text{tgt}}
\right).
\end{equation}

Taking the derivative of $\mathcal{L}$ with respect to $\mom'_{t+\t, i}$ and setting it to
zero yields
\begin{equation}
\frac{1}{\mass_i}(\mom'_{t+\t, i} - \mom_{t+\t, i})
+ \alpha \frac{\mom'_{t+\t, i}}{\mass_i}
+ \boldsymbol{\beta} \times \boldsymbol{r}_{t+\t, i}
= 0 .
\end{equation}

Multiplying by $\mass_i$ and rearranging gives
\begin{equation} \label{eq:lagrangian-mom}
\mom'_{t+\t, i}
=
\frac{1}{1+\alpha}
\bigl(
\mom_{t+\t, i} - \mass_i(\boldsymbol{\beta} \times \boldsymbol{r}_{t+\t, i})
\bigr).
\end{equation}

\paragraph{Angular momentum constraint.}
Inserting \cref{eq:lagrangian-mom} into the angular-momentum constraint \eqref{eq:angmom_constraint} yields
\begin{equation}
\frac{1}{1+\alpha}
\left(
\mathbf{L} - \mathbf{I}\boldsymbol{\beta}
\right)
= \mathbf{L}_{\text{tgt}}
\end{equation}
with inertia $\mathbf{I} = \sum_{i=1}^N \mass_i \left( \|\boldsymbol{r}_{t+\t, i}\|^2 \mathbb{I}_3 - \boldsymbol{r}_{t+\t, i} \boldsymbol{r}_{t+\t, i}^\top \right)$ and angular momentum $\mathbf{L} = \sum_{i=1}^N \boldsymbol{r}_{t+\t, i} \times \mom_{t+\t, i}$. 

Solving for $\boldsymbol{\beta}$ gives
\begin{equation}
\boldsymbol{\beta}(\alpha)
=
\mathbf{I}^{-1}\mathbf{L}
-
(1+\alpha) \mathbf{I}^{-1}\mathbf{L}_{\text{tgt}} .
\end{equation}

\paragraph{Momentum decomposition.}
For ease of notation, let's define:
\begin{align}
\lambda:= 1+\alpha,\qquad 
\mom^{(0)}_{t+\t, i}
:=
\mom_{t+\t, i} - \mass_i\bigl((\mathbf{I}^{-1}\mathbf{L})\times \boldsymbol{r}_{t+\t, i}\bigr),\qquad
\mom^{(1)}_{t+\t, i}
:=
\mass_i\bigl((\mathbf{I}^{-1}\mathbf{L}_{\text{tgt}})\times \boldsymbol{r}_{t+\t, i}\bigr).
\end{align}

Using the expression for $\boldsymbol{\beta}(\lambda)$, we obtain
\begin{equation}
\mom_{t+\t, i} - \mass_i(\boldsymbol{\beta}(\lambda)\times \boldsymbol{r}_{t+\t, i})
=
\mom^{(0)}_{t+\t, i} + \lambda \mom^{(1)}_{t+\t, i}.
\end{equation}

And for the corrected momenta
\begin{equation}
\mom'_{t+\t, i}(\lambda)
=
\frac{1}{\lambda}
\bigl(
\mom^{(0)}_{t+\t} + \lambda \mom^{(1)}_{t+\t,i}
\bigr).
\end{equation}

\paragraph{Kinetic energy as a function of $\lambda$.}
The kinetic energy of the corrected momenta is
\begin{equation}
K(\lambda)
=
\sum_{i=1}^N \frac{\|\mom'_{t+\t,i}(\lambda)\|^2}{2\mass_i}
=
\frac{1}{\lambda^2}
\sum_{i=1}^N \frac{\|\mom^{(0)}_{t+\t,i} + \lambda \mom^{(1)}_{t+\t,i}\|^2}{2\mass_i}.
\end{equation}

Introduce the scalar quantities
\begin{align}
A
=
\sum_{i=1}^N \frac{\|\mom^{(0)}_{t+\t,i}\|^2}{2\mass_i},\qquad
B
=
\sum_{i=1}^N \frac{\mom^{(0)}_{t+\t,i} \cdot \mom^{(1)}_{t+\t,i}}{\mass_i},\qquad
C
=
\sum_{i=1}^N \frac{\|\mom^{(1)}_{t+\t,i}\|^2}{2\mass_i}.
\end{align}

With these definitions, we have
\begin{equation}
K(\lambda)
=
\frac{1}{\lambda^2}
\left(
A + B\lambda + C\lambda^2
\right).
\end{equation}

\paragraph{Quadratic equation for $\lambda$.}
Imposing the energy constraint \eqref{eq:energy_constraint},
$K(\lambda) = K_{\text{tgt}}$, leads to a quadratic equation:
\begin{equation}
(C - K_{\text{tgt}})\lambda^2 + B\lambda + A = 0 .
\end{equation}

Once $\lambda$ is determined, the energy multiplier is recovered as
\begin{equation}
\alpha = \lambda - 1 .
\end{equation}

\paragraph{Final corrected momenta.}
The corrected momenta satisfying both constraints are
\begin{equation}
\mom'_{t+\t, i}
=
\frac{1}{1 + \alpha}
\Bigl(
\mom_{t+\t, i} - \mass_i\bigl(\boldsymbol{\beta}(\alpha)\times \boldsymbol{r}_{t+\t, i}\bigr)
\Bigr),
\qquad
\boldsymbol{\beta}(\alpha) = \mathbf{I}^{-1}\mathbf{L} - (1+\alpha) \mathbf{I}^{-1}\mathbf{L}_{\mathrm{tgt}} .
\end{equation}

We note that this optimization problem can be efficiently solved as a quadratic equation, however it does not represent the direct projection onto the correct manifold preserving all relevant quantities, as we do not correct predicted positions. This is mainly motivated by inference speed, since updating positions as well would lead to a non-linear problem, which must be solved iteratively and requires evaluating the \gls{MLFF} potential multiple times, which is computationally expensive.

\subsection{Thermostats} \label{app:subsec:thermostats}

We perform NVT simulations based on three different thermostats throughout this work:
\paragraph{Langevin thermostat.} The Langevin thermostat~\citep{bussi2007accurate} applies \emph{local} stochastic corrections to momenta, allowing to sample the constant-temperature (NVT) ensemble. This makes it the most robust choice for our experiments. During simulations, we apply all our inference filters (see \cref{sec:architecture-filters}), i.e., drift removal, coupled conservation of total energy and total angular momentum, as well as random rotations.

\paragraph{CSVR thermostat.} The CSVR thermostat~\citep{bussi2007canonical} applies \emph{global} stochastic velocity rescaling, to sample from the NVT ensemble with the desired target temperature. We find that global velocity rescaling in some cases leads to interference with our global correction for total energy, and therefore use only the drift removal filter, random rotation filter, and an alternative filter that only preserves angular momentum.

\paragraph{Nosé-Hoover thermostat.} The Nosé-Hoover thermostat~\citep{martyna1996explicit} applies \emph{global} corrections in a fully deterministic and reversible way. We note that the lack of stochasticity amplifies model errors, which are caused by the non-symplectic nature of our map. In some of our experiments this leads to noticeable artifacts. Therefore, we use only the drift removal filter, random rotation filter, and an alternative filter that only preserves angular momentum for this thermostat.

\subsection{Units} We use the default \texttt{ase}~\citep{larsen2017ase} units throughout our molecular experiments, i.e., energy in eV, distances in Å, forces in eV/Å.

\subsection{Compute Infrastructure} We perform training and simulation of our models on a single NVIDIA A100 80GB GPU or NVIDIA H100 80GB GPU, depending on availability. For fair comparison, the timings were always measured on a single NVIDIA A100 80GB GPU.

\subsection{Software Licences} In this work, we use \texttt{jax}~\citep{bradburry2018jax} (Apache-2.0) and the its machine learning library \texttt{flax}~\citep{heek2024flax} (Apache-2.0). We adopt the transformer architecture from~\citet{frank2025sampling} (MIT), which is implemented on top of \texttt{e3x}~\citep{unke2024e3x} (Apache-2.0). For simulation utilities, we use \texttt{ase}~\citep{larsen2017ase} (GNU LGPLv2.1).

\pagebreak
\section{Sufficiency of the Proposed Loss} \label{app:sec:sufficiency}
In \cref{eq:derivative-both-sides,eq:mean-identity,eq:u-total-derivative}, we derive a necessary condition that every Hamiltonian flow map needs to fulfill. In the following, we will derive that this is also a sufficient condition following ideas from \citet{geng2025mean}.

While~\citet{geng2025mean} derive a consistency condition by differentiating the average velocity field with respect to the end time (a backward consistency condition), our method relies on differentiating with respect to the start time (a forward consistency condition). ~\citet[Appendix B.3]{geng2025mean} show the sufficiency of their loss, and we show here that our forward formulation provides an equally sufficient condition for learning correct Hamiltonian dynamics. We start from~\cref{eq:derivative-both-sides}, which is equivalent to the loss term used in~\cref{eq:loss}, as we show in~\cref{sec:methods}. Once we revert the derivative with respect to an integration by $t$, an integration constant appears on both sides such that
\begin{equation}
\frac{\mathrm{d}}{\mathrm{d}\apos}
\left[
\t\hdisp(\pos_\apos, \mom_\apos, \t)
\right]
=
\frac{\mathrm{d}}{\mathrm{d}\apos}
\int_\apos^\bpos
\begin{pmatrix}
\vel_\tau \\
\force_\tau
\end{pmatrix}
\mathrm{d}\tau
\Longrightarrow
\t\hdisp(\pos_\apos, \mom_\apos, \t)+C_1
=
\int_\apos^\bpos
\begin{pmatrix}
\vel_\tau \\
\force_\tau
\end{pmatrix}
\mathrm{d}\tau
+C_2.
\end{equation}
For the special case of $\apos=\bpos\rightarrow\t=0$, we have $\int_\apos^\bpos
\begin{pmatrix}
\vel_\tau \\
\force_\tau
\end{pmatrix}
\mathrm{d}\tau
=0$ and therefore $C_1=C_2$.

As both integration constants are equal, this shows that~\cref{eq:displacement-def}$\Longrightarrow$~\cref{eq:derivative-both-sides} and therefore that our loss is sufficient to learn the correct integral. Crucially, for Hamiltonian systems, this implies that if the training loss is perfectly minimized, the predictions of the network will model the true dynamics of the system. The only systematic source of error would be a bias in the given force labels, or a generalization error introduced by the model.

\section{Comparison with Few-Step Diffusion Models}
\label{app:sec:few-step-diffusion-similarity}

\begin{figure}[h]
    \centering
    \includegraphics[width=\linewidth]{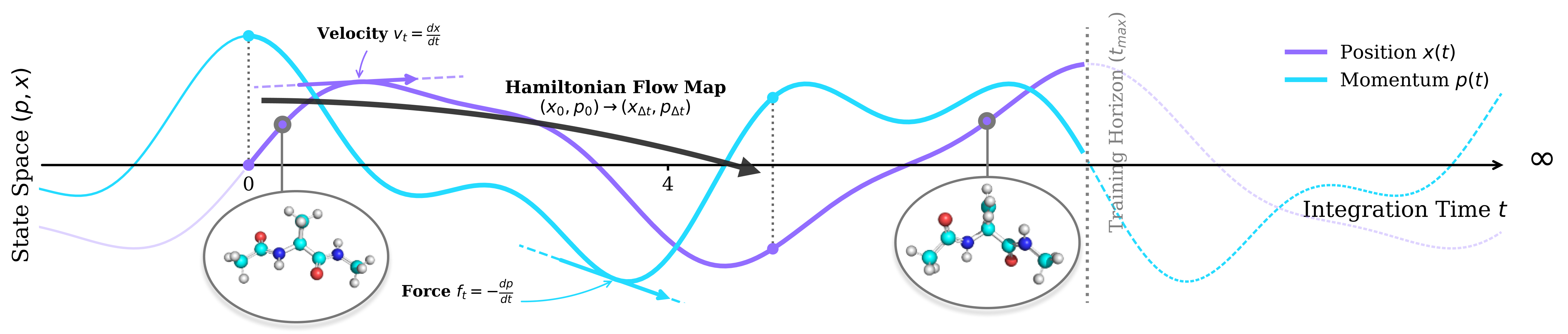}
    \caption{Our loss formulation can be understood in a similar way like few-step generative models. Instead of integrating the probability flow ODE, we integrate Hamiltonian dynamics. In our setting, the forward evolution of momenta and positions must be considered as a coupled integration problem, i.e., the Hamiltonian Flow Map (HFM) predicts a future state in phase space.}
    \label{app:fig:hamiltonian-flow-map}
\end{figure}

\paragraph{Generative sampling vs. deterministic dynamics.} Our approach shares a conceptual foundation with accelerated sampling in diffusion and flow-matching models~\citep{sohldickstein2015nonequilibrium, ho2020ddpm, song2021score, albergobuilding, lipman2023flow}. Recent advances in this field employ few-step generation techniques, such as consistency models or rectified flows, to bypass the expensive numerical integration of the probability flow-ODE~\citep{song2023consistency, song2024improved, lu2025simplifying,liu2023flow}. We visualize our training target in~\cref{app:fig:hamiltonian-flow-map}. 

As described in~\cref{sec:computational-approach}, our learning process can be viewed as a form of self-distillation across time. \citet{zhang2025alphaflow} provide a rigorous analysis of this mechanism in the context of generative modeling, demonstrating that the \emph{Mean Flow} loss~\cite{geng2025mean} analytically decomposes into two distinct gradient components: \textit{Trajectory Flow Matching} and \textit{Trajectory Consistency}. In our context, the loss decomposes into a force matching and an integral fitting part.

Unlike generative approaches that utilize noise injection and denoising scores to construct a transport map between distributions, our training objective is entirely deterministic. Because classical Hamiltonian dynamics dictates a precise, causal evolution for every point in phase space, our objective can be seen as a physics-informed loss that directly penalizes deviations from the integrated equations of motion.

\paragraph{Probability flow-ODE vs. Hamiltonian dynamics.} Generative flow models typically rely on a probability flow-ODE designed to transport a prior distribution to a target data distribution (and thus always requires data pairs). The vector field governing this flow is often a design choice, and can be constructed to be straight or smooth to facilitate easy integration~\citep{liu2023flow}, which has a major impact on the required number of integration steps. In contrast, the vector field in our setting is dictated by physical laws, specifically Hamilton's equations. Unlike diffusive dynamics often found in generative models, Hamiltonian dynamics respect the conservation of energy and symplecticity. Consequently, our objective is not to find any flow that transports samples between distributions, but to learn the specific physical flow, i.e., the true dynamics of the system, which are unknown a priori.

\paragraph{Finite interval vs. unbounded time.} Generative flows operate on a fixed, finite time interval, typically $t \in [0, 1]$, solving a boundary value problem where the endpoints are known. Hamiltonian simulation, conversely, is an initial value problem over an effectively unbounded time horizon ($t \to \infty$). We do not aim to reach a specific target distribution (we do not even know the target during training), but rather to iterate the learned map indefinitely in an auto-regressive manner. This requires the learned map to be stable under repeated composition, crucially evoking the need to avoid catastrophic accumulation of errors, a constraint that is generally less important in single-shot generative tasks.

\paragraph{The Lyapunov limit.} For generative flow models, the complexity of few-shot maps is determined by the interplay of source and target distributions, as well as the choice of the imposed vector field or interpolant. In Hamiltonian mechanics, the difficulty is governed by the system's chaoticity, quantified by the maximal Lyapunov exponent: Even for simple physical systems that might seem easy to integrate at first, the divergence of trajectories typically grows exponentially, imposing a hard physical limit beyond which the flow map becomes effectively unlearnable by a smooth neural network. Generative flow models can sacrifice accuracy for speed to a certain extent by smoothing out complex transport paths. In contrast, our method must faithfully reconstruct the evolution of phase space up to some $\Delta t$ that lies within the chaotic limit.

\section{Model Architecture} \label{app:sec:model-architecture}

In this section, we describe the three model architectures used throughout this work. However, we want to emphasize that our proposed framework is largely architecture-agnostic: any existing architecture, which is suitable for the underlying system (e.g. architectures for atomistic systems from the generative modeling or machine learning force field literature) can be used to learn Hamiltonian Flow Maps (HFMs) with minimal architectural changes. We provide a list of all hyperparameters and additional implementation details in the corresponding subsections of each experiment in \cref{app:sec:experimental-setup}.

\subsection{Multi-Layer Perceptron}

For the toy datasets, we learn Hamiltonian Flow Maps (HFMs) using a simple Multi-Layer Perceptron.

\paragraph{Embeddings.} We obtain the initial embedding of the $i$-th particle as the combination of features based on the time, position and momentum information using three separate two-layer MLPs,
\begin{equation}
    \bh_i = \MLP(\emb_{t}(\t)) + \MLP(\pos_i) + \MLP(\mom_i) \in \mathbb{R}^H,
\end{equation}
where $\emb_{t}(\t) \in \mathbb{R}^H$ denotes Gaussian random Fourier features~\citep{tancik2020fourier} of the time $\t \in [0, \t_{\max}]$. As we encode absolute positions $\pos_i \in \mathbb{R}^3$, these embeddings do not obey any form of Euclidean symmetries: they do not transform equivariantly under rotation of the input positions, and they are also not invariant to translations of the system.

\paragraph{Refinement layers.} The initial embedding is further refined using a three-layer MLP,
\begin{equation}
    \bh_i = \MLP(\bh_i) \in \mathbb{R}^H,
\end{equation}
which produces the final latent representation shared by all prediction heads.

\paragraph{Prediction heads.} Finally, we predict the mean velocities and forces for each particle using two separate prediction heads,
\begin{align}
\bar{\vel}_{i} &= \MLP(\bh_i) \in \mathbb{R}^3,\\
\bar{\force}_{i} &= \MLP(\bh_i) \in \mathbb{R}^3,
\end{align}
where each prediction head consists of a two-layer MLP.

\subsection{Translation-invariant Transformer}
For the gravitational N-body particle system and the molecular systems, we adopt the translation-invariant Transformer architecture for 3D geometries proposed by~\citet{frank2025sampling}, which operates on pairwise displacement vectors rather than the absolute particle positions. To adapt this architecture to our setting, we introduce two modifications: 
\begin{itemize}
    \item[(1)] particle-wise conditioning on time, velocities, and masses, and 
    \item[(2)] multiple prediction heads with task-specific adaptive layer normalizations.
\end{itemize}
For convenience, we summarize the components of the original architecture below and describe our modifications in detail. We provide an overview of the model architecture in \cref{app:fig:transformer}.

\begin{figure}[tb]
    \centering
    \includegraphics[width=0.6\linewidth]{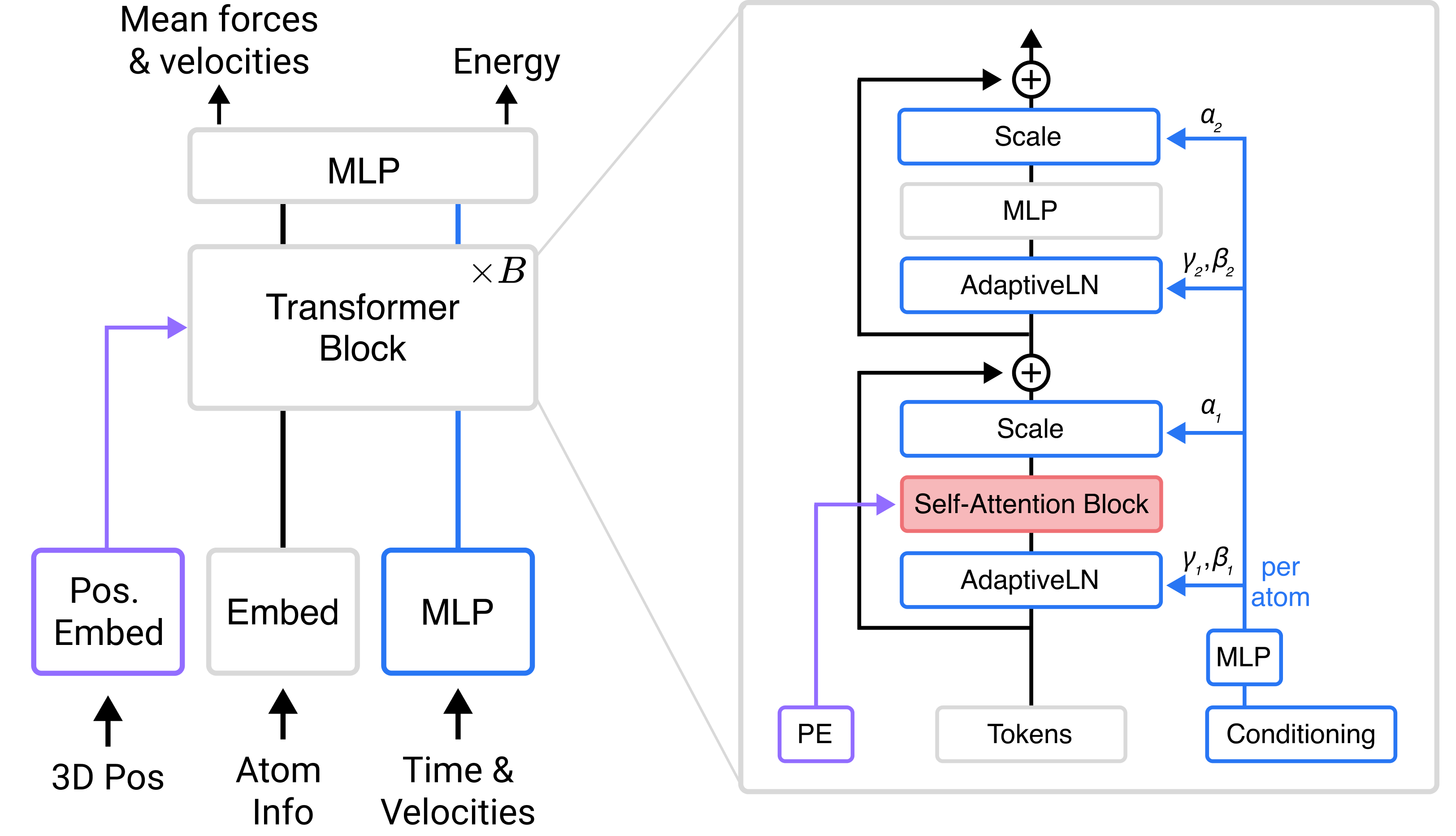}
    \caption{Translation-invariant transformer architecture adapted from \citet{frank2025sampling}. PE: Relative positional embeddings are used for scaling attention coefficients in the transformer. Conditioning: Time and Velocity inputs are passed into each transformer block via adaptive layer norms.}
    \label{app:fig:transformer}
\end{figure}

\paragraph{Token embeddings.} The initial token representation of the $i$-th particle is obtained via
\begin{equation}
    \bh_i = \emb_z(z_i) \in \mathbb{R}^H,
\end{equation}
where $\emb_z(z_i) \in \mathbb{R}^H$ denotes a learnable embedding based on the particle type $z_i \in \mathbb{N}_+$.

\paragraph{Positional embeddings.} We construct translation-invariant relative positional embeddings (rPE) based on the 
distance $d_{ij} = \|\pos_i - \pos_j\|_2$ and the and normalized displacement vector $\boldsymbol{\hat{r}}_{ij} = (\pos_i - \pos_j) / \|\pos_i - \pos_j\|_2$ between particle $i$ and $j$ using a two-layer MLP,
\begin{equation}
    \rpe = \MLP([\boldsymbol{\phi}(d_{ij}); \boldsymbol{\hat{r}}_{ij}] \in \mathbb{R}^H,
\end{equation}
where $\boldsymbol{\phi} : \mathbb{R} \rightarrow \mathbb{R}^{M_\text{RBF}}$ is a radial basis function, $[\cdot;\cdot]$ denotes concatenation, and $\odot$ denotes element-wise multiplication. As the relative positional embeddings are based on distance and normalized displacement vector between pairs of particles, these embeddings are translation-invariant.

\paragraph{Attention mechanism.} The self-attention update is defined as
\begin{align}
    \bh_i = \bh_i + \sum_{j=1}^N \softmax_j \left( \dfrac{(\dense_Q \bh_i)^\top (\dense_K \bh_j \odot \dense_K^E \rpe)}{\sqrt{H}} \right) \left( \dense_V\bh_j \odot \dense_V^E\rpe \right),
\end{align}
where $\dense_Q, \dense_K, \dense_V, \dense_K^E, \dense_V^E \in \mathbb{R}^{H\times H}$ are learnable weight matrices and $\odot$ denotes element-wise multiplication. For ease of notation, we present the self-attention update with a single attention head, but apply multi-head attention with $n_{\text{heads}}$~\citep{vaswani2017attention}.

\paragraph{Conditioning strategy.} We perform particle-wise conditioning based on time $\t \in [0, t_{\max}]$, as well as the mass $\mass_i > 0$ and velocity $\vel_i \in \mathbb{R}^3$ of the $i$-th particle. First, we construct the time-based conditioning token based on a two-layer MLP,
\begin{equation}
\boldsymbol{c}^{t} = \MLP(\emb_{t}(\t)) \in \mathbb{R}^H,
\end{equation}
where $\emb_{\t}(\t) \in \mathbb{R}^H$ denotes Gaussian random Fourier features~\citep{tancik2020fourier} of the time $\t$. Next, we encode the magnitude of the velocity $\vel_i$ of the $i$-th particle using a Gaussian basis expansion, similar to~\citet{thiemann2025forcefree}, followed by an element-specific linear transformation,
\begin{equation}
\emb_{\vel}(\vel_i)= \dense_{z_i} \psi(\|\vel_i\|_2) \in \mathbb{R}^{M_\text{Gaussian}},
\end{equation}
where $\psi(\cdot)=[\psi_1(\cdot), \dots, \psi_{M_\text{Gaussian}}(\cdot)]^\top\in \mathbb{R}^M$ is a set of $M_\text{Gaussian}$ Gaussian basis functions, $\dense_{z_i}\in \mathbb{R}^{M_\text{Gaussian}\times M_\text{Gaussian}}$ is a learnable weight matrix, and $z_i$ is the type of the $i$-th particle. The velocity $\vel_i$ is then encoded using a two-layer MLP,

\begin{equation}
\boldsymbol{c}^{\vel}_i = \MLP([\emb_{\vel}(\vel_i); \vel_i]) \in \mathbb{R}^H,
\end{equation}

where $\left[\cdot ; \cdot\right]$ denotes concatenation. Similarly, the mass $\mass_i$ of the $i$-th particle is encoded by another two-layer MLP 

\begin{equation}
\boldsymbol{c}^{\mass}_i = \MLP(\mass_i) \in \mathbb{R}^H.
\end{equation}

Finally, the particle-specific conditioning token $\boldsymbol{c}_i$ is obtained via concatenation followed by a two-layer MLP,

\begin{equation}
\boldsymbol{c}_i=\MLP([\boldsymbol{c}^{t}; \boldsymbol{c}^{\vel}_i ; \boldsymbol{c}^{\mass}_i]) \in \mathbb{R}^H.
\end{equation}

Given the particle-specific conditioning token $\boldsymbol{c}_i$, we can compute scale, shift and gating parameters via

\begin{equation}
\left[ \boldsymbol{\gamma}_i; \boldsymbol{\beta}_i;\boldsymbol{\alpha}_i \right] = \dense(\mathrm{SiLU}(\boldsymbol{c}_i))
\end{equation}

with weight matrix $\dense \in \mathbb{R}^{3H\times H}$ initialized to all zeros and $\boldsymbol{\gamma}_i, \boldsymbol{\beta}_i,\boldsymbol{\alpha}_i \in \mathbb{R}^H$. These parameters are used for conditioning based on adaptive layer norm ($\AdaLN$) and adaptive scale ($\AdaScale$) throughout the transformer blocks, defined as
\begin{equation}
    \AdaLN(\bh, \boldsymbol{\gamma}, \boldsymbol{\beta}) = \mathrm{LN}(\bh) \odot (1 + \boldsymbol{\gamma}) + \boldsymbol{\beta},
\end{equation}
and
\begin{equation}
    \AdaScale(\bh, \boldsymbol{\alpha}) = \bh \odot \boldsymbol{\alpha},
\end{equation}
where $\mathrm{LN}$ is layer normalization~\citep{ba2016layer} without scale and shift parameters and $\odot$ is element-wise multiplication.

\paragraph{Prediction heads.} Given the token representation $\bh^{[B]}_i \in \mathbb{R}^H$ for the $i$-th particle after $B$ transformer blocks, we predict the mean velocities and mean forces using two separate prediction heads with task-specific adaptive layer normalizations. We calculate task-specific scale and shift parameters for both tasks based on the particle-specific conditioning token $\boldsymbol{c}_i$,
\begin{equation} \label{app:eq:task-specific-adaln-adascale}
    \left[ \boldsymbol{\gamma}^{\vel}_i; \boldsymbol{\beta}^{\vel}_i; \boldsymbol{\gamma}^{\force}_i; \boldsymbol{\beta}^{\force}_i \right] = \dense(\mathrm{SiLU}(\boldsymbol{c}_i)),
\end{equation}
where $\boldsymbol{\gamma}^{\vel}_i, \boldsymbol{\beta}^{\vel}_i, \boldsymbol{\gamma}^{\force}_i, \boldsymbol{\beta}^{\force}_i \in \mathbb{R}^H$ and $\dense \in \mathbb{R}^{4H\times H}$ initialized to all zeros. Each head applies $\AdaLN$ followed by a two-layer MLP:
\begin{align}
\bar{\vel}_{i} &= \MLP(\AdaLN(\bh^{[B]}_i, \boldsymbol{\gamma}^{\vel}_i, \boldsymbol{\beta}^{\vel}_i)) \in \mathbb{R}^3,\\
\bar{\force}_{i} &= \MLP(\AdaLN(\bh^{[B]}_i, \boldsymbol{\gamma}^{\force}_i, \boldsymbol{\beta}^{\force}_i)) \in \mathbb{R}^3.
\end{align}
We optionally attach an additional prediction head to predict the particle-wise energy contributions,
\begin{equation}
    E_i = \MLP(\AdaLN(\bh^{[B]}_i, \boldsymbol{\gamma}^{E}_i, \boldsymbol{\beta}^{E}_i)) \in \mathbb{R},
\end{equation}
where the scale and shift parameters $\boldsymbol{\gamma}^{E}_i, \boldsymbol{\beta}^{E}_i  \in \mathbb{R}^H$ for the energy predictions are obtained as in \cref{app:eq:task-specific-adaln-adascale}.
Based on these predictions, we can then compute the potential energy of the system as the sum of energy contributions per particle,
\begin{equation}
    E _{\text{pot}}= \sum_{i=1}^N E_i,
\end{equation}
and use this prediction in our energy and angular momentum conservation filter during inference (see \cref{app:subsec:coupled-conservation-filter}).

\subsection{$\mathrm{SO}(3)$-Equivariant Transformer}

We adopt the $\mathrm{SO}(3)$-equivariant Transformer architecture for 3D geometries proposed by~\citet{frank2025sampling}, because of the similarities with the translation-invariant Transformer architecture we use for learning Hamiltonian Flow Maps (HFMs). As we employ the architecture as conservative machine learning force field (\gls{MLFF}), we remove the time-based conditioning branch from the model. The model uses equivariant formulations for activation functions~\citep{weiler20183d}, dense layers~\citep{unke2024e3x}, and layer normalizations~\citep{liao2024equiformerv2}. We give an overview in \cref{app:fig:mlff}.

\begin{figure}[tb]
    \centering
    \includegraphics[width=0.5\linewidth]{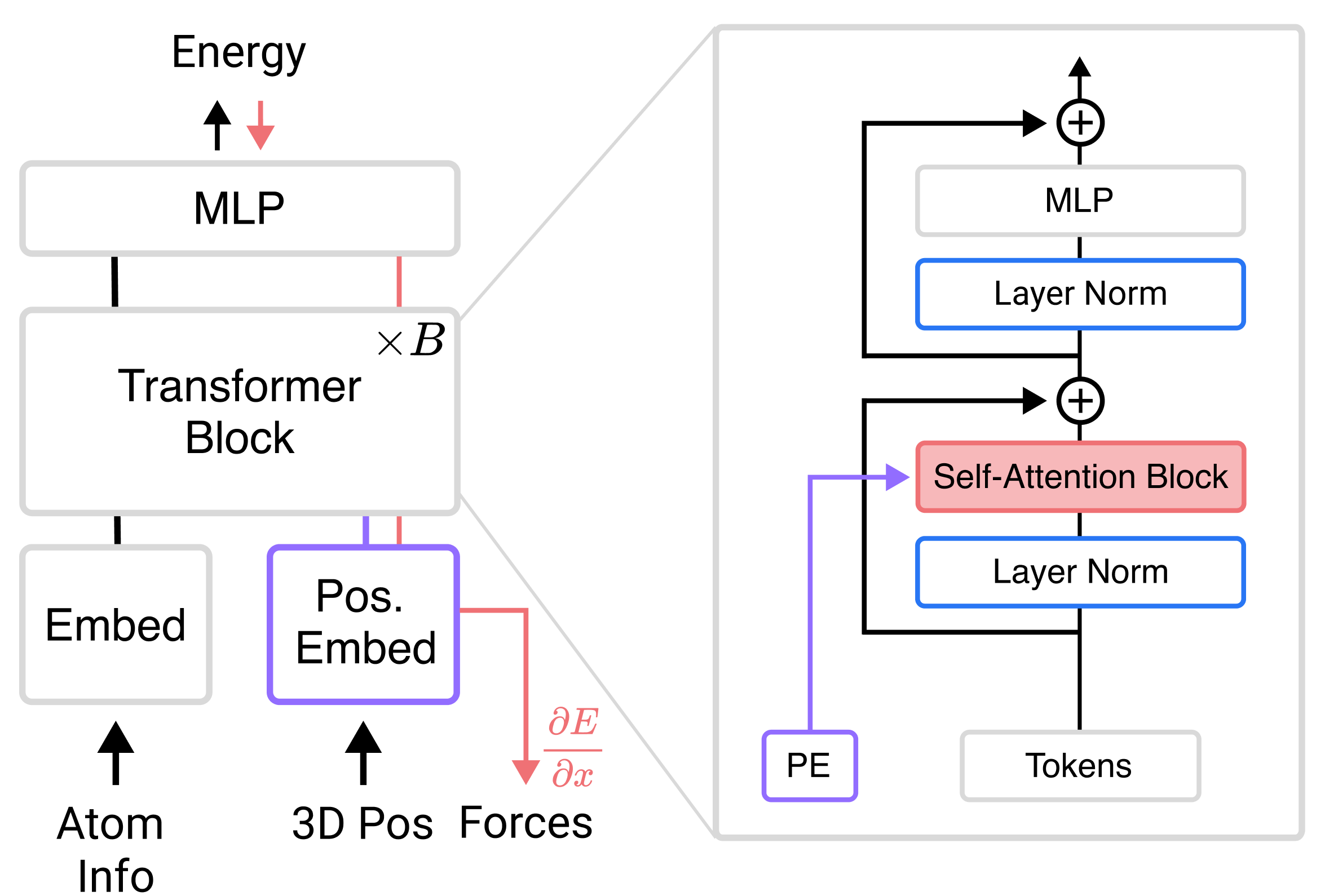}
    \caption{Equivariant \gls{MLFF} Transformer architecture adapted from \citet{frank2025sampling}. PE: Spherical harmonics embeddings are used for scaling attention coefficients in the transformer. We maintain rotational equivariant representations throughout the network.}
    \label{app:fig:mlff}
\end{figure}

Following the notation in~\citet{unke2024e3x}, the model operates on $\mathrm{SO}(3)$-equivariant tokens $\bh \in \mathbb{R}^{(L + 1)^2 \times H}$, where $L$ denotes the maximal degree of the spherical harmonics. We denote the features corresponding to the $\ell$-th degree as $\bh^{(\ell)} \in \mathbb{R}^{(2\ell + 1) \times H}$, where the $(2\ell + 1)$ entries corresponds to the orders $m = -\ell, \dots, +\ell$ per degree $\ell$. The maximum degree $L$ is chosen to ensure high fidelity at a reasonable computational cost. In this work, we use $L = 1$, similar to PaiNN~\citep{schutt2021equivariant} or TorchMDNet~\citep{tholke2021equivariant}, which restricts the representation to scalars and vectors. We refer interested readers to~\citet{duval2023hitchhiker} and~\citet{tang2025complete} for further discussion.

\paragraph{Token embeddings.} The initial token representation of the $i$-th particle is obtained as a concatenation of $L+1$ components

\begin{align}
    \bh_i = \bigoplus_{\ell = 0}^L \bh_i^{(\ell)} \in \mathbb{R}^{(L+1)^2 \times H}, \qquad
    \bh^{(\ell)}_i = 
    \begin{cases}
        \emb_z(z_i),& l=0,\\
        \boldsymbol{0}^{(\ell)},& l>0,
    \end{cases} 
    \in \mathbb{R}^{(2\ell + 1)\times H}.
\end{align}

Therefore, the features of degree $\ell = 0$ are given by a learnable embedding $\emb_z(z_i) \in \mathbb{R}^{1\times H}$ based on the particle type $z_i \in \mathbb{N}_+$ and the features of degree $\ell > 0$ are initialized with the zero matrix $\boldsymbol{0}^{(\ell)} \in \mathbb{R}^{(2l+1) \times H}$, i.e., all zeros.

\paragraph{Positional embeddings.} We construct $\mathrm{SO}(3)$-equivariant Euclidean positional embeddings (PE(3)) based on the distance $d_{ij} = \|\pos_i - \pos_j\|_2$ and normalized displacement vector $\boldsymbol{\hat{r}}_{ij} = (\pos_i - \pos_j) / \|\pos_i - \pos_j\|_2$ between particle $i$ and $j$ as a concatenation of $L+1$ components
\begin{equation}
\pe = \bigoplus_{\ell = 0}^L \boldsymbol{\phi}_\ell(d_{ij}) \, \odot \boldsymbol{Y}_\ell(\hat{r}_{ij}) \in \mathbb{R}^{(L+1)^2\times M_\text{RBF}}, \label{eq:so3-pe}
\end{equation}
where $\boldsymbol{\phi}_\ell: \mathbb{R} \mapsto \mathbb{R}^{1 \times M_\text{RBF}}$ is a radial basis function, $\boldsymbol{Y}_\ell \in \mathbb{R}^{(2\ell + 1) \times 1}$ are spherical harmonics of degree $\ell=0,\dots, L$, and $\odot$ is element-wise multiplication with implied broadcasting along the feature dimension s.t. $\boldsymbol{\phi}_\ell(d_{ij}) \, \odot \boldsymbol{Y}_\ell(\hat{r}_{ij}) \in \mathbb{R}^{(2\ell + 1) \times B}$. 

As discussed in~\citet{frank2025sampling}, these positional embeddings transform equivariantly under rotation of the input positions. Moreover, these positional embeddings are also invariant to translations as displacement vectors are used as inputs.
    
\paragraph{Attention mechanism.} In the following, we will make use of equivariant dense layers~\citep{unke2024e3x}, which act on a feature vector $\bh\in \mathbb{R}^{(L+1)^2\times H_{\text{in}}}$ per degree channel $\ell$ with $\bh^{(\ell)} \in \mathbb{R}^{(2\ell+1)\times H_{\text{in}}}$ as
\begin{align}
    \dense \bh=\bigoplus_{\ell = 0}^L \tilde{\bh}^{(\ell)},\qquad
    \tilde{\bh}^{(\ell)} = \begin{cases}
    \bh^{(\ell)}\dense_{(\ell)} + \mathbf{b}, & \ell = 0, \\
    \bh^{(\ell)}\dense_{(\ell)}, & \ell > 0,
    \end{cases}
\end{align}
where $\tilde{\bh}^{(\ell)}\in \mathbb{R}^{(2\ell+1)\times H_{\text{out}}}$ is a $\ell$-th degree feature component, $\dense_{(\ell)} \in \mathbb{R}^{H_{\text{in}} \times H_{\text{out}}}$ is a learnable weight matrix and $\mathbf{b} \in \mathbb{R}^{1\times H_{\text{out}}}$ is a bias applied only to the scalar channel. First, we define a pairwise gating mechanism $\boldsymbol{E} \in \mathbb{R}^{(L+1)^2\times H}$ as
\begin{align}
    \boldsymbol{E}_{ij}^{(\ell, m)} = \dense_K^E \pe^{(\ell = 0)}\quad \text{for all } \ell=0,\dots, L \text{ and } m = -\ell, \dots, \ell,
\end{align}
i.e., the scalar gate gets replicated across all $(\ell, m)$ channels. The self-attention update is then defined as
\begin{align}
    \bh_i &= \bh_i + \sum_{j=1}^N \softmax_j \left( \dfrac{\left \langle \dense_Q\bh_i, \dense_K \bh_j \odot \boldsymbol{E}_{ij} \right \rangle_F}{\sqrt{(L+1)^2H}} \right) \left( \dense_V \bh_j \otimes \dense_V^E\pe \right),
\end{align}
where $\dense_Q, \dense_K, \dense_V, \dense_V^E$ are learnable equivariant dense layers, $\langle \cdot \rangle_F$ is the Frobenius inner product, and $\otimes$ is the Clebsch-Gordan tensor product~\citep{thomas2018tensor,fuchs2020se,frank2022so3krates}. For ease of notation, we present the self-attention update with a single attention head, but apply multi-head attention with $n_{\text{heads}}$~\citep{vaswani2017attention}.

\paragraph{Prediction head.} Given the token representation $\bh_i^{[B]} \in \mathbb{R}^{(L+1)^2 \times H}$ for the $i$-th particle after $B$ transformer blocks, we predict energy contributions per particle based on the features $\bh^{[B](\ell = 0)}_i \in \mathbb{R}^{1\times H}$ of degree $\ell = 0$ using a two-layer MLP,
\begin{equation}
E_{i} = \MLP(\bh^{[B](\ell = 0)}_i) \in \mathbb{R}.
\end{equation}
Based on the predictions, we compute the potential energy of the system as the sum of energy contributions over all particles,
\begin{equation}
    E _{\text{pot}}= \sum_{i=1}^N E_i + E_{\text{rep}},
\end{equation}
where $E_{\text{rep}}$ denotes an empirical correction term inspired by the Ziegler–Biersack–Littmark (ZBL) potential~\citep{ziegler1985stopping}, which augments the energy prediction with physical knowledge about nuclear repulsion~\citep{unke2021spookynet}. Finally, the forces are computed as the negative gradient of the total potential energy with respect to the positions,
\begin{equation}
    \force_{i} = - \nabla_{\pos_i} E _{\text{pot}},
\end{equation}
using automatic differentiation in \texttt{jax}~\citep{bradburry2018jax}.
\section{Experimental Setup} \label{app:sec:experimental-setup}

\subsection{Single Particle Systems}
\label{app:sec:experimental-setup-single-particle}
The potential energies are defined as $\potential_{\text{s}}$ (Spring pendulum) and $\potential_{\text{b}}$ (Barbanis potential):
\begin{equation}
\potential_{\text{s}}(\mathbf{x}) = m g y + \frac{1}{2} k \left(\sqrt{x^2 + y^2} - l_0\right)^2
\label{eq:spring}
\end{equation}
\begin{equation}
\potential_{\text{b}}(\mathbf{x}) = \frac{1}{2} (\omega_x^2 x^2 + \omega_y^2 y^2) + \lambda x^2 y^2
\label{eq:barbanis}
\end{equation}
To simplify the training of our neural networks, we use a self-consistent dimensionless coordinate system without physical units for these experiments. For our experiments, we set $\lambda=10.0$ and $g=9.81$. We train a Hamiltonian Flow map in the conserved energy setting with $E_{tot}=1.5$ (Barbanis), $E_{tot}=-5$ (Spring pendulum). 

\paragraph{Dataset.}

We generate training data by uniformly sampling positions and rejecting them if the potential energy exceeds the desired total energy. We compute force labels using the ground truth potentials. Afterwards, we uniformly sample the direction of momenta and adjust the norm such that the total energy matches exactly. Note that all samples are drawn independently and there is no temporal or trajectory information in the datasets. We sample 100k decorrelated samples in total for each dataset.

\paragraph{Architecture.} We learn Hamiltonian Flow Maps (HFMs) using the Multi-Layer Perceptron-based network architecture with hidden dimensionality $H = 1024$.

\paragraph{Hyperparameters.}
Both HFM models are trained for 1000 epochs with batch size 512. For both single particle systems, we set $t_{\max}=2.5$ as the maximum timestep.

\paragraph{Simulation.}
We start simulations from random initial conditions with the correct total energy, generated via rejection sampling and adjustment of momenta magnitude as explained above. We do not use any filters for 1D systems. Ground truth trajectories are generated with Velocity Verlet integration of the true potential using a small timestep of $\t=0.01$.

\subsection{Gravitational N-Body System}
\label{app:sec:experimental-setup-gravity}

The potential is given as a gravitational interaction without boundary conditions: 
\begin{equation}
V_{\text{g}}(\mathbf{\pos}) = - \sum_{i}\sum_{i < j} \frac{G \mass_i \mass_j}{\lVert  \pos_i - \pos_j \rVert}
\label{eq:gravity}
\end{equation}

\paragraph{Dataset.}
The dataset creation is described in~\citet{brandstettergeometric}, where the ground truth data is generated using Velocity Verlet integration with small timesteps using natural units. The training set consists of 10.000 trajectories, and validation and test set of 2.000 trajectories each, while each trajectory contains 5.000 integration steps. For training with our loss, we disregard the time information in the dataset and train our model using decorrelated (randomly sampled) forces and geometries from the train dataset.

\paragraph{Architecture.}
We learn Hamiltonian Flow Maps (HFMs) using a translation-invariant Transformer architecture. Model hyperparameters are given in \cref{app:tab:gravity-architecture}. Since masses are all the same, we don't use mass embeddings. We use a larger cutoff (40) with more radial basis functions (64), since distances between interacting particles can be far.

\begin{table}[h!]
\centering
\caption{
Architectural details for the translation-invariant transformer on gravitational $N$-body system. $B$ is the number of blocks, $n_\text{heads}$ is the number of attention heads, $H$ is the hidden dimension, and $M_\text{RBF}$ and $M_\text{Gaussian}$ are the number of radial basis and Gaussian basis functions, respectively.
}
\vspace{3pt}
\begin{tabular}{lccccc}
\toprule
\textbf{Model} & $B$ & $n_\text{heads}$ & $H$ & $M_\text{RBF}$ & $M_\text{Gaussian}$ \\
\midrule
Transformer  & 6 & 8 & 256 & 64 & 8 \\
\bottomrule
\end{tabular}
\label{app:tab:gravity-architecture}
\end{table}

\paragraph{Hyperparameters.}
We set $\t_{\max}=0.1$ as the maximum timestep and train with batch size 64 for 300 epochs.

\paragraph{Simulation.}
We use all filters for our HFM model: drift removal, random rotation and coupled energy and angular momentum correction with the ground truth potential. Ground truth is generated with Velocity Verlet integration of the true potential using a small timestep of $\t=0.001$. The Velocity Verlet integrator in~\cref{fig:gravity} uses the ground truth analytic potential and therefore produces results that are indistinguishable from the ground truth for small timesteps.

\subsection{Small Molecules}
\label{app:sec:experimental-setup-small-mols}

\paragraph{Dataset.} We take four small organic molecules from the MD17 dataset~\cite{chmiela2017machine}, which is widely used as a force prediction benchmark for \glspl{MLFF}. We use the split from~\citet{fu2023forces}, i.e., 9,500 configurations for training, 500 configurations for validation, and 10,000 configurations for testing. We sample momenta $\mom$ from the Maxwell--Boltzmann distribution with mean $\mu_T = 500\,\mathrm{K}$ and standard deviation $\sigma_T = 150\,\mathrm{K}$ as described in \cref{app:implementation}.

\paragraph{Architecture.} We learn Hamiltonian Flow Maps (HFMs) using a translation-invariant Transformer architecture. Additionally, we employ an $\mathrm{SO}(3)$-equivariant Transformer as a MLFF baseline. Model hyperparameters are given in \cref{app:tab:architecture-small-molecules}.

\begin{table}[h!]
\centering
\caption{
Architectural details for the Translation-invariant and $\mathrm{SO}(3)$-equivariant Transformer on small molecules. $B$ is the number of blocks, $n_\text{heads}$ is the number of attention heads, $H$ is the hidden dimension, and $M_\text{RBF}$ and $M_\text{Gaussian}$ are the number of radial basis and Gaussian basis functions, respectively. Since the $\mathrm{SO}(3)$-Transformer does not have a conditioning branch, we write `--' for $M_\text{Gaussian}$.
}
\vspace{3pt}
\begin{tabular}{lccccc}
\toprule
\textbf{Model} & $B$ & $n_\text{heads}$ & $H$ & $M_\text{RBF}$ & $M_\text{Gaussian}$ \\
\midrule
Transformer  & 6 & 8 & 256 & 10 & 8 \\
$\mathrm{SO}(3)$-Transformer & 3 & 4 & 128 & 10 & \textemdash \\
\bottomrule
\end{tabular}
\label{app:tab:architecture-small-molecules}
\end{table}

\paragraph{Hyperparameters.}
 We use batch size of 500 and 50,000 epochs for the HFM models, corresponding to a total training time of up to 1d11h on a single NVIDIA A100 80BG GPU. We set $\t_{\max}=10$fs for all systems. For the $\mathrm{SO}$(3)-equivariant Transformer, we train for 5,000 epochs with a batch size of 50 corresponding to a total training time of up to 16h on a single NVIDIA A100 80BG GPU. For \gls{MLFF} training, we follow standard practice~\citep{batzner2022e3} and use as our training objective a weighted sum of energy and force loss terms, i.e., $\mathcal{L}=\lambda_E\mathcal{L}_E + \lambda_F\mathcal{L}_F$, and set $\lambda_E = 0.01$ and $\lambda_F=0.99$.

\paragraph{Simulation.}
We perform 300\,ps of \gls{MD} simulation in the NVT ensemble using a Langevin thermostat. We set the target temperature to $T = 500\,\mathrm{K}$ and apply a friction of $\gamma = 1/100\,\mathrm{fs}$ in all experiments. We run multiple simulations in parallel and report averaged quantities with standard deviations. Similar to training, we start the simulation with zero net linear momentum, and apply inference-time filters during simulation: drift removal, random rotation and coupled energy and angular momentum correction. We use our own trained \glspl{MLFF} as potentials for the energy correction filter. 

\paragraph{Evaluation.}
Following \citet{fu2023forces}, we calculate the interatomic distance distribution $h(r)$ averaged over frames from equilibrated trajectories and calculate the MAE w.r.t. the reference data.  Let $r$ be the distance, $N$ be the total number of particles, $i, j$ indicate the pairs of atoms that contribute to the distance statistics, and $\delta$ the Dirac Delta function to extract value distributions. For a given configuration $\pos$, the distribution of interatomic distances $h(r)$ is defined as

\begin{equation}
    h(r) = \dfrac{1}{N(N-1)} \sum_{i=1}^N \sum_{j=1, j \neq i}^N \delta \left( r - \|\pos_i-\pos_j\|_2 \right).
\end{equation}

Finally, we calculate the MAE as
\begin{equation}
    \text{MAE}_{h(r)} = \int_{r=0}^\infty \left| \langle h(r)\rangle - \langle \hat{h}(r)\rangle \right| dr,
\end{equation}
where $\langle \cdot \rangle$ denotes the average over frames of an equilibrated trajectories, i.e.,
$\langle h(r)\rangle := \frac{1}{T}\sum_{t=1}^T h_t(r)$ and $\langle \hat{h}(r)\rangle := \frac{1}{T}\sum_{t=1}^T \hat{h}_t(r)$, with $\langle h(r)\rangle$ computed from the reference and $\langle \hat{h}(r)\rangle$ computed from the generated trajectory by the model.

\subsection{Paracetamol}
\label{app:sec:experimental-setup-paracetamol}

\paragraph{Dataset.}
We adopt the MD17 dataset~\cite{chmiela2017machine} for Paracetamol with 106,490 samples. We use 80\% of the data ($\approx 85$k configurations) for training. We randomly select 256 training samples to simulate the low data regime. We sample momenta $\mom$ from the Maxwell--Boltzmann distribution with mean $\mu_T = 500\,\mathrm{K}$ and standard deviation $\sigma_T = 150\,\mathrm{K}$ as described in \cref{app:implementation}.

\paragraph{Architecture.} We learn Hamiltonian Flow Maps (HFMs) using a translation-invariant Transformer architecture. Additionally, we employ an $\mathrm{SO}(3)$-equivariant Transformer as an \gls{MLFF} baseline. Hyperparameters are given in \cref{app:tab:paracetamol-architecture}.

\begin{table}[h!]
\centering
\caption{
Architectural details for the Translation-invariant and $\mathrm{SO}(3)$-equivariant Transformer on paracetamol. $B$ is the number of blocks, $n_\text{heads}$ is the number of attention heads, $H$ is the hidden dimension, and $M_\text{RBF}$ and $M_\text{Gaussian}$ are the number of radial basis and Gaussian basis functions, respectively. Since the $\mathrm{SO}(3)$-Transformer does not have a conditioning branch, we write `--' for $M_\text{Gaussian}$.
}
\vspace{3pt}
\begin{tabular}{lccccc}
\toprule
\textbf{Model} & $B$ & $n_\text{heads}$ & $H$ & $M_\text{RBF}$ & $M_\text{Gaussian}$ \\
\midrule
Transformer  & 6 & 8 & 256 & 10 & 8 \\
$\mathrm{SO}(3)$-Transformer & 3 & 4 & 128 & 10 & \textemdash \\
\bottomrule
\end{tabular}
\label{app:tab:paracetamol-architecture}
\end{table}

\paragraph{Hyperparameters.}
We use a batch size of 128 and 2,000 epochs for the HFM models, resulting in a total training time of up to 22h on a single NVIDIA A100 80BG GPU. We use 640,000 epochs for the HFM model trained on 256 samples, amounting to 1d14h training time. In both cases, we set $\t_{\max}=10$fs. For the $\mathrm{SO}$(3)-equivariant Transformer, we train for 500 epochs with a batch size of 64, corresponding to a total training time of up to 13h. We train the \gls{MLFF} for 160,000 epochs with batch size 64, which amounts to 1d13h training time. Following~\citep{batzner2022e3}, we use as our training objective a weighted sum of energy and force loss terms, i.e., $\mathcal{L}=\lambda_E\mathcal{L}_E + \lambda_F\mathcal{L}_F$, and set $\lambda_E = 0.01$ and $\lambda_F=0.99$.

\paragraph{Simulation.}
We perform 3\,ns of \gls{MD} simulation in the NVT ensemble using a Langevin thermostat. We set the target temperature to $T = 500\,\mathrm{K}$ and apply a friction of $\gamma = 1/100\,\mathrm{fs}$ in all experiments. We also perform 3\,ns of \gls{MD} simulation in the NVE ensemble. Similar to training, we start the simulation with zero net linear momentum, and apply inference-time filters during simulation: drift removal, random rotation and coupled energy and angular momentum correction. We use our own trained \glspl{MLFF} as the potential for the energy correction. 

\paragraph{ScoreMD.} We benchmark against the recent diffusion model from \citet{plainer2025consistent}, which learns a diffusion process on Boltzmann-distributed samples, can generate iid samples via the standard denoising diffusion process, and also enables force estimation for simulation. To ensure a fair comparison, we use the graph transformer architecture from their work. We train the model with 128 hidden units, 3 layers, and 16-dimensional feature embeddings. For the dataset with 256 samples, we use a batch size of 256; for the dataset with $\approx 85$k samples, we use a batch size of 1024. To stabilize force estimation, we evaluate the network at diffusion time $t=0.05$. After training, we run Langevin simulations with the same setup as in our framework, using 25 parallel simulations to speed up sampling. We did not use any filters when evaluating ScoreMD.

\subsection{Alanine Dipeptide}
\label{app:sec:experimental-setup-alanine}

\paragraph{Dataset.} We use the dataset provided \citet{kohler2021smooth}, which contains data from a 1\,$\mu$s simulation in implicit solvent of which we use 80\% for training, and 10\% for validation. We sample momenta $\mom$ from the Maxwell--Boltzmann distribution with mean $\mu_T = 500\,\mathrm{K}$ and standard deviation $\sigma_T = 150\,\mathrm{K}$ as described in \cref{app:implementation}.

\paragraph{Architecture.} We learn Hamiltonian Flow Maps (HFMs) using a translation-invariant Transformer architecture. Model hyperparameters are given in \cref{app:tab:alanine-dipeptide-architecture}. In contrast to the other experiments, we do not train a separate \glspl{MLFF} for the coupled energy and angular momentum conservation filter, but instead just add an additional output head to predict the energy.

\begin{table}[h!]
\centering
\caption{
Architectural details for the translation-invariant transformer on alanine dipeptide. $B$ is the number of blocks, $n_\text{heads}$ is the number of attention heads, $H$ is the hidden dimension, and $M_\text{RBF}$ and $M_\text{Gaussian}$ are the number of radial basis and Gaussian basis functions, respectively.
}
\vspace{3pt}
\begin{tabular}{lccccc}
\toprule
\textbf{Model} & $B$ & $n_\text{heads}$ & $H$ & $M_\text{RBF}$ & $M_\text{Gaussian}$ \\
\midrule
Transformer  & 6 & 8 & 256 & 10 & 8 \\
\bottomrule
\end{tabular}
\label{app:tab:alanine-dipeptide-architecture}
\end{table}

\paragraph{Hyperparameters.}
We use a batch size of 512 and 2,000 epochs for the HFM models, resulting in a total training time of up to 3d12h on a single NVIDIA H100 80BG GPU, or 4d22h on an A100 80GB GPU. The energy loss weight is $0.01$.

\paragraph{Simulation.}
We perform 10 parallel simulations of 100\,ns of \gls{MD} simulation in the NVT ensemble using a Langevin thermostat, totalling to 1\,$\mu$s. We set the target temperature to $T = 300\,\mathrm{K}$ and apply a friction of $\gamma = 1/1000\,\mathrm{fs}$ in all experiments, which is the same as in the reference data \citep{kohler2021smooth}. Similar to training, we start the simulation with zero net linear momentum, and apply inference-time filters during simulation: drift removal, random rotation and coupled energy and angular momentum correction. We use the same HFM model with a separate output head to also predict the potential energy for the energy correction filter. 

\subsection{Ac-Ala3-NHMe}
\paragraph{Dataset.} We adopt the MD22 dataset~\cite{chmiela2023accurate} for Ac-Ala3-NHMe with 85,109 samples. We use 80\% of the data ($\approx 68$k configurations) for training. We sample momenta $\mom$ from the Maxwell--Boltzmann distribution with mean $\mu_T = 500\,\mathrm{K}$ and standard deviation $\sigma_T = 150\,\mathrm{K}$ as described in \cref{app:implementation}.

\paragraph{Architecture.} We learn Hamiltonian Flow Maps (HFMs) using a translation-invariant Transformer architecture. Additionally, we employ an $\mathrm{SO}(3)$-equivariant Transformer as an \gls{MLFF} baseline. Hyperparameters are given in \cref{app:tab:ala3-architecture}.

\begin{table}[h!]
\centering
\caption{
Architectural details for the Translation-invariant and $\mathrm{SO}(3)$-equivariant Transformer on Ac-Ala3-NHMe. $B$ is the number of blocks, $n_\text{heads}$ is the number of attention heads, $H$ is the hidden dimension, and $M_\text{RBF}$ and $M_\text{Gaussian}$ are the number of radial basis and Gaussian basis functions, respectively. Since the $\mathrm{SO}(3)$-Transformer does not have a conditioning branch, we write `--' for $M_\text{Gaussian}$.
}
\vspace{3pt}
\begin{tabular}{lccccc}
\toprule
\textbf{Model} & $B$ & $n_\text{heads}$ & $H$ & $M_\text{RBF}$ & $M_\text{Gaussian}$ \\
\midrule
Transformer  & 6 & 8 & 256 & 10 & 8 \\
$\mathrm{SO}(3)$-Transformer & 3 & 4 & 128 & 10 & \textemdash \\
\bottomrule
\end{tabular}
\label{app:tab:ala3-architecture}
\end{table}

\paragraph{Hyperparameters.}
We use a batch size of 128 and 5,000 epochs for the HFM model, resulting in a total training time of up to 40h on a single NVIDIA H100 80BG GPU. We set $\t_{\max}=10$fs. For the $\mathrm{SO}$(3)-equivariant Transformer, we train for 500 epochs with a batch size of 64, corresponding to a total training time of up to 10h on a single NVIDIA H100 80BG GPU. Following~\citep{batzner2022e3}, we use as our training objective a weighted sum of energy and force loss terms, i.e., $\mathcal{L}=\lambda_E\mathcal{L}_E + \lambda_F\mathcal{L}_F$, and set $\lambda_E = 0.01$ and $\lambda_F=0.99$.

\paragraph{Simulation.} We perform 3\,ns of \gls{MD} simulation in the NVT ensemble using a Langevin thermostat. We set the target temperature to $T = 500\,\mathrm{K}$ and apply a friction of $\gamma = 1/100\,\mathrm{fs}$ in all experiments. Similar to training, we start the simulation with zero net linear momentum, and apply inference-time filters during simulation: drift removal, random rotation and coupled energy and angular momentum correction. We use our own trained \glspl{MLFF} as the potential for the energy correction. 

\subsection{Chignolin and BBA}
\paragraph{Dataset.} In this setup, we use the pre-trained coarse-grained generative force field ScoreMD \citep{plainer2025consistent} and distill it. We take the data from \citet{deshaw20211fastfolding} and use the same training split as in \citet{plainer2025consistent}, reduce it to $C_\alpha$ atoms, and use ScoreMD to recompute coarse-grained force and energy labels for these training positions. As such, this allows us to train directly on coarse-grained samples where energy or forces are typically unavailable. Chignolin was simulated with a temperature of $340$K and BBA with $325$K.

\paragraph{Architecture.} We learn Hamiltonian Flow Maps (HFMs) using a translation-invariant Transformer architecture. Model hyperparameters are given in \cref{app:tab:cg-architecture}.
Similarly to the setup for alanine dipeptide, we do not train a separate \glspl{MLFF} for the coupled energy and angular momentum conservation filter, but instead just add output head to predict the energy.

\begin{table}[h!]
\centering
\caption{
Architectural details for the translation-invariant transformer and MLFF on Chignolin and BBA. $B$ is the number of blocks, $n_\text{heads}$ is the number of attention heads, $H$ is the hidden dimension, and $M_\text{RBF}$ and $M_\text{Gaussian}$ are the number of radial basis and Gaussian basis functions, respectively.
}
\vspace{3pt}
\begin{tabular}{llcccccc}
\toprule
\textbf{System} & \textbf{Model} & $B$ & $n_\text{heads}$ & $H$ & $M_\text{RBF}$ & $M_\text{Gaussian}$ & Cutoff in \AA{} \\
\midrule
Chignolin & Transformer  & 6 & 8 & 256 & 10 & 8 & 15\\
Chignolin & $\mathrm{SO}(3)$-Transformer & 3 & 4 & 128 & 10 & \textemdash & 15 \\
BBA & Transformer  & 6 & 8 & 256 & 10 & 8 & 20 \\
BBA & $\mathrm{SO}(3)$-Transformer & 3 & 4 & 128 & 10 & \textemdash & 20\\

\bottomrule
\end{tabular}
\label{app:tab:cg-architecture}
\end{table}

\paragraph{Hyperparameters.}
We use a batch size of 512 and 2,000 epochs for the Chignolin HFM models, resulting in a total training time of up to 1d on a single NVIDIA H100 80BG GPU. As for BBA, we use a batch size of 512 with 400 epochs and a training time of up to 1d4h on a single NVIDIA H100 80BG GPU.
\section{Additional Experiments and Ablations} \label{app:sec:ablations}
\subsection{Analysis of Conservation Laws in the Microcanonical (NVE) Ensemble} \label{app:subsec:nve-E-L-conservation}

\cref{app:fig:paracetamol_nve_trajcast} provides the temporal evolution for total energy and angular momentum corresponding to the experiment in~\cref{fig:paracetamol_nve_nstep}. Our setup with filters maintains the conservation of both quantities over long simulation runs (500\,ps).

Given identical starting conditions, our HFM model should theoretically trace the same phase-space trajectories as a reference \gls{MLFF} integrated via Velocity Verlet, up to a certain time limit related to the divergence in chaotic systems. We compare HFM and Velocity Verlet (using our \gls{MLFF}) trajectories starting from the same initial conditions for paracetamol in the NVE ensemble in~\cref{app:fig:paracetamol_nve_trajcast}. Following~\citet{thiemann2025forcefree}, we monitor selected interatomic distances and angles.

\begin{figure}[H]
    \centering
    \includegraphics[width=0.95\linewidth]{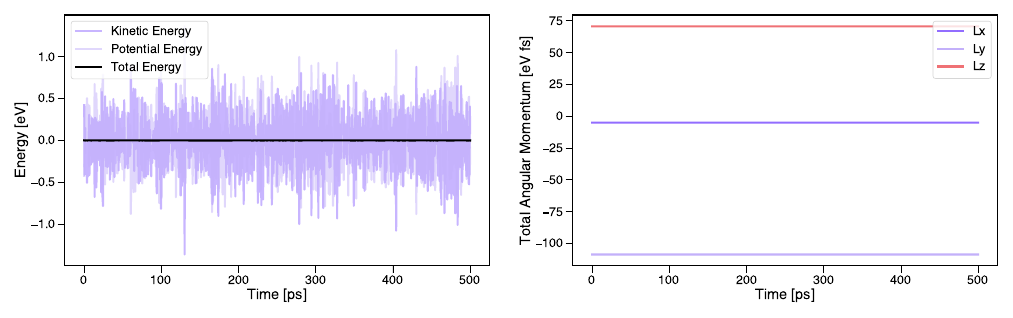}
    \caption{Conservation of total energy and total angular momentum in the NVE ensemble for paracetamol. \textbf{Left:} Temporal evolution of kinetic, potential and total energy over time. \textbf{Right:} Temporal evolution of angular momentum components over time.}
    \label{app:fig:paracetamol_nve_totalLE}
\end{figure}

\begin{figure}[H]
    \centering
    \includegraphics[width=.95\linewidth]{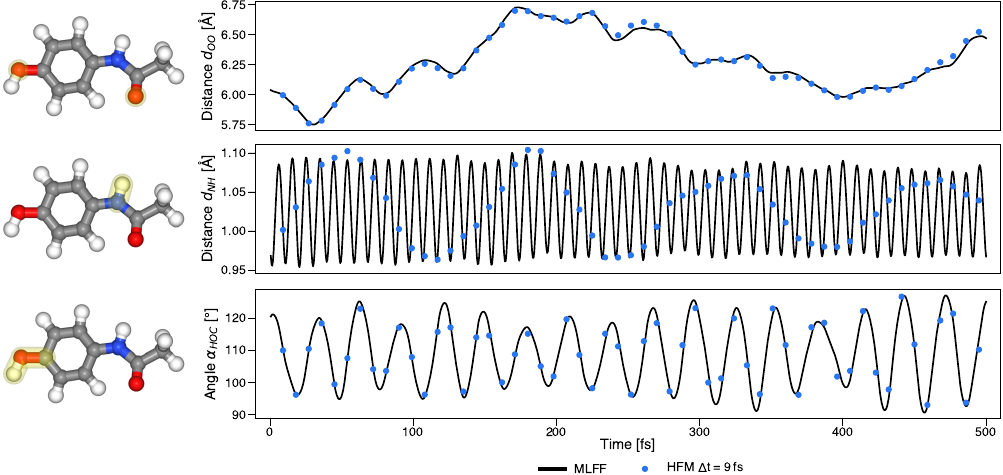}
    \caption{Short NVE simulations for paracetamol. We integrate our HFM model with $\t=9\,\mathrm{fs}$ steps in an NVE setting, and compare to integration using Velocity Verlet. The figure is inspired by~\citet{thiemann2025forcefree}. \textbf{First row:} oxygen-oxygen distance. \textbf{Second row:} nitrogen-hydrogen distance. \textbf{Third row:} hydroxyl group angle.}
    \label{app:fig:paracetamol_nve_trajcast}
\end{figure}

\subsection{Wall Clock Time Comparison for Simulation of Aspirin}
\label{app:subsec:wall-clock-time}

In~\cref{tab:timings}, we perform a wall clock time comparison between our model and \glspl{MLFF} for aspirin. We measure the time for a single inference step, and also use this information to compute the total simulation length in ns for a single GPU day. We evaluate the speed of our own \gls{MLFF} architecture, as well as two representative baselines from the literature: SchNet~\cite{schutt2018schnet}, a very fast and lightweight invariant \gls{MLFF} with limited expressivity, as well as NequIP~\cite{batzner2022e3}, a standard E(3)-equivariant \gls{MLFF}. We re-implement both baseline \glspl{MLFF} in JAX for a fair framework-independent runtime comparison. We use a single NVIDIA A100 80GB GPU for all timings, and report the average over 5000 measurements.

To ensure energy conservation, our inference pipeline includes an auxiliary \gls{MLFF} call to compute potential energy corrections. We must consider the cost for an additional \gls{MLFF} forward pass to compute the potential energy per simulation step. In this case, our own \gls{MLFF} is used for the total energy correction, while using the HFM model to predict the energy would also be possible. The timings for HFM in~\cref{tab:timings} already account for this fact: The 2.1ms total inference time decomposes into 0.9ms for the MLFF forward pass (note that we don't need an additional backward pass to compute the conservative forces, since we are only interested in the potential energy) and 1.2ms for the translation-invariant Transformer that predicts mean forces and velocities. Consequently, even with this overhead, HFM-9fs achieves a simulation throughput of 378.5 ns/day, significantly outperforming classical integrators, even if an extremely lightweight model like SchNet with limited expressivity is employed.

\begin{table}[h]
    \centering
        \caption{Time per forward pass (ms/step) and resulting simulation length (ns/day) for representative invariant~\cite{schutt2018schnet} and E(3)-equivariant~\cite{batzner2022e3} MLFFs and for our HFM with increasing integration timestep. MLFFs use $\t = 0.5\,\mathrm{fs}$. For fair comparison, all models are implemented in JAX. Timings are measured for aspirin on NVIDIA A100 80GB GPU.}
    \setlength{\tabcolsep}{3pt}
    \begin{tabular}{cccccccc}
    \toprule
    & SchNet & NequIP & Our MLFF & HFM-0.5fs & HFM-1fs & HFM-5fs & HFM-9fs \\
    \midrule
    $\frac{\mathrm{ms}}{\mathrm{step}}$ & 0.3 & 2.1 & 1.4 & 2.1 &  2.1 & 2.1 & 2.1 \\
    \midrule
    $\frac{\mathrm{ns}}{\mathrm{day}}$ & 148.9 & 20.6 & 31.7 & 21.0 & 42.1 & 210.3 & 378.5 \\
    \bottomrule
    \end{tabular}
    \label{tab:timings}
\end{table}

\subsection{Analysis of Force Error for Small Molecules}
\label{app:subsec:force-error}

In \cref{app:tab:force-mae}, we evaluate the instantaneous force prediction error of our model. It is important to note that HFM is trained to model integrated flow maps, rather than minimizing instantaneous force errors alone. Consequently, as the mixing parameter $q_{\tau=0}$ increases, the model prioritizes force matching over time consistency, resulting in a lower Force MAE. Compared to a dedicated \gls{MLFF} trained using force matching alone, our models still achieve decent Force MAE. Following~\citet{geng2025mean}, we use $q_{\t=0}$ = 0.75 in all our experiments to strike balance between force matching and time consistency.

\begin{table}[h]
\caption{Force MAE in units of [meV/\AA] for different $q_{\t=0}$. Sampling $\t=0$ more frequently yields better Force MAE. Setting $q_{\t=0}$ = 0.75 strikes a good balance between force matching and time consistency in our experiments.} \label{app:tab:force-mae}
\centering
\begin{tabular}{l|cccc}
\hline
Method & Aspirin & Ethanol & Naphthalene & Salicylic Acid \\
\hline
MLFF & 3.8 & 1.5 & 1.5 & 2.2 \\
HFM ($\t = 0$, $q_{\t=0}$ = 0.75) & 4.4 & 2.4 & 2.2 & 6.0 \\
HFM ($\t = 0$, $q_{\t=0}$ = 0.50) & 4.5 & 3.0 & 2.8 & 6.3 \\
HFM ($\t = 0$, $q_{\t=0}$ = 0.25) & 7.0 & 2.8 & 3.0 & 4.7 \\
\hline
\end{tabular}
\end{table}

\subsection{Analysis of Structural Observables for Other Thermostats}
\label{app:subsec:other-thermostats}

To asses the structural accuracy of our simulatios, we follow \citet{fu2023forces} and calculate the distribution $h(r)$ over interatomic distances $r$ and report the MAE w.r.t.\ \emph{ab-initio} reference data. We investigate the impact of thermostat choice within the NVT ensemble.

We observe that a global and deterministic Nosé-Hoover thermostat~\cite{martyna1996explicit} can lead to simulation artifacts in our framework causing larger deviations in MAE of $h(r)$ histograms (\cref{tab:mae_nose_hoover}). Similar effects have been observed in other long-timestep models~\citep{bigi2025flashmd} and are attributed to a failure of the thermostat to maintain kinetic energy equipartition across atom types. This is likely caused by non-conservative force predictions, as similar effects are observed using direct force prediction in \glspl{MLFF}~\citep{bigi2025dark}. Our results using the stochastic CSVR thermostat~\citep{bussi2007canonical} in \cref{tab:mae_csvr} are comparable to those using the Langevin thermostat~\citep{bussi2007accurate} in~\cref{tab:fane_langevin}.

Furthermore, during early experiments, we found that global momenta rescaling in global thermostats is generally less robust than local Langevin dynamics, as local stochasticity seems to be helpful for dampening model errors. Consequently, we avoid explicit energy rescaling in our filters when using global thermostats (CSVR or Nosé-Hoover in  \cref{tab:mae_csvr,tab:mae_nose_hoover}), as both mechanisms interfere with one another and degrade stability. Instead, we use an alternative filter that only preserves angular momentum.

\begin{table}[H]
\centering
\caption{Interatomic distances $h(r)$ MAE [unitless] for 300\,ps of MD simulation in the NVT ensemble using a \textbf{CSVR thermostat} w.r.t. the reference data from \emph{ab-initio} calculations. Results are averaged over 5 simulation runs with standard deviations shown in parentheses. Lower values indicate better performance.}
\begin{tabular}{l|c|cccccc}
\hline
Dataset & MLFF & \multicolumn{6}{c}{Hamiltonian Flow Map} \\
 & 0.5\,fs & 0.5\,fs & 1\,fs & 3\,fs & 5\,fs & 7\,fs & 9\,fs \\
\hline
Aspirin & 0.026 \scriptsize{(0.003)} & 0.099 \scriptsize{(0.003)} & 0.024 \scriptsize{(0.003)} & 0.050 \scriptsize{(0.002)} & 0.070 \scriptsize{(0.003)} & 0.070 \scriptsize{(0.005)} & 0.064 \scriptsize{(0.010)} \\
Ethanol & 0.074 \scriptsize{(0.005)} & 0.119 \scriptsize{(0.003)} & 0.057 \scriptsize{(0.006)} & 0.079 \scriptsize{(0.002)} & 0.113 \scriptsize{(0.004)} & 0.099 \scriptsize{(0.005)} & 0.125 \scriptsize{(0.001)} \\
Naphthalene & 0.043 \scriptsize{(0.005)} & 0.053 \scriptsize{(0.002)} & 0.037 \scriptsize{(0.001)} & 0.041 \scriptsize{(0.004)} & 0.040 \scriptsize{(0.001)} & 0.046 \scriptsize{(0.001)} & 0.051 \scriptsize{(0.001)} \\
Salicylic Acid & 0.038 \scriptsize{(0.003)} & 0.025 \scriptsize{(0.001)} & 0.021 \scriptsize{(0.001)} & 0.064 \scriptsize{(0.005)} & 0.073 \scriptsize{(0.004)} & 0.054 \scriptsize{(0.006)} & 0.059 \scriptsize{(0.003)} \\
\hline
\end{tabular}    
\label{tab:mae_csvr}
\end{table}

\begin{table}[H]
\centering
\caption{Interatomic distances $h(r)$ MAE [unitless] for 300\,ps of MD simulation in the NVT ensemble using a \textbf{\textit{Nosé-Hoover} thermostat} w.r.t. the reference data from \emph{ab-initio} calculations. Results are averaged over 5 simulation runs with standard deviations shown in parentheses. Lower values indicate better performance.}
\begin{tabular}{l|c|cccccc}
\hline
Dataset & MLFF & \multicolumn{6}{c}{Hamiltonian Flow Map} \\
 & 0.5\,fs & 0.5\,fs & 1\,fs & 3\,fs & 5\,fs & 7\,fs & 9\,fs \\
\hline
Aspirin & 0.020 \scriptsize{(0.003)} & 0.236 \scriptsize{(0.054)} & 0.221 \scriptsize{(0.008)} & 0.048 \scriptsize{(0.006)} & 0.088 \scriptsize{(0.014)} & 0.094 \scriptsize{(0.023)} & 0.272 \scriptsize{(0.002)} \\
Ethanol & 0.067 \scriptsize{(0.003)} & 0.716 \scriptsize{(0.017)} & 0.695 \scriptsize{(0.045)} & 0.057 \scriptsize{(0.002)} & 0.367 \scriptsize{(0.064)} & 0.246 \scriptsize{(0.153)} & 0.244 \scriptsize{(0.035)} \\
Naphthalene & 0.036 \scriptsize{(0.004)} & 0.085 \scriptsize{(0.003)} & 0.077 \scriptsize{(0.002)} & 0.176 \scriptsize{(0.043)} & 0.172 \scriptsize{(0.053)} & 0.121 \scriptsize{(0.011)} & 0.501 \scriptsize{(0.006)} \\
Salicylic Acid & 0.024 \scriptsize{(0.002)} & 0.159 \scriptsize{(0.062)} & 0.102 \scriptsize{(0.087)} & 0.210 \scriptsize{(0.020)} & 0.154 \scriptsize{(0.067)} & 0.058 \scriptsize{(0.004)} & 0.062 \scriptsize{(0.003)} \\
\hline
\end{tabular}
\label{tab:mae_nose_hoover}
\end{table}

\subsection{Analysis of Simulation Stability for Small Molecules}
\label{app:subsec:simulation-stability}

We further validate the robustness of HFM by assessing simulation stability within the NVT ensemble. We adopt the stability metric by~\citet{fu2023forces}. We compare runs from five different initial conditions and three thermostats. \Cref{tab:stab_langevin_app,tab:stab_csvr_app,tab:stab_nose_app} report the mean trajectory duration before collapse for Langevin, CSVR, and Nosé-Hoover thermostats. A mean of 300\,ps refers to no simulation collapse. The results confirm that HFM maintains high stability even for large timesteps.

\begin{table}[H]
\centering
\caption{Simulation stability under the \textbf{Langevin thermostat} with a filter for explicit coupled conservation of total energy and angular momentum. We report the mean stable trajectory duration in ps (max 300). Values are presented as: Mean (Standard Deviation) [Number of collapsed runs].}
\begin{tabular}{l|c|ccccc}
\hline
Dataset & MLFF & \multicolumn{5}{c}{Hamiltonian Flow Map} \\
 & 0.5\,fs & 1\,fs & 3\,fs & 5\,fs & 7\,fs & 9\,fs \\
\hline
Aspirin & 300.0 \scriptsize{(000.0) [0]} & 300.0 \scriptsize{(000.0) [0]} & 300.0 \scriptsize{(000.0) [0]} & 300.0 \scriptsize{(000.0) [0]} & 300.0 \scriptsize{(000.0) [0]} & 300.0 \scriptsize{(000.0) [0]} \\
Ethanol & 300.0 \scriptsize{(000.0) [0]} & 300.0 \scriptsize{(000.0) [0]} & 300.0 \scriptsize{(000.0) [0]} & 300.0 \scriptsize{(000.0) [0]} & 300.0 \scriptsize{(000.0) [0]} & 300.0 \scriptsize{(000.0) [0]} \\
Naphthalene & 300.0 \scriptsize{(000.0) [0]} & 300.0 \scriptsize{(000.0) [0]} & 300.0 \scriptsize{(000.0) [0]} & 300.0 \scriptsize{(000.0) [0]} & 300.0 \scriptsize{(000.0) [0]} & 300.0 \scriptsize{(000.0) [0]} \\
Salicylic Acid & 300.0 \scriptsize{(000.0) [0]} & 300.0 \scriptsize{(000.0) [0]} & 300.0 \scriptsize{(000.0) [0]} & 254.1 \scriptsize{(091.9) [1]} & 200.5 \scriptsize{(039.6) [5]} & 196.7 \scriptsize{(127.7) [2]} \\
\hline
\end{tabular}
\label{tab:stab_langevin_app}
\end{table}

\begin{table}[H]
\centering
\caption{Simulation stability under the \textbf{CSVR thermostat} without explicit energy correction filter. We report the mean stable trajectory duration in ps (max 300). Values are presented as: Mean (Standard Deviation) [Number of collapsed runs].}
\begin{tabular}{l|c|ccccc}
\hline
Dataset & MLFF & \multicolumn{5}{c}{Hamiltonian Flow Map} \\
 & 0.5\,fs & 1\,fs & 3\,fs & 5\,fs & 7\,fs & 9\,fs \\
\hline
Aspirin & 300.0 \scriptsize{(000.0) [0]} & 300.0 \scriptsize{(000.0) [0]} & 300.0 \scriptsize{(000.0) [0]} & 300.0 \scriptsize{(000.0) [0]} & 300.0 \scriptsize{(000.0) [0]} & 300.0 \scriptsize{(000.0) [0]} \\
Ethanol & 300.0 \scriptsize{(000.0) [0]} & 300.0 \scriptsize{(000.0) [0]} & 300.0 \scriptsize{(000.0) [0]} & 300.0 \scriptsize{(000.0) [0]} & 300.0 \scriptsize{(000.0) [0]} & 300.0 \scriptsize{(000.0) [0]} \\
Naphthalene & 300.0 \scriptsize{(000.0) [0]} & 300.0 \scriptsize{(000.0) [0]} & 300.0 \scriptsize{(000.0) [0]} & 300.0 \scriptsize{(000.0) [0]} & 300.0 \scriptsize{(000.0) [0]} & 300.0 \scriptsize{(000.0) [0]} \\
Salicylic Acid & 300.0 \scriptsize{(000.0) [0]} & 240.4 \scriptsize{(077.2) [2]} & 300.0 \scriptsize{(000.0) [0]} & 300.0 \scriptsize{(000.0) [0]} & 260.1 \scriptsize{(079.9) [1]} & 221.3 \scriptsize{(102.7) [2]} \\
\hline
\end{tabular}
\label{tab:stab_csvr_app}
\end{table}

\begin{table}[H]
\centering
\caption{Simulation stability under the \textbf{Nosé-Hoover thermostat} without explicit energy correction filter. We report the mean stable trajectory duration in ps (max 300). Values are presented as: Mean (Standard Deviation) [Number of collapsed runs].}

\begin{tabular}{l|c|ccccc}
\toprule
Dataset & MLFF & \multicolumn{5}{c}{Hamiltonian Flow Map} \\
 & 0.5\,fs & 1\,fs & 3\,fs & 5\,fs & 7\,fs & 9\,fs \\
\midrule
Aspirin & 300.0 \scriptsize{(000.0) [0]} & 300.0 \scriptsize{(000.0) [0]} & 300.0 \scriptsize{(000.0) [0]} & 300.0 \scriptsize{(000.0) [0]} & 300.0 \scriptsize{(000.0) [0]} & 271.4 \scriptsize{(057.3) [1]} \\
Ethanol & 300.0 \scriptsize{(000.0) [0]} & 300.0 \scriptsize{(000.0) [0]} & 300.0 \scriptsize{(000.0) [0]} & 300.0 \scriptsize{(000.0) [0]} & 300.0 \scriptsize{(000.0) [0]} & 300.0 \scriptsize{(000.0) [0]} \\
Naphthalene & 300.0 \scriptsize{(000.0) [0]} & 294.1 \scriptsize{(011.8) [1]} & 300.0 \scriptsize{(000.0) [0]} & 300.0 \scriptsize{(000.0) [0]} & 300.0 \scriptsize{(000.0) [0]} & 300.0 \scriptsize{(000.0) [0]} \\
Salicylic Acid & 300.0 \scriptsize{(000.0) [0]} & 178.9 \scriptsize{(105.4) [3]} & 300.0 \scriptsize{(000.0) [0]} & 300.0 \scriptsize{(000.0) [0]} & 287.3 \scriptsize{(025.3) [1]} & 300.0 \scriptsize{(000.0) [0]} \\
\bottomrule
\end{tabular}

\label{tab:stab_nose_app}
\end{table}

\subsection{Ablation for $\tau$ Sampling Distribution on Small Molecules} 
\label{app:sec:tau_sampling}

In this subsection, we ablate how different base distributions $q(\tau)$ for the timestep $\t$ affect the integration accuracy on several small molecules. In this work, we experiment with three different choices of $q(\tau)$ (see \cref{app:implementation}). Therefore, for each molecule, we train three models that only differ in $q(\tau)$, while keeping all other hyperparameters fixed.

To quantify the model's ability to recover the correct Hamiltonian flow, we compute the Root Mean Square Deviation (RMSD) between the final state ($\pos_{\text{HFM}}$, $\mom_{\text{HFM}}$) predicted by our HFM model in a single step and a reference state ($\pos_{\text{ref}}$, $\mom_{\text{ref}}$) obtained via integrating the baseline \gls{MLFF} with Velocity Verlet over the same time interval $\t$. To account for the inherent diffusivity of the system, where particles naturally drift further apart over longer timescales, we normalize this prediction error by the accumulated RMSD of the reference trajectory. The normalized metric for positions at timestep $t$ is defined as: 

\begin{equation} 
\text{Normalized RMSD}_{\boldsymbol{x}}(t) = \frac{\text{RMSD}(\boldsymbol{x}_{\text{HFM}}(t), \boldsymbol{x}_{\text{ref}}(t))}{\sum_{k=0}^{N-1} \text{RMSD}(\boldsymbol{x}_{\text{ref}}(\frac{(k+1)t}{N}), \boldsymbol{x}_{\text{ref}}(\frac{kt}{N}))}, 
\end{equation} 

where the denominator approximates the path length of the ground truth trajectory. We employ the same formula for momenta as well. This normalization ensures that errors at larger $\t$ are not inflated due to larger total displacements, providing a fair comparison of integral error across different timescales. 

The results in \cref{app:fig:integral-t-sampling} indicate that our proposed mixture distribution allows for consistently low integration error within the interval $[0, \t_{\max}]$ on all systems. In contrast, the uniform distribution has consistently low integration error on aspirin and naphthalene, but higher integration error on ethanol and salicylic acid. We argue that this is due to instabilities we have observed while using the uniform distribution during training, which underlines that the choice of $q(\tau)$ has a large impact on the structure of the loss landscape~\citep{zhang2025alphaflow}. Finally, the logit-normal distribution yields low integration error for small $\t$, but the integration error increases significantly as $\t \rightarrow \t_{\max}$, since these timesteps were rarely sampled during training. Hence, we use our mixture distribution in all our experiments.

\begin{figure}[H]
    \centering
    \includegraphics[width=.98\linewidth]{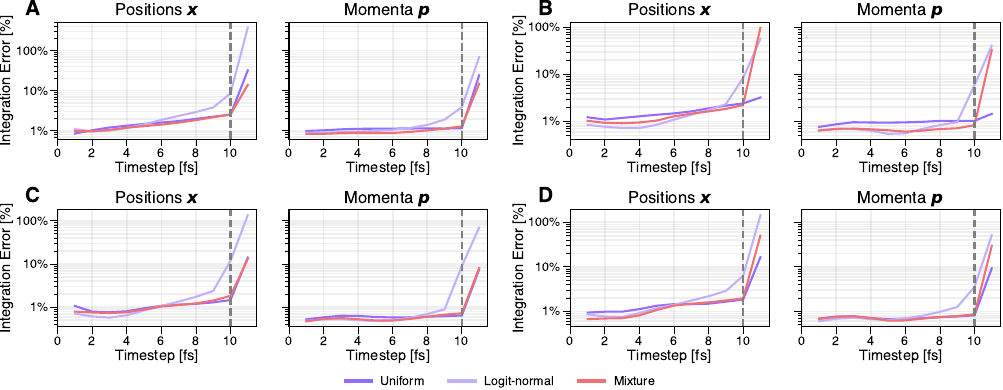}
    \caption{We show the normalized integration error (RMSD) for different sampling distributions $q(\tau)$ on small molecules. (\textbf{A}) Aspirin.
    (\textbf{B}) Ethanol. (\textbf{C}) Naphthalene. (\textbf{D}) Salicylic Acid. The dashed vertical line at $\t_{\max}=10\,\mathrm{fs}$ marks the training limit.}
    \label{app:fig:integral-t-sampling}
\end{figure}

\subsection{Analysis of Training Dynamics on Paracetamol}

In \cref{app:fig:integral-checkpoint}, we analyze the convergence of the HFM model by monitoring the trajectory-level deviation from Velocity Verlet integration across different training checkpoints using the metric from~\cref{app:sec:tau_sampling}. We observe a systematic reduction in normalized integration error as training progresses, indicating that the model successfully refines its approximation of the underlying Hamiltonian flow. Interestingly, the integration error decreases for all timesteps over the course of the training, hinting at the fact that the model learns time consistency and force accuracy simultaneously and not separately.

\begin{figure}[H]
    \centering
    \includegraphics[width=.99\linewidth]{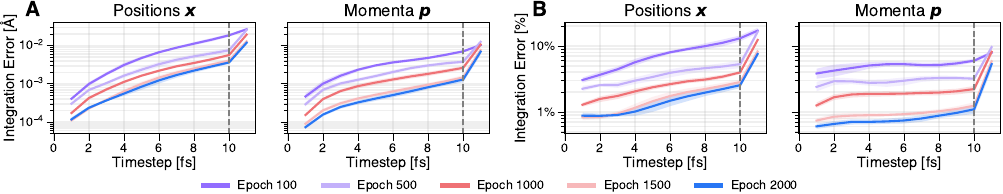}
    \caption{Evolution of integration error during training on paracetmol. (\textbf{A}) Absolute integration error.
    (\textbf{B}) Normalized integration error. The dashed vertical line at $\t_{\max}=10\,\mathrm{fs}$ marks the training limit.}
    \label{app:fig:integral-checkpoint}
\end{figure}

Furthermore, we investigate the impact of the training horizon $\t_{\max}$ in \cref{app:fig:integral-t-max}. The results highlight a trade-off between long-range stability and short-range precision: while a larger $\t_{\max}$ enhances the model's predictive capability at large timesteps, it possibly leads to a degradation in accuracy for smaller timesteps (see \cref{app:fig:integral-t-max}, right panel around 5fs).

\begin{figure}[H]
    \centering
    \includegraphics[width=.99\linewidth]{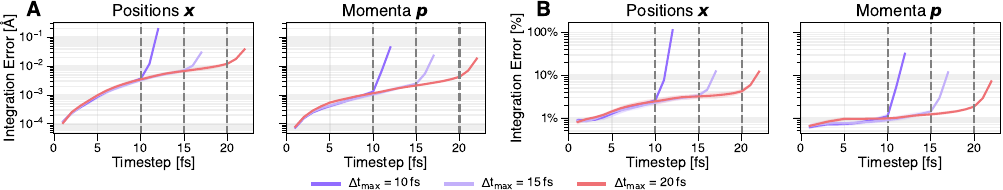}
    \caption{Ablation of maximum training timestep $\t_{\max}$ on paracetamol. (\textbf{A}) Absolute integration error.
    (\textbf{B}) Normalized integration error. The dashed vertical line at $\t_{\max}=10\,\mathrm{fs}$ marks the training limit.}
    \label{app:fig:integral-t-max}
\end{figure}

\subsection{Analysis of Time Consistency on Paracetamol}

In \cref{app:fig:trajcetory-consistency}, we evaluate the internal consistency of the learned Hamiltonian Flow Maps. A physically valid time-evolution operator must satisfy the semi-group property of the flow \citep{boffi2025build}. In our notation, where $u_{\Delta t}$ denotes the flow map advancing the system state $z_t = (x_t, p_t)$ by time $\Delta t$, this condition implies that a single forward step should be equivalent to two sequential half-steps:
\begin{equation}
u_{\Delta t}(z_t) \approx u_{\Delta t/2} \big( u_{\Delta t/2}(z_t) \big).
\end{equation}

To quantify violations of the semi-group property, we compute the consistency error as the RMSD between the state obtained from one full step $\Delta t$ and the state obtained from two composed steps of $\Delta t/2$. Analogous to the integration error analysis in~\cref{app:sec:tau_sampling}, we report both the absolute RMSD and the normalized RMSD, where the latter is scaled by the accumulated displacement of the composed trajectory to account for the natural scale of motion.

\cref{app:fig:trajcetory-consistency} displays these metrics for an HFM trained with $\Delta t_{\max}=10\,\mathrm{fs}$. As a reference, the dashed grey line shows the consistency error of a standard Velocity Verlet integrator based on our baseline \gls{MLFF}, comparing a single 0.5\,fs step against two 0.25\,fs steps. The results demonstrate that our model maintains a comparable degree of internal consistency, up to around 4\,fs steps. However, as shown in the other experiments, larger timesteps can still be used effectively in simulations.
 
\begin{figure}[H]
    \centering
    \includegraphics[width=.99\linewidth]{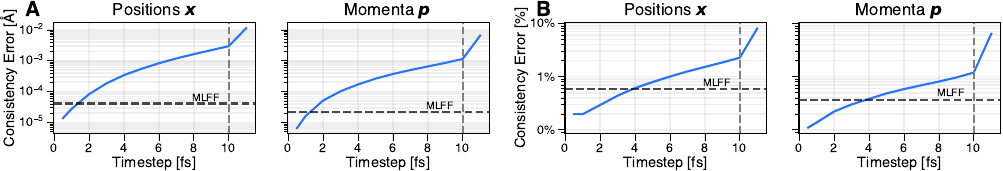}
    \caption{Analysis of time consistency on Paracetamol. We measure the deviation between a single predicted step of size $\Delta t$ and two composed steps of size $\Delta t / 2$. (\textbf{A}) Absolute consistency RMSD error.
    (\textbf{B}) Normalized consistency RMSD error. The dashed vertical line at $\t_{\max}=10\,\mathrm{fs}$ marks the training limit. The dashed horizontal line indicates the consistency error of Velocity Verlet with $0.5\,\mathrm{fs}$ steps.}
    \label{app:fig:trajcetory-consistency}
\end{figure}

\subsection{Analysis of Vibrational Density of States on Paracetamol}
In \cref{app:fig:paracetamol_spectra}, we perform a granular analysis of the vibrational density of states to assess the spectral fidelity of our HFM model for different training regimes. We employ the HFM trained with $\Delta t_{\max}=10\,\mathrm{fs}$ and evaluated in simulation for $\t=9\,\mathrm{fs}$ consistent with~\cref{fig:paracetamol_nvt_training_data} in the main text. The plots display the per-atom power spectra, computed by taking the Fourier transform of the velocity autocorrelation function for each atom type in the paracetamol molecule. This allows us to disentangle the contributions of different chemical elements (C, H, N, O) to the vibrational dynamics.

We compare the spectra generated by HFM (blue) against those obtained from integrating a reference MLFF baseline with Velocity Verlet (black) for different training set sizes ($\approx85k$ or $256$ training samples). The results demonstrate that our model recovers the correct frequency modes even in the data-scarce regime, where we train on $256$ decorrelated samples.

\begin{figure}[H]
    \centering
    \includegraphics{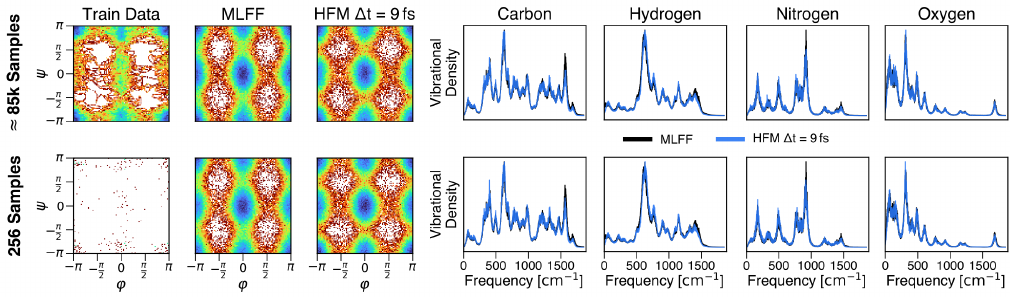}
    \caption{Data efficiency and spectral fidelity in NVT simulations (3 ns) for paracetamol. We compare the per-atom vibrational spectra of paracetamol generated by integrating the reference MLFF with Velocity Verlet (black) and our HFM model with $\Delta t=9\,\mathrm{fs}$ (blue). Rows correspond to different training set sizes.}
    \label{app:fig:paracetamol_spectra}
\end{figure}

\subsection{Analysis of Temperature Robustness on Paracetamol}

In \cref{app:fig:paracetamol_spectra_temp}, we test the temperature robustness of a single HFM model on paracetamol. Specifically, we report results for $T\in \{300,350,400,450,550,600,650,700\}$\,K, and compare the HFM model evaluated at $\Delta t=9\,\mathrm{fs}$ with our MLFF baseline. Across these temperatures, our HFM model closely matches the distribution of dihedral angles $(\varphi, \psi)$ and vibrational spectra obtained by the MLFF baseline. These results suggest high temperature robustness of our HFM model.

\subsection{Ablation: Effect of Inference Filters on Observables} \label{app:subsec:filters-fix-stuff}
Our inference pipeline employs projection filters to strictly enforce the conservation of total angular momentum and total energy. Here, we analyze the impact of these filters on thermodynamic consistency and dynamic as well as static observables. We further investigate whether using conservation filters might artificially stabilize invalid trajectories, e.g., those generated by Velocity Verlet integration with larger timesteps.

\begin{figure}[H]
    \centering
    \includegraphics{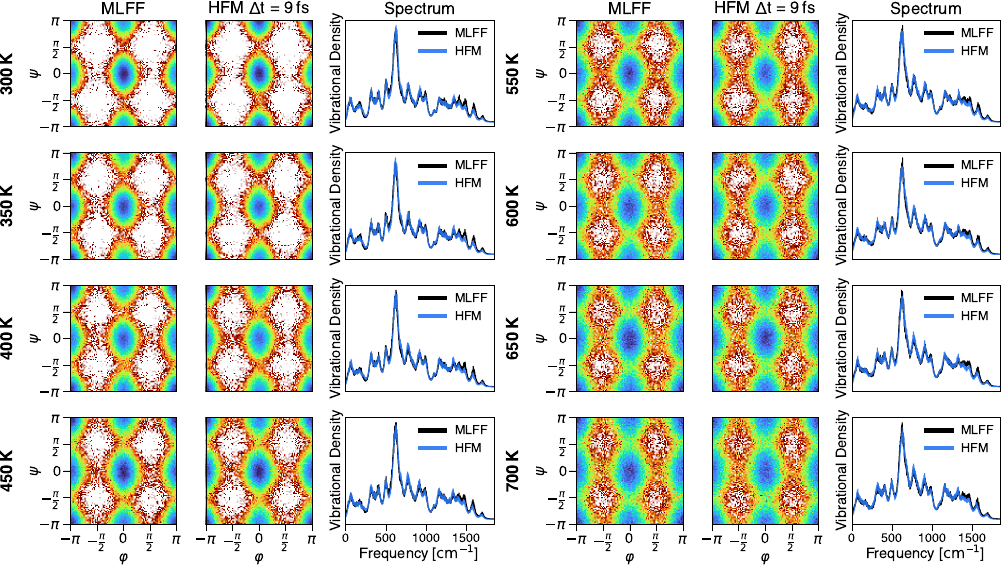}
    \caption{Temperature robustness in NVT simulations (3 ns) for paracetamol. We compare the ramachandran plot over dihedral angles $\varphi, \psi$ and vibrational spectra of paracetamol generated by integrating the reference MLFF with Velocity Verlet (black) and our HFM model with $\Delta t=9\,\mathrm{fs}$ (blue). We report results for a wide range of temperatures $T\in \{300,350,400,450,550,600,650,700\}$\,K.
    }
    \label{app:fig:paracetamol_spectra_temp}
\end{figure}

First, we examine element-wise kinetic energy distributions, using the Langevin thermostat to simulate paracetamol in the NVT ensemble. In \cref{app:tab:kinetic-energy-deviation}, we report the deviation of the per-element kinetic temperature from the target thermostat temperature. Without conservation filters (left), we observe significant deviations, particularly for hydrogen atoms at small timesteps, indicating a failure to maintain equipartition. Enabling the conservation filters (right) restores thermodynamic consistency, keeping temperature deviations minimal across all atom types and timesteps.

\begin{table}[h]
\centering
\caption{Kinetic energy deviation ($\Delta T$) from the target thermostat temperature for paracetamol simulations using the Langevin thermostat. We compare integration without (left) and with (right) energy and angular momentum conservation filters. Values are reported as mean (standard deviation) in Kelvin. The filters effectively restore equipartition.}
\label{app:tab:kinetic-energy-deviation}
\resizebox{\linewidth}{!}{%
\begin{tabular}{l|ccccc|ccccc}
\hline
\multicolumn{1}{c|}{Timestep [fs]} & \multicolumn{5}{c|}{Without energy / angular momentum conservation} & \multicolumn{5}{c}{With energy / angular momentum conservation} \\
 & $\Delta T_{\mathrm{all}}$ & $\Delta T_{\mathrm{H}}$ & $\Delta T_{\mathrm{C}}$ & $\Delta T_{\mathrm{N}}$ & $\Delta T_{\mathrm{O}}$ & $\Delta T_{\mathrm{all}}$ & $\Delta T_{\mathrm{H}}$ & $\Delta T_{\mathrm{C}}$ & $\Delta T_{\mathrm{N}}$ & $\Delta T_{\mathrm{O}}$ \\
\hline
$\Delta t = 1$ & -13.7 \scriptsize{(0.0)} & -22.6 \scriptsize{(0.0)} & -6.1 \scriptsize{(0.0)} & -10.0 \scriptsize{(0.0)} & -6.0 \scriptsize{(0.0)} & -0.8 \scriptsize{(0.0)} & -8.8 \scriptsize{(0.0)} & 6.2 \scriptsize{(0.0)} & 3.2 \scriptsize{(0.0)} & 4.9 \scriptsize{(0.0)} \\
$\Delta t = 3$ & -2.1 \scriptsize{(0.2)} & -3.6 \scriptsize{(0.2)} & -0.7 \scriptsize{(0.5)} & -1.7 \scriptsize{(1.0)} & -1.1 \scriptsize{(0.4)} & -0.7 \scriptsize{(0.1)} & -2.3 \scriptsize{(0.3)} & 0.8 \scriptsize{(0.3)} & -0.5 \scriptsize{(0.8)} & 0.3 \scriptsize{(0.4)} \\
$\Delta t = 5$ & 1.9 \scriptsize{(0.4)} & 0.9 \scriptsize{(0.6)} & 3.6 \scriptsize{(0.5)} & 1.6 \scriptsize{(1.5)} & -0.4 \scriptsize{(0.8)} & -0.0 \scriptsize{(0.4)} & -0.8 \scriptsize{(0.6)} & 1.4 \scriptsize{(0.5)} & -0.2 \scriptsize{(1.2)} & -2.4 \scriptsize{(1.2)} \\
$\Delta t = 7$ & 1.3 \scriptsize{(0.5)} & -0.3 \scriptsize{(0.7)} & 2.9 \scriptsize{(0.4)} & 5.1 \scriptsize{(0.8)} & -0.3 \scriptsize{(1.5)} & 0.5 \scriptsize{(0.3)} & -1.2 \scriptsize{(0.4)} & 2.2 \scriptsize{(0.4)} & 3.7 \scriptsize{(1.4)} & -0.3 \scriptsize{(0.6)} \\
$\Delta t = 9$ & 2.6 \scriptsize{(0.2)} & 1.4 \scriptsize{(0.6)} & 4.0 \scriptsize{(0.5)} & 6.9 \scriptsize{(0.8)} & -0.2 \scriptsize{(1.0)} & 0.4 \scriptsize{(0.4)} & -1.0 \scriptsize{(0.5)} & 2.0 \scriptsize{(0.5)} & 4.4 \scriptsize{(1.6)} & -2.0 \scriptsize{(0.9)} \\
\hline
\end{tabular}
}
\end{table}

Second, we assess whether the structural dynamics predicted by our model rely on these filters. \Cref{app:fig:paracetamol_spectra_no_cc} displays the statistics and vibrational spectra generated without explicit energy or angular momentum conservation. The spectra remain well aligned with the ground truth \gls{MLFF} results (see \cref{app:fig:paracetamol_spectra} for comparison). This suggests that the HFM model learns valid dynamics intrinsic to the Hamiltonian flow, while the filters serve to correct thermodynamic drifts.

\begin{figure}[h]
    \centering
    \includegraphics{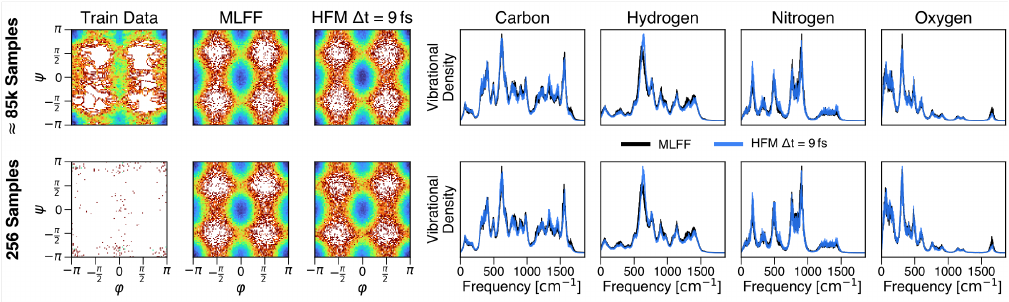}
    \caption{Robustness of vibrational spectra \textbf{without filters}. We show the per-atom vibrational density of states for paracetamol ($\Delta t=9\,\mathrm{fs}$) generated \textbf{without} energy and angular momentum conservation filter. Comparing this to \cref{app:fig:paracetamol_spectra} confirms that the structural dynamics are robustly captured by the model even in the absence of explicit constraints.}
    \label{app:fig:paracetamol_spectra_no_cc}
\end{figure}

\begin{figure}[H]
    \centering
    \includegraphics[width=.6\linewidth]{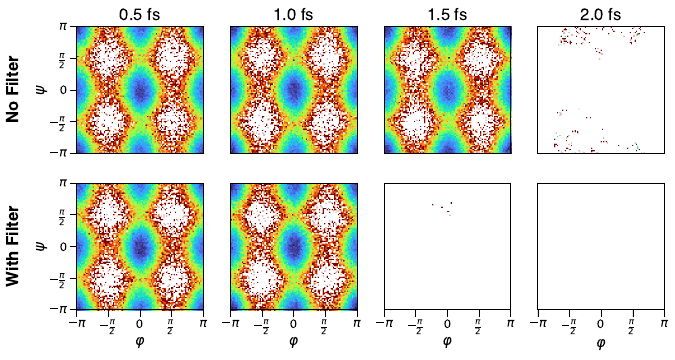}
    \caption{Effect of inference filters on Velocity Verlet stability: We attempt to stabilize integration of our \gls{MLFF} using Velocity Verlet using inference filters. \textbf{Top}: The unfiltered baseline becomes unstable between timesteps of $1.5$ and $2.0\,\mathrm{fs}$. \textbf{Bottom}: Applying filters does not rescue the simulation but rather causes earlier divergence (for smaller timestep), confirming that filters do not in general artificially stabilize incorrect integration steps.}
    \label{app:fig:paracetamol_vv_check}
\end{figure}

Finally, we address the question whether our filters might stabilize arbitrary invalid trajectories and thereby enable larger timesteps. To test this, we apply our filters to a standard Velocity Verlet integration of the reference \gls{MLFF} with different larger timesteps. As shown in \cref{app:fig:paracetamol_vv_check}, the filters do not enable larger timesteps for Velocity Verlet. In contrast, applying rigid energy constraints to the erroneous steps of Velocity Verlet causes the simulation to diverge earlier (instability onset $< 1.5\,\mathrm{fs}$) than for the unfiltered baseline. This confirms that the large-timestep stability of HFM stems from the learned flow map itself, not from the post-hoc application of inference filters.

\subsection{Ablation: Effect of Integration Timestep on Free Energy Surface for Alanine Dipeptide}
\label{app:subsec:ablations-aldp}

\paragraph{Free energy surfaces.} 
In this section, we complement the main paper and report additional Ramachandran plots for an HFM trained with our objective up to $\t_{\mathrm{max}}$ and evaluated at different timesteps $\t$. The corresponding plots are shown in \cref{app:fig:aldp_ramachandran}, with numerical results in \cref{app:tab:aldp-pmf-js}. Overall, models evaluated up to $\t=13$\,fs recover the low-probability mode and match the reference distribution well, except for $\t\approx 10$\,fs, where we observe a systematic deviation. Since this behavior is consistent across hyperparameters and training setups for this system, we attribute it to an inherent instability at this timescale. As $\t$ approaches $\t_{\mathrm{max}}$, the agreement degrades further: the simulations no longer recover all basins and, in some cases, evaluations at $\t=\t_{\mathrm{max}}$ diverge.

\begin{figure}[H]
    \centering
    \includegraphics{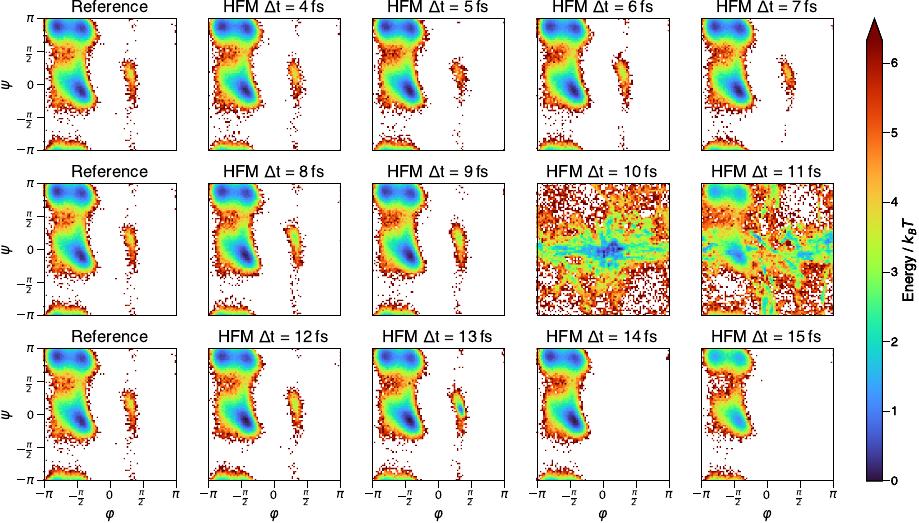}
    \caption{Comparing the free energy surfaces projected onto the dihedral angles $\varphi,\psi$ for an HFM evaluated at different step sizes $\t$. Each simulation was initialized from $10$ starting points, totaling $1$\,$\mu$s of simulation time, and was downsampled to 100k samples for visualization.}
    \label{app:fig:aldp_ramachandran}
\end{figure}

To analyze the deviation of our HFM model for $\t = 10$\,fs and $\t = 10$\,fs, we conduct additional reference simulations using the Amber ff99SB-ILDN force field in \texttt{OpenMM}~\citep{eastman2017openmm}, which was also used to generate the ground truth data for training~\citep{kohler2021smooth}. From those simulations, we compute the element-wise spectra to analyze vibrational modes (see \cref{app:fig:aldp_freqs}).

We find that the fast hydrogen-X vibrations have periods of $10$\,fs and $11$\,fs, which coincide exactly with the integration timesteps where our HFM model becomes instable in our experiments. We argue that as we hit the period of the hydrogen vibrations exactly with our integration timestep, it is much harder for the model to make an accurate prediction, since it needs to learn that large intra-step oscillations cancel out exactly for this timestep. Especially if our model makes a relatively small frequency error, i.e., it predicts a slightly larger or smaller time step, the relative error in phase space may accumulate quickly. For larger integration timesteps, this phenomenon vanishes and rollouts become stable again.

\begin{figure}[H]
    \centering
    \includegraphics[width=\textwidth]{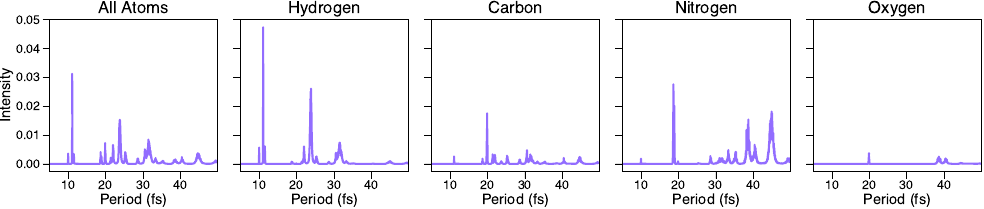}
    \caption{Vibrational frequencies for $1$ns simulation of alanine dipeptide using the Amber ff99SB-ILDN force field in OpenMM.}
    \label{app:fig:aldp_freqs}
\end{figure}

\paragraph{Accuracy vs.\ timestep for alanine dipeptide.}
To quantify agreement between the predicted free energy surface and the reference distribution from \citet{kohler2021smooth}, we report the mean Jensen--Shannon divergence (MJS) and the potential mean force (PMF) error following \citet{durumeric2024learning} with the hyperparameters from \citep{plainer2025consistent}. Both metrics compare the 2D free energy surface projection, with the PMF error weighting low-probability regions more strongly due to the use of $\log$. In \cref{app:tab:aldp-pmf-js}, we observe that the deviation from the reference increases with the step size~$\t$. For $\t \approx 10$\,fs, simulations tend to become unstable and no longer yield meaningful results. 

\begin{table}[H]
\centering
\caption{Alanine dipeptide accuracy across different timesteps $\t$ (fs). We report the PMF error and MJS. Lower values are better.}
\label{app:tab:aldp-pmf-js}
\resizebox{\linewidth}{!}{%
\begin{tabular}{l|cccccccccccc}
\hline
Metric & 4 & 5 & 6 & 7 & 8 & 9 & 10 & 11 & 12 & 13 & 14 & 15 \\
\hline
PMF & 0.041 & 0.049 & 0.045 & 0.044 & 0.049 & 0.048 & 29.676 & 10.480 & 0.053 & 0.169 & 0.117 & 0.170 \\
MJS & 0.0039 & 0.0046 & 0.0042 & 0.0041 & 0.0048 & 0.0045 & 0.5380 & 0.2419 & 0.0051 & 0.0120 & 0.0067 & 0.0094 \\
\hline
\end{tabular}
}
\end{table}

\paragraph{3D conformations.} \Cref{app:fig:aldp_random_frames} shows ten randomly selected snapshots from the alanine dipeptide trajectory generated by the HFM at $\t=12$\,fs.

\begin{figure}[H]
    \centering
    \includegraphics[width=0.6\linewidth]{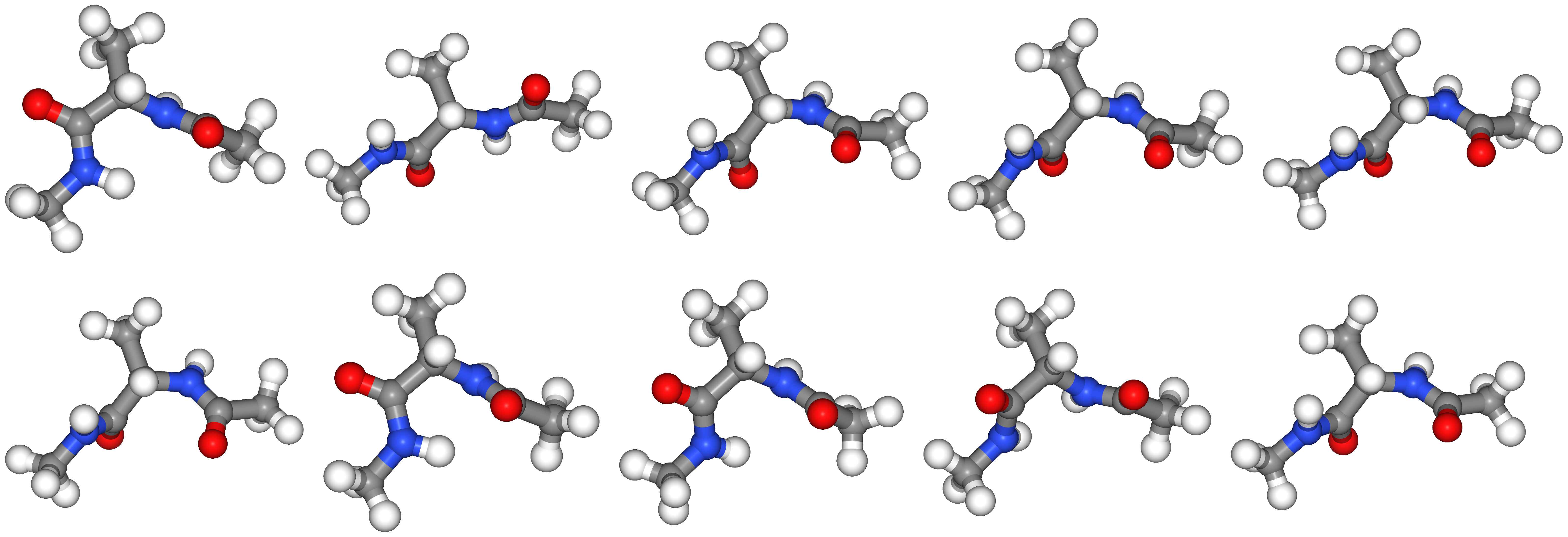}
    \caption{3D renderings of ten randomly selected frames from the alanine dipeptide simulation using the HFM evaluated at $\t=12$\,fs.}
    \label{app:fig:aldp_random_frames}
\end{figure}

\subsection{Lyapunov Time Analysis}

To evaluate whether the integrator preserves physical error accumulation without introducing spurious instabilities, we analyze the divergence of perturbed rollouts by measuring the system's characteristic error growth rate. A structurally stable integrator should match the intrinsic chaotic timescales of the underlying continuous physical system and not substantially exceed them. Following the methodology outlined in~\citet{bigi2025flashmd} Appendix M, we quantify this behavior by calculating the empirical Lyapunov times. 

As summarized in~\cref{app:tab:lyapunov_ethanol}, the Lyapunov times obtained via HFM closely match the reference molecular dynamics (MD) baseline across the evaluated integration step sizes. This demonstrates that HFM accurately reproduces the true chaotic dynamics and error propagation of the system.

\begin{table}[H]
\centering
\caption{Empirical Lyapunov times for Ethanol. We report the characteristic error growth rate in fs across different integration step sizes.}
\label{app:tab:lyapunov_ethanol}
\begin{tabular}{l|c|cccc}
\hline
Dataset & MLFF & \multicolumn{4}{c}{Hamiltonian Flow Map} \\
& 0.5\,fs & 0.5\,fs & 1\,fs & 2\,fs & 5\,fs \\
\hline
Ethanol & 199 & 186 & 186 & 192 & 211 \\
\hline
\end{tabular}
\label{tab:lyapunov_ethanol}
\end{table}

\subsection{Comparison to Learning from Trajectory Data} \label{app:subsec:data-efficiency}
To assess the \textit{data efficiency} of our approach, we extend our evaluation by comparing our trajectory-free method against trajectory-based training under a fixed computational budget for generating training data. In molecular modeling, representative geometries are often relatively cheap to obtain (e.g., via semi-empirical methods), making expensive ground-truth force evaluations the primary bottleneck. While our approach uses these forces directly as training inputs for isolated states, trajectory-based models require explicit simulations, which demands iteratively computing forces and next states. We conduct experiments for MD17 Paracetamol~\cite{chmiela2017machine} and MD22 Ac-Ala3-NHMe~\cite{chmiela2023accurate}.

Following the setup of TrajCast~\cite{thiemann2025forcefree}, we generate baseline trajectories with 0.5fs steps until the force evaluation budget is exhausted by initializing NVE runs from 5 randomly subsampled geometries of the MD17/MD22 datasets. We use our pre-trained MLFF as proxy for computing ground truth forces. We extract all pairs of states with a temporal separation of exactly 7~fs. As shown in~\cref{app:tbl:data-efficiency-para} and~\cref{app:tbl:data-efficiency-ala3}, our trajectory-free consistency objective consistently achieves lower errors for the same force labeling costs. This enhanced data efficiency is particularly pronounced in the low-data regime (\cref{app:fig:data-efficiency-para} and~\cref{app:fig:data-efficiency-ala3}). Even if trained with a smaller budget of ground truth force evaluations, our model remains significantly closer to the true physical distribution during sampling rollouts.

\begin{table}[H]
    \centering
    \caption{Data efficiency comparison of our training objective (HFM) and the Trajectory Matching (TM) objective~\cite{thiemann2025forcefree,bigi2025flashmd} across different training data budgets for Paracetamol in NVT simulations (1 ns, 10 replicas, 7fs steps). Training data budgets are measured in terms of the number of force evaluations for creating them. We compute the potential mean force (PMF) error and mean Jensen-Shannon (MJS) divergence to the free energy surface produced with an accurate small-stepsize MLFF.}
    \label{app:tbl:data-efficiency-para}
    \begin{tabular}{l|cccc}
        \hline
        Force Evaluations & PMF (HFM) $\downarrow$ & MJS (HFM) $\downarrow$ & PMF (TM) $\downarrow$ & MJS (TM) $\downarrow$ \\
        \hline
        128  & 0.136 & 0.0132 & \faBolt & \faBolt \\
        256  & 0.129 & 0.0121 & 5.772   & 0.2696  \\
        1024 & 0.127 & 0.0120 & 0.326   & 0.0312  \\
        85k  & 0.124 & 0.0119 & 0.122   & 0.0115  \\
        \hline
    \end{tabular}
\end{table}

\begin{table}[H]
    \centering
    \caption{Data efficiency comparison of our training objective (HFM) and the Trajectory Matching (TM) objective~\cite{thiemann2025forcefree,bigi2025flashmd} across different training data budgets for Ac-Ala3-NHMe in NVT simulations (3 ns, 10 replicas, 7fs steps). Training data budgets are measured in terms of the number of force evaluations for creating them. We compute the potential mean force (PMF) error and mean Jensen-Shannon (MJS) divergence to the free energy surface produced with an accurate small-stepsize MLFF.}
    \label{app:tbl:data-efficiency-ala3}
    \begin{tabular}{l|cccc}
        \hline
        Training Data & PMF (HFM) $\downarrow$ & MJS (HFM) $\downarrow$ & PMF (TM) $\downarrow$ & MJS (TM) $\downarrow$ \\
        \hline
        2048 & 0.258 & 0.0246 & 1.484 & 0.1054 \\
        85k  & 0.184 & 0.0159 & 0.523 & 0.0496 \\
        256k & N/A   & N/A    & 0.185 & 0.0176 \\
        \hline
    \end{tabular}
\end{table}

\begin{figure}[H]
    \centering
    \includegraphics{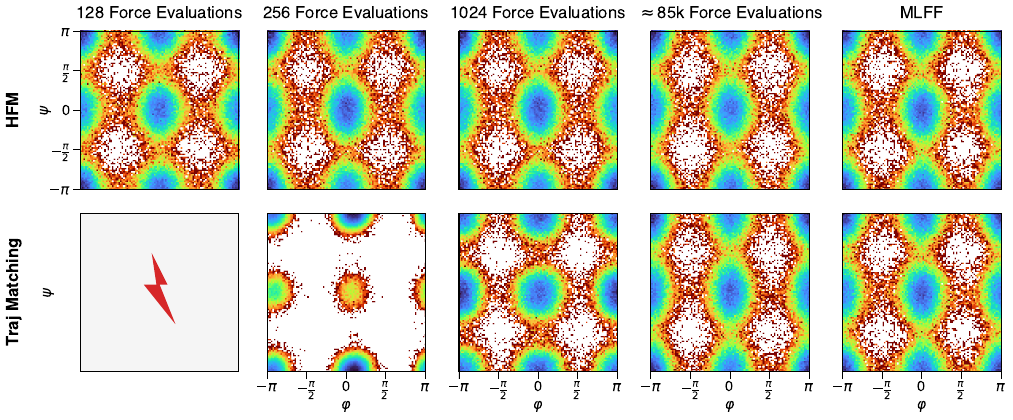}
    \caption{Ramachandran plots for Paracetamol in NVT simulations (1 ns, 10 replicas, 7fs steps) comparing our training objective (HFM) and the Trajectory Matching (TM) objective~\cite{thiemann2025forcefree,bigi2025flashmd} across different training data budgets in terms of the number of force evaluations.}
    \label{app:fig:data-efficiency-para}
\end{figure}

\subsection{Analysis of Transition Probabilities for Chignolin} \label{app:subsec:transition-rates-chignolin}

For proteins, transition rates between metastable states are a critical time-dependent property. On Chignolin, we evaluate this by estimating the transition probabilities with a Markov chain following the definition in \citet{plainer2025consistent}. As shown in~\cref{app:tab:transition_states}, HFM obtains very low mean Jensen-Shannon divergence (JSD) values compared to the MLFF reference simulation. Even at a large timestep of $\Delta t=25$\,fs, we achieve a JSD of $8.11 \cdot 10^{-4}$, indicating accurate recovery of metastable transitions (see ~\cref{app:fig:transition_states}). To test the limits of our model, we include results up to $\Delta t=45$\,fs.

\begin{figure}[H]
    \centering
    \includegraphics{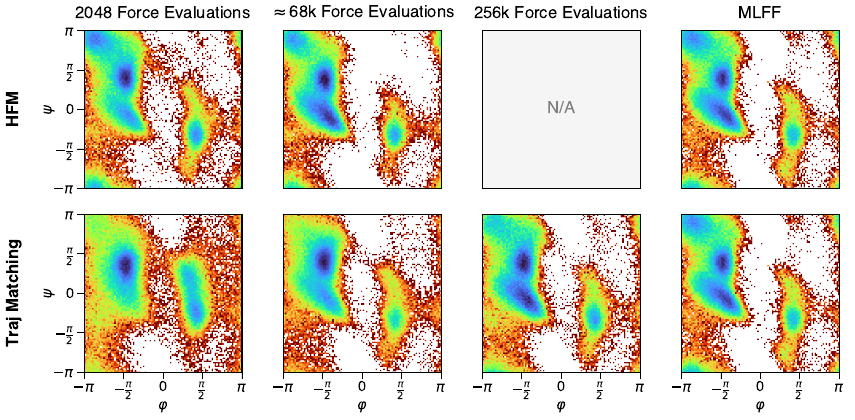}
    \caption{Ramachandran plots for Ac-Ala3-NHMe in NVT simulations (3 ns, 10 replicas, 7fs steps) comparing our training objective (HFM) and the Trajectory Matching (TM) objective~\cite{thiemann2025forcefree,bigi2025flashmd} across different training data budgets in terms of the number of force evaluations.}
    \label{app:fig:data-efficiency-ala3}
\end{figure}

\begin{figure}[H]
    \centering
    \includegraphics[width=\linewidth]{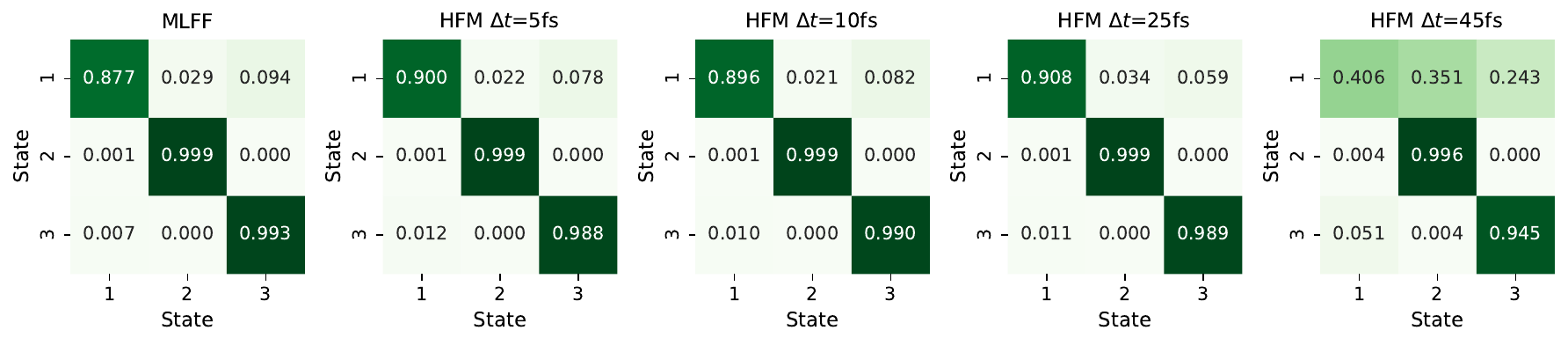}
    \caption{Comparison of transition probabilities for Chignolin. We assigned each 3D conformation to a single state and then estimated the transition probabilities with a Markov chain.}
    \label{app:fig:transition_states}
\end{figure}

\begin{table}[H]
    \centering
    \caption{Computing the mean Jensen-Shannon divergence (JSD) of transition rates of Chignolin comparing HFM to an MLFF reference simulation. We have assigned each 3D conformation of chignolin to one metastable state and estimate the transition probabilities with a Markov chain. We then compute the mean JSD of these transition probabilities.}
    \label{app:tab:transition_states}
    \begin{tabular}{lrrr}
        \toprule
        \textbf{Method} & \textbf{Mean JSD}  \\
        \midrule
        HFM $\Delta t=5$\,fs & $3.63 \cdot 10^{-4}$  \\
        HFM $\Delta t=10$\,fs & $2.35 \cdot 10^{-4}$ \\
        HFM $\Delta t=25$\,fs & $8.11 \cdot 10^{-4}$ \\
        HFM $\Delta t=45$\,fs & $5.11 \cdot 10^{-2}$ \\        
        \bottomrule
    \end{tabular}
\end{table}

\subsection{Detailed Analysis of Rollouts for BBA}
\label{app:subsec:bba-individual-rollouts}

We analyze the individual rollouts (per replica) for HFM and the CG MLFF to better understand the difference in sampling of the modes. Due to the larger time-steps, the HFM generates fewer samples per rollout but explores the distribution well (\cref{app:fig:hfm-bba-rollouts}). The CG MLFF seems less stable and gets stuck in some cases which leads to visible artifacts (\cref{app:fig:mlff-bba-rollouts}). 

\begin{figure}[H]
    \centering
    \includegraphics{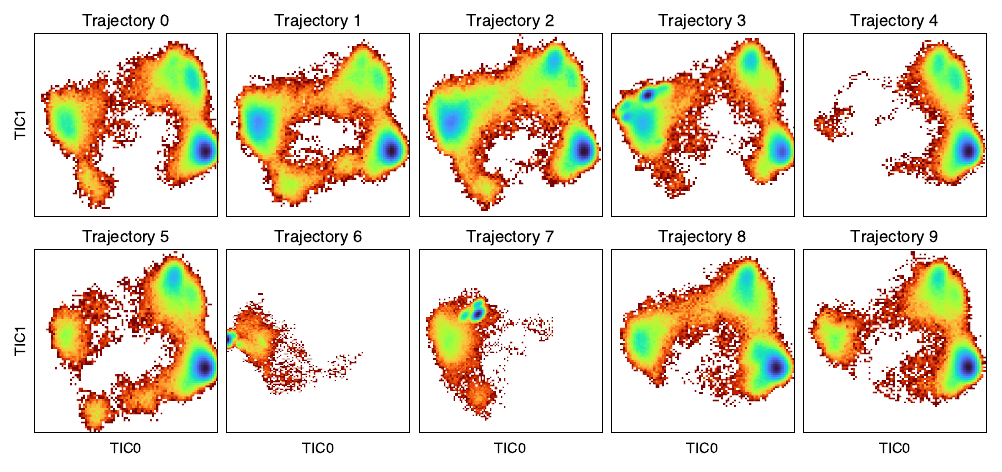}
    \caption{BBA rollouts using CG MLFF with 10 replicas.}
    \label{app:fig:mlff-bba-rollouts}
\end{figure}

\begin{figure}[H]
    \centering
    \includegraphics{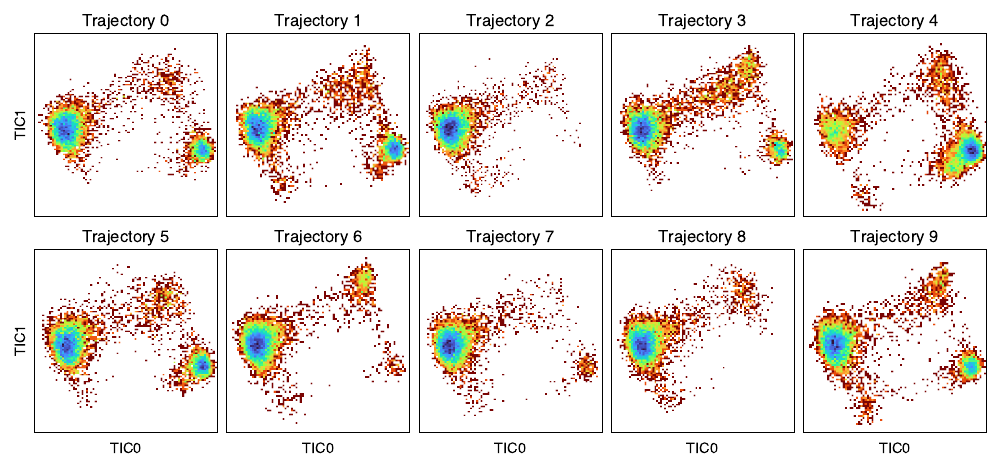}
    \caption{BBA rollouts using HFM with 10 replicas.}
    \label{app:fig:hfm-bba-rollouts}
\end{figure}

\subsection{3D Conformations of BBA}
\label{app:subsec:bba-3d-conformations}
In \cref{app:fig:hfm-bba-3d}, we verify the physical plausibility of the states our HFM model produces by plotting 3D renderings of random states of the BBA simulation. We can see that no atoms clash and the correct secondary structure is recovered.

\begin{figure}[H]
    \centering
    \includegraphics[width=\linewidth]{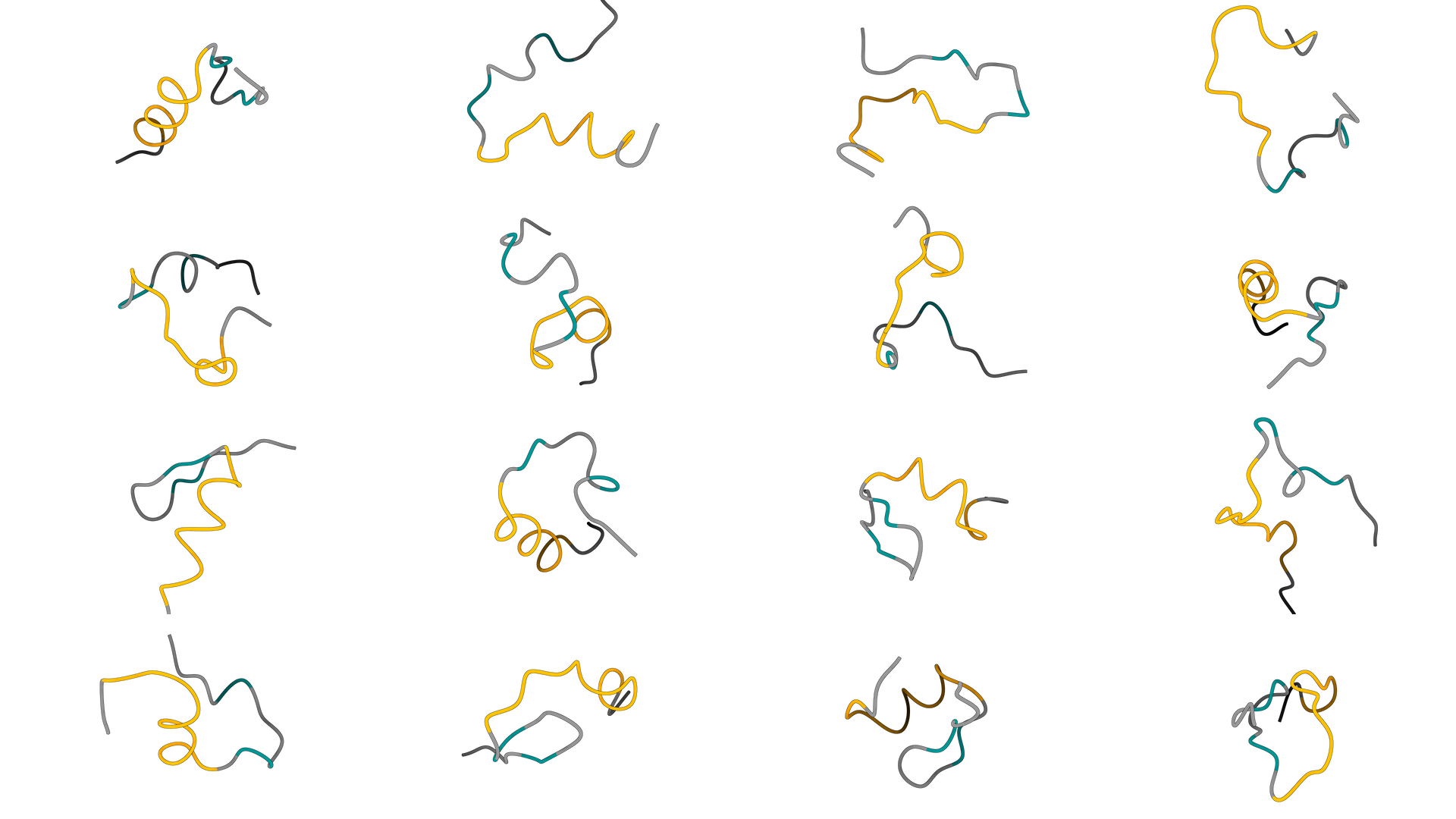}
    \caption{3D renderings of 16 randomly selected frames from the BBA simulation using the HFM evaluated at $\Delta t=15$\,fs.}
    \label{app:fig:hfm-bba-3d}
\end{figure}
}

\end{document}